\documentclass[preprint,12pt]{elsarticle}

\usepackage{amssymb}
\usepackage{amsmath}

\usepackage{bm}
\usepackage{rotating}
\usepackage{graphicx}
\usepackage{booktabs}
\usepackage{makecell}
\usepackage{caption}
\usepackage{pifont}

\usepackage{pdflscape}
\usepackage{pifont}

\newcommand{\cmark}{\ding{51}}
\newcommand{\xmark}{\ding{55}}

\usepackage{tikz}
\newcommand{\pmark}{%
\tikz[baseline=-0.6ex]{
    \draw (0,0) circle (0.075cm);
    \fill (0,0.075cm) arc[start angle=90,end angle=270,radius=0.075cm] -- cycle;
}%
}

\usepackage{subcaption}
\usepackage{xspace}
\newcommand{\modelname}{\textit{S\kern-0.08em C\kern-0.08em C\kern-0.08em M}\xspace}

\newcommand{\kws}{\textit{K\kern-0.08em W\kern-0.08em S}\xspace}

\newcommand{\kpi}{\textit{K\kern-0.08em P\kern-0.08em I}\xspace}

\newcommand{\kpis}{\textit{K\kern-0.08em P\kern-0.08em I\kern-0.08em s}\xspace}

\DeclareRobustCommand{\method}[1]{{\fontsize{8}{12}\selectfont \textbf{#1}}}

\usepackage{booktabs}     
\usepackage{tabularx}     
\usepackage{caption}      
\usepackage{array}        
\usepackage{multirow}     
\usepackage{caption}   
\usepackage{float}     

\usepackage{makecell}
\usepackage{hyperref}

\usepackage{algorithm}%
\usepackage{algorithmicx}%
\usepackage{algpseudocode}%

\usepackage{graphicx}%
\usepackage{multirow}%
\usepackage{amsmath,amssymb,amsfonts}%
\usepackage{amsthm}%
\usepackage{mathrsfs}%
\usepackage[title]{appendix}%
\usepackage{xcolor}%
\usepackage{textcomp}%
\usepackage{manyfoot}%
\usepackage{booktabs}%
\usepackage{listings}%

\usepackage{longtable}
\usepackage{booktabs}
\usepackage{tabularx}

\usepackage{tabularx}
\usepackage{subcaption}

\journal{Nuclear Physics B}

\begin{document}

\begin{frontmatter}



\title{\modelname: Stream Cruise Control Method for Automated Drift Detection and Adaptation}

\author[unt]{Mohammad Abu-Shaira\corref{cor1}}
\author[unt]{Weishi Shi}

\cortext[cor1]{Corresponding author}

\affiliation[unt]{
    organization={Department of Computer Science and Engineering, University of North Texas},
    addressline={1155 Union Circle \#311366},
    city={Denton},
    postcode={76203},
    state={Texas},
    country={USA}
}




\begin{abstract}
Real-world datasets often exhibit evolving distributions, known as concept drift. Ignoring drift degrades predictive performance, while reliance on fixed hyperparameters further limits model adaptability under changing conditions. Adaptive learning addresses this challenge by continuously updating models online, allowing them to incrementally adjust and remain effective as data distributions evolve. This paper presents the Stream Cruise Control Method (\modelname), a comprehensive framework for drift detection and adaptation in online regression. \modelname{} enables automated adaptation through early-response, pre-update drift detection, drift magnitude quantification, KPI-window-based thresholding for local false-alarm mitigation, dynamic hyperparameter tuning, and model recalibration. \modelname{} also adopts an in-memory design for real-time adaptability, unlike purely reactive methods that typically activate adaptation only after performance degradation is observed. By using dynamic thresholding and remaining agnostic to data distributions, \modelname{} supports KPI-based monitoring across varying data streams, including high-dimensional and large-scale settings. \modelname{} is integrated with four online regression models and evaluated on 18 synthetic datasets covering abrupt, incremental, and alternating gradual drift, together with eight real-world datasets. The evaluation uses both $R^2$ and MSE and compares against eight detector--adaptation baselines. Results show improved predictive performance and effective drift handling across the evaluated online regression settings.
\end{abstract}




\end{frontmatter}

\section{Introduction}
The increasing dynamism and non-stationarity of modern environments highlight the growing need for online learning methods, as traditional batch approaches often fall short under such conditions. Online learning addresses these challenges by processing data incrementally, enabling real-time responses and continuous adaptation to evolving patterns. As a result, conventional batch-trained models become inadequate for applications requiring timely updates. For example, relying solely on historical data is ineffective for tasks such as traffic management in smart cities or real-time stock trend forecasting. Likewise, adaptive algorithms are crucial for autonomous vehicles, allowing them to respond swiftly to changing road conditions and traffic dynamics.

The presence of hyperparameters in online learning models poses a significant challenge, particularly given that online learning typically operates under non-stationary conditions, where data is not independent and identically distributed (\textit{i.i.d.}). Such evolving conditions, together with the dynamic nature of the data environment, lead to a phenomenon known as concept drift \cite{webb2016characterizing, ditzler2012incremental}, wherein the underlying data distribution evolves over time. Consequently, it is impractical to expect users to continually tune these parameters with each occurrence of concept drift \cite{barbaro2018tuning}.

While online learning models are designed to address concept drift by continually updating with incoming data, their adaptation is not instantaneous. In many cases, performance degradation occurs before the model fully adjusts to the new data distribution. To mitigate this lag, adaptation mechanisms are employed. In this context, adaptive learning refers to the process of updating predictive models in real time to respond to shifts in the underlying data distribution \cite{gama2014survey}. The ability to adapt to such concept drift can be seen as a critical component and a natural extension for online learning systems \cite{giraud2000note} that learn predictive models incrementally with each example.

Designing adaptive predictive models for dynamic environments requires careful attention to several critical factors. The system must be able to detect anomalies that may signal potential drift, while also distinguishing these signals from random noise to avoid unnecessary adjustments. It should remain flexible in adapting to genuine changes yet robust enough to maintain stability in noisy conditions. Equally important is operational efficiency, ensuring that adaptation occurs within the constraints of processing time and available memory resources \cite{gama2014survey}.

Practitioners often address concept drift by coupling drift detection methods with periodic model updates or explicit adaptation techniques \cite{celik2023online}. Another common strategy is to employ ensemble learning algorithms that combine the predictions of multiple learners to improve robustness under changing data distributions \cite{de2019overview}. While these strategies can be effective in certain scenarios, they still exhibit notable limitations in dynamic and resource-constrained streaming environments.

Most existing drift detection techniques experience substantial detection delay due to their reactive nature. Specifically, they require a sufficient number of post-drift observations before recognizing that a drift has occurred \cite{gama2004learning,bifet2007learning,ross2012exponentially}. By that point, the model may have already processed several affected observations, causing predictive performance to degrade before any corrective action is initiated \cite{gonccalves2014comparative,barros2018survey}. As a result, subsequent adaptation mechanisms may be activated only after the model has already been adversely influenced by the new data distribution, making the adjustment process delayed rather than early-responsive.

In addition, many detectors provide only a binary decision, such as drift or no drift, rather than estimating the magnitude or severity of the distributional change \cite{gama2014survey,lu2019learning}. Consequently, they do not directly determine how much adaptation is required after drift is detected. The adaptation step, when included, is typically handled as a separate mechanism and is not explicitly controlled by a drift-magnitude-aware adaptation quantity \cite{zliobaite2016overview,khamassi2018discussion}. This separation limits the ability of existing detector-adaptation pipelines to perform timely and proportionate model updates.

Ensemble methods, on the other hand, can improve robustness by maintaining multiple learners, but they often introduce substantial computational overhead due to their more complex structure. This overhead can hinder real-time deployment, especially in high-velocity or resource-constrained data stream environments. These limitations underscore the need for a unified, efficient, and versatile adaptation framework capable of reducing detection latency and handling diverse drift types without imposing excessive computational demands.

To address these challenges, this paper introduces \modelname, a model-agnostic, meta-adaptive framework for online regression under concept drift. \modelname operates as a pre-update control layer that monitors predictive performance, detects potential drift, quantifies its magnitude, and uses this information to regulate the hyperparameters of the underlying online learner before the model is updated with incoming data. Rather than directly replacing the learner or modifying its internal structure, \modelname acts at a higher control level to guide how the learner adapts to evolving data. Unlike conventional drift detectors such as DDM, EDDM, ADWIN, KSWIN, and ITA, which primarily provide a reactive drift signal without estimating drift severity or prescribing how the model should adapt, \modelname integrates drift detection, drift-magnitude quantification, and adaptive hyperparameter control into a unified process. Similarly, unlike standalone adaptation strategies such as RESET and WINDOW, or hyperparameter adaptation methods such as OHL and SSPT that typically require an external trigger, \modelname jointly determines when adaptation is needed and how strongly the learner should be adjusted. This enables early-responsive and proportionate adaptation while preserving the lightweight nature required for online data stream environments.

Algorithmically, existing drift-handling methods are largely decoupled pipelines, where detection, adaptation, and parameter tuning are treated as independent components. For instance, detectors such as DDM, EDDM, ADWIN, FHDDM, and KSWIN primarily identify statistically significant changes in error streams or data distributions, but they do not directly determine how the model should adapt. Adaptation mechanisms, such as model reset or sliding-window retraining, are typically applied as external policies, while hyperparameter tuning is either performed offline or requires additional optimization procedures. As a result, these approaches lack a unified mechanism to translate detected changes into calibrated, model-specific adaptation actions.

In contrast, \modelname introduces a distinct algorithmic mechanism rather than merely combining existing components. First, \modelname maintains a bounded \kpi-Win memory that stores recent predictive behavior and uses the current incoming instance as a transient pre-update observation, allowing the framework to assess potential drift before the underlying model is modified. Second, instead of relying on fixed warning/drift thresholds or two-window statistical tests, \modelname derives a CFAR-inspired adaptive threshold from the distribution of recent \kpis, linking the detection boundary to a user-specified nominal per-decision sensitivity level under the local \kpi-window approximation. Third, \modelname computes a drift magnitude score and places the incoming observation into severity regions, distinguishing stable behavior, incremental drift, and abrupt drift. Fourth, this magnitude is not used only for reporting; it is mapped through a scale-map control mechanism into bounded, model-specific hyperparameter adjustments, such as the OLR-WA combination weight, the RLS forgetting factor, or the PA regularization parameter. Finally, if the optimized hyperparameter is insufficient to restore stable performance, \modelname performs bounded recalibration using only a limited number of additional incoming observations. Therefore, unlike conventional detector-adaptation pipelines, \modelname jointly determines when adaptation is needed, how severe the change is, how strongly the learner should be adjusted, and whether additional recalibration is required, all within a lightweight pre-update control layer.

Online hyperparameter-free models are important because repeatedly fine-tuning model hyperparameters after each occurrence of concept drift is impractical in strict online settings. In this work, we distinguish between \textit{hyperparameters} and \textit{configuration parameters}. Configuration parameters are fixed at deployment time and reflect user-defined operational preferences rather than quantities optimized during learning. For example, the threshold multiplier $z$ controls drift sensitivity: a lower value such as $z=1.5$ makes the framework more sensitive to subtle changes, whereas a higher value such as $z=2.5$ makes it more conservative. Similarly, the selected KPI reflects the user's monitoring objective, such as accuracy, loss, $R^2$, or MSE. In contrast, \textit{hyperparameters} are model-specific settings that directly govern the learning dynamics and usually require tuning or validation, such as the learning rate, regularization strength, forgetting factor, or model-combination weights. Therefore, describing \modelname as tuning-free means that, once the deployment-level configuration is specified, \modelname adaptively controls the underlying model hyperparameters during the stream without requiring repeated offline tuning, validation-based search, or manual retuning after drift. This makes \modelname suitable for practical online data streams where data arrive sequentially and revisiting past observations is limited or infeasible.

Building on these design principles, \modelname is implemented as a modular plug-in that can be integrated with online regression models through their adaptive hyperparameters and update rules. Rather than requiring changes to the internal structure of the learner, \modelname observes the learner's predictive behavior, adjusts the appropriate hyperparameter according to the estimated drift severity, and returns the updated model state for continued online learning. In this sense, \modelname can be characterized as a meta-adaptive layer, because it adapts the adaptation behavior of the underlying online learner by regulating its model-specific hyperparameters rather than replacing or redesigning the learner itself. This design gives the framework bounded memory and constant-size control state,
while supporting model-agnostic deployment in streaming environments.

In this study, \modelname is evaluated with multiple online regression models under abrupt, incremental, and alternating gradual drift scenarios, as well as under varying dataset characteristics. The experimental evaluation compares \modelname against standalone drift detectors, adaptation strategies, and online hyperparameter tuning methods to assess both predictive performance and drift-handling effectiveness.

\noindent The main contributions of this work are as follows:

\begin{itemize}

\item \textbf{A bounded-memory pre-update control framework for online regression.}
We introduce \modelname{} as a pluggable meta-adaptive layer that operates before model updates, enabling early intervention when predictive performance begins to degrade. Unlike conventional drift-handling pipelines that react only after a drift alarm, \modelname{} monitors each incoming point or mini-batch before updating the learner, allowing timely adaptation while preserving computational efficiency.

\item \textbf{A KPI-driven and distribution-agnostic drift detection mechanism.}
We propose a detection strategy that operates directly on predictive performance indicators rather than relying on raw distributional comparisons or fixed detector assumptions. Using a bounded \kpi-window memory and CFAR-inspired local thresholding, \modelname{} dynamically calibrates the threshold \(\tau\) from recent \kpi{} behavior, making the detection boundary responsive to different KPIs, data streams, and short-term variability. Because \modelname{} performs KPI-based detection, it avoids direct high-dimensional feature-space comparisons and is therefore lightweight and suitable for high-dimensional data streams.

\item \textbf{Drift magnitude estimation for proportional adaptation.}
Instead of producing only a binary drift/no-drift decision, \modelname{} quantifies the severity of performance deviation and uses it as a control signal. This allows the framework to distinguish stable behavior, minor drift, and severe drift. Since \modelname{} monitors performance at the point or mini-batch level, it can also respond to local drift when a locally affected group of samples produces measurable degradation in the monitored \kpi{}.

\item \textbf{Dynamic hyperparameter control through deployment-time configuration.}
We develop a scale-map-based mechanism that transforms drift magnitude into bounded, model-specific hyperparameter adjustments during stream processing. This reduces reliance on unnecessary fixed model hyperparameters and replaces repeated offline tuning with lightweight deployment-time configuration parameters that control adaptation behavior. As a result, \modelname{} avoids costly hyperparameter search while remaining suitable for real-time streaming environments.

\item \textbf{A unified adaptation strategy with bounded recalibration.}
We integrate detection, magnitude quantification, hyperparameter tuning, and
model recalibration into a single control loop. Recalibration is triggered
conditionally when severe drift persists after hyperparameter adjustment and
uses a limited recent memory, supporting adaptive stability, scalability, and
resource-efficient operation under evolving stream conditions.

\end{itemize}

The source code and datasets used in this study are publicly available through our GitHub repository\footnote{\url{https://github.com/anonymous273800/SCCM}}.

\section{Related Work}

In concept drift research, much of the focus has been on classification problems, leaving regression, which involves continuous target variables, relatively less explored \cite{celik2021adaptation}. This section reviews three main categories of related work: concept drift detectors, concept drift adaptation techniques, and automated drift adaptation frameworks. The discussion also highlights the methods used as experimental baselines in this study, including ADWIN, KSWIN, RESET, WINDOW, OHL, and SSPT.

\subsection{Concept Drift Detectors}

Drift detection techniques are commonly categorized into distribution-oriented and performance-oriented methods \cite{maheshwari2024drift}. Distribution-oriented methods identify changes in the underlying data distribution, while performance-oriented methods monitor predictive behavior, such as increases in error, to infer the occurrence of concept drift.

DDM \cite{gama2004learning} is an early performance-based detector derived from the PAC learning assumption that the learner's error rate should decrease as more samples are observed under a stationary distribution. It uses warning and drift thresholds to identify possible concept changes. However, DDM is primarily designed for classification and is more effective for abrupt drift than gradual drift. EDDM \cite{baena2006early} extends DDM by monitoring the distance between classification errors, making it more suitable for gradual drift. Nevertheless, it remains classification-oriented and depends on predefined parameters.

ADWIN \cite{bifet2007learning} is a widely used adaptive-window drift detector with statistical guarantees. It detects changes by comparing statistics across subwindows within a dynamically maintained window and removes older observations when a significant change is detected. Although ADWIN is generic and effective in many streaming settings, it may suffer from detection delay and memory overhead due to the need to maintain and update a variable-size window. In this study, ADWIN is used as one of the main drift-detection baselines.

PL \cite{bach2008paired} employs a stable learner and a reactive learner. The stable learner is trained using a longer history of data, while the reactive learner focuses on recent observations. Drift is detected when the reactive learner performs better than the stable learner. Although useful, this strategy introduces additional computational and memory costs because multiple learners must be maintained.

KSWIN \cite{raab2020reactive} is a non-parametric drift detector based on the Kolmogorov--Smirnov statistical test. It compares recent observations with a reference window to detect distributional changes. Its main advantage is that it does not assume a specific data distribution. However, its performance is sensitive to the window size, split ratio, and significance level. Since the KS test is univariate, KSWIN may also be less suitable for high-dimensional data unless dimensions are monitored separately. In this study, KSWIN is used as the second main drift-detection baseline.

ITA \cite{dasu2006information} detects distributional changes using information-theoretic measures such as relative entropy. It partitions the data space using a kdq-tree and compares distributions through Kullback--Leibler divergence. While ITA is generic and non-parametric, it can be computationally expensive for high-dimensional or high-speed data streams.

\renewcommand{\arraystretch}{1}

\begin{longtable}{p{2.1cm} p{11cm}}
\caption{Summary of Concept Drift Detection Techniques} 
\label{tab:related-work-summary-drift-detection-techniques}\\
\toprule
\textbf{Method} & \textbf{Features (\texttt{+}) and Limitations (\texttt{-})} \\
\midrule
\endfirsthead

\multicolumn{2}{c}%
{{\bfseries \tablename\ \thetable{} -- continued}} \\
\toprule
\textbf{Method} & \textbf{Features (\texttt{+}) and Limitations (\texttt{-})} \\
\midrule
\endhead

\midrule \multicolumn{2}{r}{{Continued on next page}} \\
\endfoot

\bottomrule
\endlastfoot

\textbf{DDM}\cite{gama2004learning} & 
\texttt{+} • Simple • Computationally efficient • Provides warning and drift levels. \newline
\texttt{-} • Mainly designed for classification • Less effective for gradual drift • Depends on fixed warning and drift thresholds • Assumes i.i.d. Bernoulli errors. \\

\textbf{EDDM}\cite{baena2006early} &
\texttt{+} • Simple • Improves gradual drift detection compared with DDM. \newline
\texttt{-} • Mainly designed for classification • Requires careful parameter tuning • Needs a minimum number of errors before detection. \\

\textbf{ADWIN}\cite{bifet2007learning} &
\texttt{+} • Generic • Adaptive windowing • Statistical guarantees • Used as a drift-detection baseline in this study. \newline
\texttt{-} • Detection delay • Computational overhead from comparing subwindows • Requires setting the confidence parameter ($\delta$). \\

\textbf{PL}\cite{bach2008paired} &
\texttt{+} • Uses stable and reactive learners • Can detect drift through performance differences. \newline
\texttt{-} • Higher computation and memory cost • Requires maintaining two learners • Depends on fixed reactive window size and replacement threshold. \\

\textbf{KSWIN}\cite{raab2020reactive} &
\texttt{+} • Non-parametric • No distributional assumption • Based on the Kolmogorov--Smirnov test • Used as a drift-detection baseline in this study. \newline
\texttt{-} • Sensitive to significance level, window size, and split ratio • May suffer from detection delay • Less suitable for high-dimensional streams because the KS test is univariate. \\

\textbf{ITA}\cite{dasu2006information} &
\texttt{+} • Generic • Non-parametric • Uses information-theoretic distribution comparison. \newline
\texttt{-} • Computationally expensive for high-dimensional streams • Requires multidimensional density estimation • Sensitive to window size and significance settings. \\

\end{longtable}

\subsection{Concept Drift Adaptation Techniques}

\subsubsection{Hyperparameters Optimization}

Hyperparameter optimization has been widely studied in batch learning, where the full dataset is available before model deployment. Common model-free methods include Grid Search \cite{hutter2019automated}, which exhaustively evaluates predefined configurations, and Random Search \cite{bergstra2012random}, which samples configurations from predefined distributions. Resource-allocation methods such as Successive Halving \cite{jamieson2016non} discard weak configurations iteratively, while Hyperband \cite{li2018hyperband} extends this idea by running Successive Halving with different budget allocations. Other approaches include gradient-based optimization \cite{bengio2000gradient}, which updates hyperparameters using gradient information, and model-based methods such as Bayesian optimization \cite{domke2012generic}, which uses a surrogate model to guide the search toward promising regions of the hyperparameter space.

However, these batch hyperparameter optimization methods are designed for offline settings and are generally incompatible with online learning assumptions when applied directly. They usually require repeated training and validation over fixed datasets, whereas online learning processes examples or mini-batches sequentially, often under a single-pass constraint, where past examples may not be stored or revisited after prediction. Therefore, directly applying Grid Search, Random Search, Bayesian optimization, or cross-validation-based tuning to data streams violates the online learning assumption of limited access to past and future data. This distinction does not imply that hyperparameters cannot be optimized online, but rather that batch optimization procedures must be replaced or substantially adapted for streaming environments.

Online hyperparameter optimization therefore focuses on adapting hyperparameters during the learning process while respecting streaming constraints. Instead of repeatedly retraining models over fixed datasets, online approaches update or select hyperparameters incrementally using the information available at each time step or mini-batch. Several studies have proposed concrete algorithmic frameworks for this purpose. For example, Online Hypergradient Learning (OHL) dynamically computes hyperparameter gradients from incoming data and updates hyperparameters through an online projected hyper-gradient descent procedure \cite{zhan2018efficient}. SPOT provides a sequential parameter optimization strategy for automatically tuning algorithm parameters \cite{bartz2005sequential}. SSPT extends self-parameter tuning to data streams by using the Nelder--Mead search procedure to continuously adjust hyperparameter configurations in a single pass \cite{veloso2021hyperparameter}. Stream-oriented resource-allocation methods, such as Hyperband for Streams, adapt the budget-allocation principle of Hyperband to streaming scenarios \cite{jie2020hypertube}. HyperTube further addresses online hyperparameter optimization under constrained computational resources by organizing the search process around tube-based resource allocation rather than serving as evidence of a missing definition or framework \cite{jie2020hypertube}.

\subsubsection{Adaptation Strategies}

Adaptive learning algorithms employ several strategies to address concept drift:

\renewcommand{\theenumi}{\roman{enumi}}
\begin{enumerate}
    \item \textit{Model Replacement Strategy (RESET)}: The old model is completely discarded and a new model is built from scratch after drift detection, exemplified by OAML \cite{celik2023online}. In this study, this strategy is referred to as RESET, where the current model is reinitialized once drift is detected.

    \item \textit{Model Retraining (WINDOW)}: Upon detecting concept drift, many models initiate retraining using recently observed data, exemplified in \cite{baier2021detecting}. In this study, this strategy is referred to as WINDOW, where the model is adapted using a recent sliding window of observations rather than the full historical stream.

    \item \textit{Ensemble Learning}:
    Ensemble learning techniques have been proposed to improve data stream analysis and concept drift adaptation by combining the predictions of multiple learners. This reduces the impact of distributional changes, since the ensemble is less affected by any single learner. These methods are usually grouped into two categories: online ensembles and block-based ensembles~\cite{sun2016online}. In block-based ensembles, the data stream is divided into blocks or subsets, and each base learner is trained on a different block rather than on the entire dataset. Three popular block-based ensembles are Accuracy Updated Ensemble (AUE), Accuracy Weighted Ensemble (AWE), and Streaming Ensemble Algorithm (SEA)~\cite{krawczyk2017ensemble}. Ensemble online learning often outperforms traditional methods in dynamic data stream analytics but is computationally expensive.

    \item \textit{Updating Model Structure}: Some approaches adapt to drift by changing the structure of the predictive model. For example, CVFDT \cite{hulten2001mining} creates alternate subtrees when existing ones become unreliable, while DWM \cite{kolter2007dynamic} dynamically adds or removes weighted experts according to their performance.
\end{enumerate}

\begin{table}[H]
\centering
\caption{Summary of Concept Drift Adaptation Techniques}
\renewcommand{\arraystretch}{1}
\begin{tabularx}{\textwidth}{p{3.1cm} p{9.8cm}}
\toprule
\textbf{Method} & \textbf{Features (\texttt{+}) and Limitations (\texttt{-})} \\
\midrule
\addlinespace

\textbf{Hyperparameter Optimization 
(OHL, SSPT)}
\cite{barbaro2018tuning,zhan2018efficient,veloso2021hyperparameter} & 
\texttt{+} • Enables adaptive hyperparameter control in online settings • OHL updates hyperparameters using hyper-gradients • SSPT searches for suitable configurations using Nelder--Mead • Used as adaptation baselines in this study. \newline
\texttt{-} • May introduce additional computational cost • Performance depends on the stability of search or gradient updates under drift.
\\

\addlinespace
\textbf{Model Replacement 
(RESET)}
\cite{celik2023online} & 
\texttt{+} • Simple • Generic • Suitable for abrupt shifts • Used as an adaptation baseline in this study. \newline
\texttt{-} • Discards past knowledge • May be inefficient under gradual or recurring drift.
\\

\addlinespace
\textbf{Model Retraining 
(WINDOW)}
\cite{baier2021detecting} & 
\texttt{+} • Generic • Uses recent data for adaptation • Useful when recent observations better represent the current concept • Used as an adaptation baseline in this study. \newline
\texttt{-} • Requires storing a recent window • Sensitive to window size • May introduce retraining latency.
\\

\addlinespace
\textbf{Ensemble Learning}
\cite{guo2023pfge} & 
\texttt{+} • Generic • Combines multiple models for robust predictions. \newline
\texttt{-} • High memory and inference overhead • Requires tuning of ensemble size and update strategy.
\\

\addlinespace
\textbf{Updating Model Structure}
\cite{hulten2001mining,kolter2007dynamic} & 
\texttt{+} • Dynamically adapts model structure, such as trees or weighted experts, based on performance. \newline
\texttt{-} • Implementation complexity • Computational overhead • Often tied to specific model families.
\\

\bottomrule
\end{tabularx}
\label{tab:related-work-summary-drift-adaptation-techniques}
\end{table}

\subsection{Automated Drift Adaptation Frameworks}

Automated drift adaptation frameworks typically integrate drift detection, adaptation, and model maintenance components. Some frameworks also incorporate hyperparameter tuning or retraining mechanisms.

StepWise \cite{ma2018robust} was designed for web-based anomaly detection. It uses iSST-EVT to automatically set detection thresholds and identify abnormal spikes related to drift. However, its use case is domain-specific, which limits its general applicability. AEF-CDA \cite{maheshwari2024drift} is an adaptive ensemble framework developed for concept drift adaptation in large-scale Internet of Medical Things data streams. While effective in its target domain, it is healthcare-specific and ensemble-based, which can increase computational and resource costs. HyperTube \cite{jie2020hypertube} supports hyperparameter optimization for deep learning under constrained computational resources through parallel online learning and retraining of selected models on previous data samples. However, its reliance on stored history and deep learning-oriented design limits its suitability for lightweight online regression.

\begin{table}[H]
\centering
\caption{Summary of Concept Drift Adaptation Frameworks}
\renewcommand{\arraystretch}{1.2}
\begin{tabularx}{\textwidth}{p{2cm} p{11cm}}
\toprule
\textbf{Method} & \textbf{Features (\texttt{+}) and Limitations (\texttt{-})} \\
\midrule
\addlinespace

\textbf{StepWise}
\cite{ma2018robust} & 
\texttt{+} • Detects anomalies and drift using automatic thresholding through iSST-EVT. \newline
\texttt{-} • Designed mainly for web-based anomaly detection • Use-case specific • Limited generalizability to online regression.
\\

\addlinespace
\textbf{AEF-CDA}
\cite{maheshwari2024drift} & 
\texttt{+} • Adaptive ensemble framework for medical data streams. \newline
\texttt{-} • Domain-specific to healthcare and IoMT • Ensemble-based • Computationally and resource intensive.
\\

\addlinespace
\textbf{HyperTube}
\cite{jie2020hypertube} & 
\texttt{+} • Enables parallel online learning with adaptive hyperparameter tuning. \newline
\texttt{-} • Targeted toward deep learning under constrained compute environments • Relies on stored historical data • Limited to selected model types.
\\

\bottomrule
\end{tabularx}
\label{tab:related-work-summary-drift-adaptation-frameworks}
\end{table}

\subsection{Adaptive Hyperparameters in Online Regression}

In online regression, adaptive hyperparameters are especially important because they control the balance between historical and recent observations under non-stationary data streams. These hyperparameters are often described using terms such as forgetting factor, decay factor, memory decay rate, learning rate, aggressiveness parameter, or momentum coefficient.

For example, Recursive Least Squares (RLS) \cite{fontenla2013online} uses a forgetting factor $\lambda \in (0,1]$ to reduce the influence of older observations through exponential weighting. Online Regression with Weighted Average (OLR-WA) \cite{abu2023olr} uses a weighting parameter $\alpha \in (0,1]$ to balance time-based and confidence-based training, where larger values emphasize recent data. Passive-Aggressive (PA) regression \cite{crammer2006online} uses the aggressiveness parameter $C$ to control the trade-off between model stability and responsiveness to new observations. Widrow-Hoff (LMS) \cite{widrow1960adaptive} relies on the learning rate to determine the magnitude of online updates. Similarly, in Stochastic Gradient Descent with Momentum (SGDM), the momentum coefficient $\beta \in [0,1)$ controls the influence of past gradients on the current update \cite{liu2020improved}.

These examples illustrate that many online regression models already contain hyperparameters that implicitly regulate adaptation. However, selecting or adjusting these hyperparameters under drift remains challenging, particularly when the stream changes over time and historical data cannot be repeatedly revisited.

The comparison in Table~\ref{tab:sccm-feature-level-comparison} provides structured evidence from representative concept drift detection and adaptation methods. While it is not intended as a bibliometric survey, it shows that several influential detectors are classification-oriented, whereas generic detectors and adaptation strategies usually lack regression-specific, magnitude-aware hyperparameter control. Thus, the gap addressed by \modelname{} is not simply the absence of drift detection, but the absence of a unified regression-oriented framework that jointly performs drift detection, drift magnitude estimation, adaptive hyperparameter tuning, and bounded recalibration.

To further clarify the positioning of \modelname{} relative to existing concept drift detection and adaptation methods, Table~\ref{tab:sccm-feature-level-comparison} summarizes the main functional differences between \modelname{} and representative drift detection and adaptation methods. The table shows that most existing methods address only one component of the drift-handling process, such as detection, retraining, replacement, or hyperparameter tuning. Existing detectors usually identify drift but do not quantify its magnitude or adapt the model. Similarly, adaptation strategies such as RESET, WINDOW, OHL, and SSPT require an external drift signal. In contrast, \modelname{} uses KPI-based monitoring, drift magnitude estimation, proportional hyperparameter tuning, and bounded recalibration in a single pre-update online learning process.

\begin{sidewaystable}[!htbp]
\centering
\captionsetup{justification=centering}
\caption{Feature-level comparison between \modelname{} and representative drift detection and adaptation methods.}
\vspace{-5pt}
\label{tab:sccm-feature-level-comparison}

\tiny
\setlength{\tabcolsep}{1.5pt}
\renewcommand{\arraystretch}{1.15}

\begin{tabularx}{0.96\textheight}{@{}
>{\raggedright\arraybackslash}p{1.25cm}
>{\centering\arraybackslash}p{0.75cm}
>{\centering\arraybackslash}p{0.45cm}
>{\centering\arraybackslash}p{0.45cm}
>{\centering\arraybackslash}p{0.45cm}
>{\centering\arraybackslash}p{0.45cm}
>{\centering\arraybackslash}p{0.45cm}
>{\raggedright\arraybackslash}X
@{}}
\toprule
\textbf{Method} & 
\rotatebox{30}{\textbf{Task Type}} & 
\rotatebox{30}{\textbf{Drift Detection}} & 
\rotatebox{30}{\textbf{Drift Magnitude}} & 
\rotatebox{30}{\textbf{Adaptation}} & 
\rotatebox{30}{\textbf{Hyperparam. Tuning}} & 
\rotatebox{30}{\textbf{Unified Framework}} & 
\textbf{\hspace{2cm}Benchmarking Note} \\
\midrule

\textbf{DDM} 
& \makecell{Classi-\\fication} 
& \cmark 
& \xmark 
& \xmark 
& \xmark 
& \xmark 
& Reactive detection-only method that raises a drift signal after sufficient error-based evidence accumulates and requires an external adaptation mechanism. \\

\textbf{EDDM} 
& \makecell{Classi-\\fication} 
& \cmark 
& \xmark 
& \xmark 
& \xmark 
& \xmark 
& Reactive detection-only method based on classification error-distance behavior, with the same external-adaptation limitation as DDM. \\

\textbf{ADWIN} 
& Generic 
& \cmark 
& \xmark 
& \xmark 
& \xmark 
& \xmark 
& Adaptive-window detector that can provide delayed reactive drift signals, but adaptation must be supplied externally. \\

\textbf{KSWIN} 
& Generic 
& \cmark 
& \xmark 
& \xmark 
& \xmark 
& \xmark 
& Reactive distributional-test detector without integrated adaptation, hyperparameter tuning, or magnitude-driven control. \\

\textbf{ITA} 
& Generic 
& \cmark 
& \xmark 
& \xmark 
& \xmark 
& \xmark 
& Reactive distributional-change detector, but it does not provide drift magnitude estimation or adaptation strength selection. \\

\textbf{PL} 
& \makecell{Classi-\\fication} 
& \cmark 
& \xmark 
& \cmark 
& \xmark 
& \pmark 
& Uses stable and reactive learners to respond to drift, but adaptation is indirectly embedded and lacks explicit magnitude-driven control. \\

\makecell[l]{\textbf{Ensemble}\\\textbf{Methods}} 
& \makecell{Mostly\\classi-\\fication} 
& \cmark 
& \xmark 
& \cmark 
& \xmark 
& \pmark 
& Adapt through model weighting, replacement, or ensemble restructuring, often reactively and with higher computational or model-management cost. \\

\textbf{RESET} 
& Generic 
& \xmark 
& \xmark 
& \cmark 
& \xmark 
& \xmark 
& Adaptation policy only; it replaces or resets the model after an external drift signal and does not detect drift by itself. \\

\textbf{WINDOW} 
& Generic 
& \xmark 
& \xmark 
& \cmark 
& \xmark 
& \xmark 
& Recent-window adaptation policy that retrains or updates using recent samples after an external drift signal is provided. \\

\textbf{OHL} 
& Generic 
& \xmark 
& \xmark 
& \cmark 
& \cmark 
& \xmark 
& Hyperparameter adaptation method that requires an external trigger and does not provide integrated drift detection or magnitude estimation. \\

\textbf{SSPT} 
& Generic 
& \xmark 
& \xmark 
& \cmark 
& \cmark 
& \xmark 
& Hyperparameter tuning strategy activated after an external drift or adaptation trigger, but it does not unify drift detection, magnitude estimation, and adaptation. \\

\textbf{StepWise} 
& \makecell{Domain-\\specific} 
& \cmark 
& \xmark 
& \cmark 
& \pmark 
& \pmark 
& Reactive automated adaptation framework, but typically designed for specific settings and does not provide a general lightweight pre-update regression control layer. \\

\textbf{AEF-CDA} 
& \makecell{Ensemble-\\based} 
& \cmark 
& \xmark 
& \cmark 
& \pmark 
& \pmark 
& Ensemble-oriented concept drift adaptation framework that reacts to drift but may require additional model-management cost and lacks explicit magnitude-driven recalibration. \\

\textbf{\makecell{Hyper-\\Tube}} 
& \makecell{Online\\tuning} 
& \xmark 
& \xmark 
& \cmark 
& \cmark 
& \pmark 
& Online hyperparameter optimization framework under resource constraints, but it does not unify KPI-driven drift detection, drift magnitude estimation, and bounded recalibration. \\

\midrule

\textbf{\modelname{} (Ours)} 
& \makecell{Regres-\\sion} 
& \cmark 
& \cmark 
& \cmark 
& \cmark 
& \cmark 
& Unified pre-update framework that evaluates incoming observations before the main model update and continuously adjusts adaptation strength through KPI-driven detection, drift magnitude estimation, adaptive tuning, and bounded recalibration. \\

\midrule
\multicolumn{8}{@{}l@{}}{\footnotesize \cmark: supported; \xmark: not supported; \pmark: partially supported.} \\

\bottomrule
\end{tabularx}

\end{sidewaystable}

\subsection{Summary}

In summary, existing drift detection methods exhibit distinct strengths and limitations. Some methods, such as DDM and EDDM, are mainly classification-oriented, while others, such as ADWIN and KSWIN, are more generic and can be applied in streaming settings. However, many detectors are reactive, sensitive to parameter settings, or affected by detection delay and computational overhead.

For drift adaptation, existing strategies include hyperparameter optimization, model replacement, model retraining, ensemble learning, and structural updates. Each strategy presents trade-offs in terms of knowledge retention, computational cost, latency, and suitability for different drift types. In particular, RESET and WINDOW provide simple and interpretable adaptation mechanisms, while OHL and SSPT represent online hyperparameter adaptation strategies. Therefore, these methods are used as benchmark adaptation baselines in the experimental evaluation.

Existing automated drift adaptation frameworks are often domain-specific, resource-intensive, or designed for particular model families. This motivates the need for a general-purpose, regression-oriented, and low-latency framework that integrates drift detection, adaptive hyperparameter control, and model adaptation in a unified online learning process. A comprehensive overview of the reviewed methods is provided in Tables~\ref{tab:related-work-summary-drift-detection-techniques}--\ref{tab:related-work-summary-drift-adaptation-frameworks}.


\section{Problem Definition}

In online learning, the data stream is presumed to be infinitely long and must be handled in a sequential manner, with only a limited amount of data stored in memory at any point in time \cite{gama2014survey}. Typically, data points are processed in small increments. The data generation process can change over time, leading to what is known as \textit{concept drift}, which refers to an unexpected change in the data distribution over time.

Consider a data stream \(\{(X_1, y_1), (X_2, y_2), \dots\}\) generated from a joint probability density function \(p(X, y)\), representing the concept to be learned. Concept drift can be represented as shown in Equation \ref{eq:concept_drift}, where \(t_n\) denotes a point in time and \(t_{n+1}\) denotes the direct subsequent point in time.
{
    \begin{equation}    
    \label{eq:concept_drift}
        \exists \textit{X,y} : p_{t_n}(X, y) \neq p_{t_{n+1}}(X, y)
    \end{equation}    
}

\begin{figure}[ht]
  \centering
  \includegraphics[width=0.65\textwidth, height=0.18\textheight]{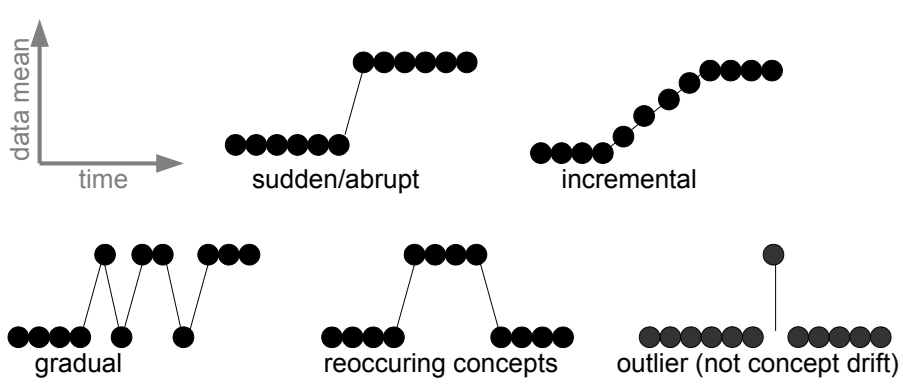}
  \caption{Concept Drift Types \cite{gama2014survey}}
  \label{fig:001_drift_types}
  \vspace{-.25cm}
\end{figure}

Authors in \cite{webb2016characterizing, gama2014survey} classify concept drift into several principal categories, as shown in Figure \ref{fig:001_drift_types}, focusing on \textit{duration} and \textit{magnitude}, which significantly influence learner selection and adaptation. Drift duration is the amount of time in which an initial concept $a_n \equiv p_{t_n}(X,y)$, at time
\(t_n\) drifts to a resulting concept $a_{n+1} \equiv p_{t_{n+1}}(X,y)$ at time \(t_{n+1}\), represented by $Duration(a_{n},a_{n+1}) = t_{n+1} - t_{n}$.

In addition to temporal drift patterns such as abrupt, incremental, and gradual drift, concept drift can also be characterized by its scope in the data stream. Global drift refers to a change that affects the overall data-generating distribution or most of the input space, whereas local drift affects only a specific subset of samples, group of instances, or subregion of the feature space \cite{webb2016characterizing,aguiar2024comprehensive}. In local drift, the overall stream behavior may remain relatively stable while only part of the input-output relationship changes.

Drift magnitude is the distance between the initial and resulting concepts over the drift period \([t_{n}, t_{n+1}]\), represented by the formula $Magnitude(t_0, t_1) = D(a_0, a_1)$, where \(D\) is a distribution distance function that quantifies
the difference between concepts at two points in time. A minor drift may require model refinement, while major ones may necessitate model abandonment \cite{webb2016characterizing}.

The effectiveness of a common concept drift-aware online learning strategy largely depends on its ability to detect concept drifts. However, due to the duration of drifts and the unknown magnitude between two drifts, along with the inherent noise in the data, relying on the original definition for drift detection becomes an intractable task. Therefore, designing a drift detection and adaptation method that is easy to implement and applicable to different types of drifts has become a major focus of our paper.

\section{Method}

This section presents the details of \modelname, explains its underlying principles, and highlights the distinctive features that make it a robust and efficient framework for automated drift detection and adaptation.

The \modelname framework comprises four core objectives. First, it provides \textit{in-memory early-response} drift detection, which aims to identify early evidence of shifts in the monitored \kpi behavior before substantial performance degradation occurs, while mitigating false alarms through a local \kpi-window-based thresholding mechanism. Second, it quantifies drift magnitude to provide insight into the significance of the detected change. Third, it performs adaptive hyperparameter optimization to sustain robust performance. Fourth, it supports model adjustment through lightweight tuned-hyperparameter inference for immediate fluctuations and more comprehensive recalibration when severe drift persists despite hyperparameter optimization.

The idea of \modelname is inspired by the cruise control function in automobiles, with its principle adapted to the context of online data streams. Similar to how cruise control continuously regulates fuel injection and braking to maintain a stable driving speed under varying road conditions, \modelname dynamically adjusts adaptive hyperparameters in real time to sustain consistent model performance. This mechanism enables the framework to seamlessly balance the incorporation of new data with the preservation of previously acquired knowledge. By doing so, \modelname not only enhances stability in the presence of fluctuations and drift but also ensures efficient long-term learning in dynamic environments.

\modelname is designed as a modular component that can be seamlessly integrated into online learning models to improve their ability to adapt to concept drift. In this paper, our focus is directed toward online regression, where we demonstrate that incorporating \modelname leads to significant performance gains. Nonetheless, the framework is model-agnostic in nature, and its applicability is not limited to regression; in principle, it can be extended to other forms of online learning tasks as well.

\modelname operates as a pre-update pluggable component, invoked before modifying the model weights $\mathbf{W}$ during the online learning process. In an online regression setting, when the model receives a new instance (or mini-batch) from the data stream, this instance is first held in memory prior to updating the model and is subsequently discarded once the update is complete. During this stage, the model retains its current weights, and \modelname evaluates the incoming instance to determine whether it indicates concept drift. If no drift is detected, the model proceeds with the standard online update step. Conversely, if drift is identified, \modelname applies corrective actions to characterize the drift and adapt the model accordingly.

\modelname systematically captures Key Performance Indicators (\kpis) associated with incoming stream observations, represented as either mini-batches or single data points. The \kpis of each encountered mini-batch are stored in a condensed, sequentially ordered window known as the \kpi-Window (\kpi-Win). Examples of such \kpis include Accuracy, Precision, Recall, F1-score, etc. The selection of stream \kpi measures is adaptable and contingent upon the specific problem domain; for instance, in \textit{regression} scenarios, common metrics encompass Mean Absolute Error (MAE), Mean Squared Error (MSE), Root Mean Squared Error (RMSE), and R-squared ($R^2$). The \kpi-Win size (\kws) is a bounded variable determined by the formulation provided in Equation~\ref{eq:window-size}. Since online settings assume a potentially infinite stream of data samples, $N$ represents the number of the most recent data points in the rolling window, and $K$ represents the mini-batch size. The scaling factor $\gamma$ depends on the specific requirements; in our experiments, $\gamma = 0.05$ was used. The \kws is constrained between a lower bound (LB) and an upper bound (UB). Values below LB are set to LB, and values above UB are capped at UB.

\begin{figure}[ht]
  \centering
\includegraphics[width=0.75\textwidth, keepaspectratio]{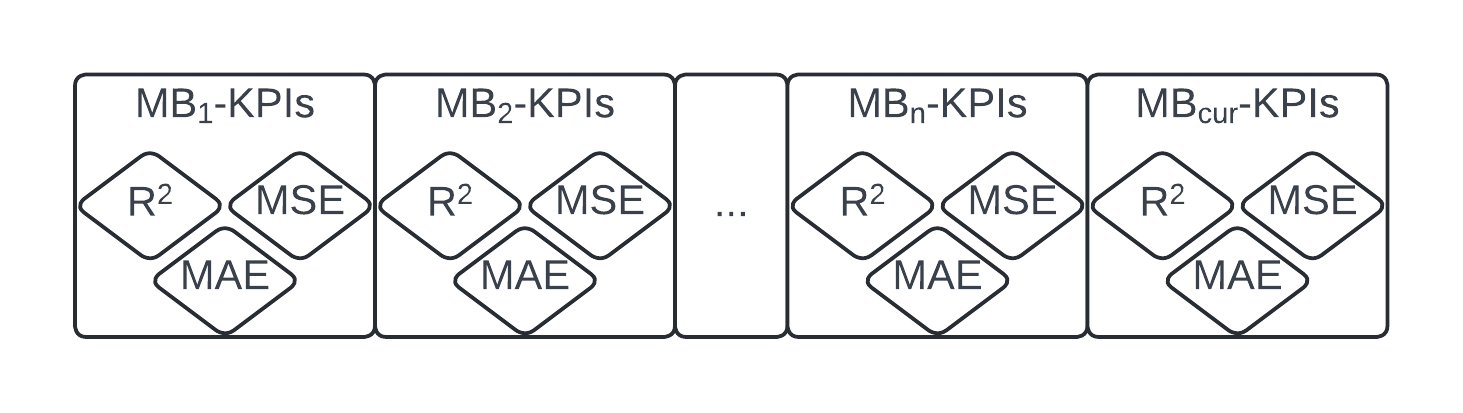}
  \vspace{-.3cm}
  \caption{\centering \kpis Window (\kpi-Win)}
  \label{fig:002_KPIs_Window}
\end{figure}

In an illustrative scenario featuring 31 entries in \kpi-Win, where each entry encompasses a set of \kpis related to the consumed mini-batch, the initial 30 entries are allocated for historical data. The final entry is designated for the instant feed or the current mini-batch, denoted as $\text{Inst}_{\scriptscriptstyle\text{KPI}}$. The $\text{Inst}_{\scriptscriptstyle\text{KPI}}$ entry is transient and dynamically updates in response to concurrent computations, while the remaining entries represent our \textit{Baseline Statistics}, serving as a point of reference for comparing performance changes. Figure~\ref{fig:002_KPIs_Window} illustrates the data structure used within \modelname, where each entry represents a bag comprising one or more \kpis.

This point-level and mini-batch-level monitoring also enables \modelname{} to respond to local drift. If a locally drifted group of samples appears in the stream and causes a measurable degradation in the monitored \kpi, the transient $\text{Inst}_{\scriptscriptstyle\text{KPI}}$ entry will deviate from the recent \kpi-window baseline. In this case, \modelname{} can classify the deviation as incremental or abrupt drift according to its severity and then activate the corresponding hyperparameter adaptation and recalibration mechanisms. Thus, \modelname{} handles local drift through its effect on predictive performance, although it does not explicitly localize the affected feature-space region.

\begin{equation}
    \label{eq:window-size}
    \begin{aligned}
        &\text{LB} \leq \kws
        &\quad= \left(\frac{N}{K}\right) \times \gamma \leq \text{UB}
    \end{aligned}
\end{equation}

\modelname estimates the recent distributional behavior of the historical \kpis in the current \kpi-window, using their empirical mean and standard deviation to compute a variable threshold $\tau$. This approach avoids the use of a fixed threshold, which may not be suitable for all \kpis or data streams due to their varying characteristics and dynamics. The variable threshold can be computed using Equation~\ref{eq:threshold}, where $\sigma$ denotes the standard deviation of the recent \kpi-window and $z$ is a sensitivity multiplier derived from the user-specified nominal level \(\rho\).

{
\begin{equation}
\label{eq:threshold}
\text{Threshold } (\tau) = z \times \sigma
\end{equation}
}
{
\begin{equation}
    \label{eq:z}
    z = \Phi^{-1}(1-\rho)
\end{equation}
}

The method establishes two thresholds, denoted as the “low” and “high” limits, and selects the appropriate one based on the nature of the employed \kpi. For \kpis where smaller values are preferred (e.g., loss), the relevant threshold corresponds to the right tail of the distribution. Conversely, for \kpis where larger values are desirable (e.g., $R^2$), the “low” limit aligned with the left tail is applied. For example, when the \kpi is $R^2$, a deviation into the left tail indicates performance degradation, suggesting potential drift that warrants further investigation. In scenarios where multiple \kpis are considered and their objectives differ (for instance, loss decreases with improvement while $R^2$ increases), both the “low” and “high” limits are jointly applied to maintain reliable drift detection.

The two limits, “low” and “high,” are determined from the dynamically computed threshold $\tau$, as defined in Equation~\ref{eq:low-high}, where  $\mu_{\scriptscriptstyle\text{KPI}}$ is the mean of the \kpi window. The role of $\tau$ is to provide an adaptive boundary that accounts for the characteristics of the employed \kpi as well as the variability of the incoming data stream. Unlike fixed thresholds, which may be either too strict or too lenient under changing conditions, a dynamic threshold adjusts to the evolving distribution of \kpi values. Additionally, $\tau$ is dynamically derived and is determined by $z$, which reflects the user-specified nominal per-decision sensitivity level under the local \kpi-window approximation. This supports robustness across different types of \kpis and data regimes by calibrating the detection boundary to recent \kpi variability and reducing reactions to benign short-term fluctuations.

\begin{equation}
\label{eq:low-high}
\begin{aligned}
\text{low} & = \mu_{\scriptscriptstyle\text{KPI}} - \tau, \\
\text{high} & = \mu_{\scriptscriptstyle\text{KPI}} + \tau.
\end{aligned}
\end{equation}

To distinguish between stable performance, incremental drift, and abrupt drift, three boundaries are defined. The innermost region, referred to as the safe band, is given by $[\mu_{\scriptscriptstyle\text{KPI}} - \zeta,\, \mu_{\scriptscriptstyle\text{KPI}} + \zeta]$. The parameter $\zeta$ is a user-defined tolerance level that determines the width of this safe region. By design, $\zeta$ reflects the user’s preference regarding sensitivity to small deviations. A smaller value of $\zeta$ yields a narrower safe band, making the detector more responsive to subtle variations, whereas a larger value creates a wider buffer that suppresses frequent, unnecessary adaptations. In our experiments, $\zeta$ was set to 0.005, which provided a practical balance between sensitivity and stability.

\begin{figure}[ht]
    \centering
    \begin{minipage}{\columnwidth}
        \centering        \includegraphics[width=\columnwidth,keepaspectratio]{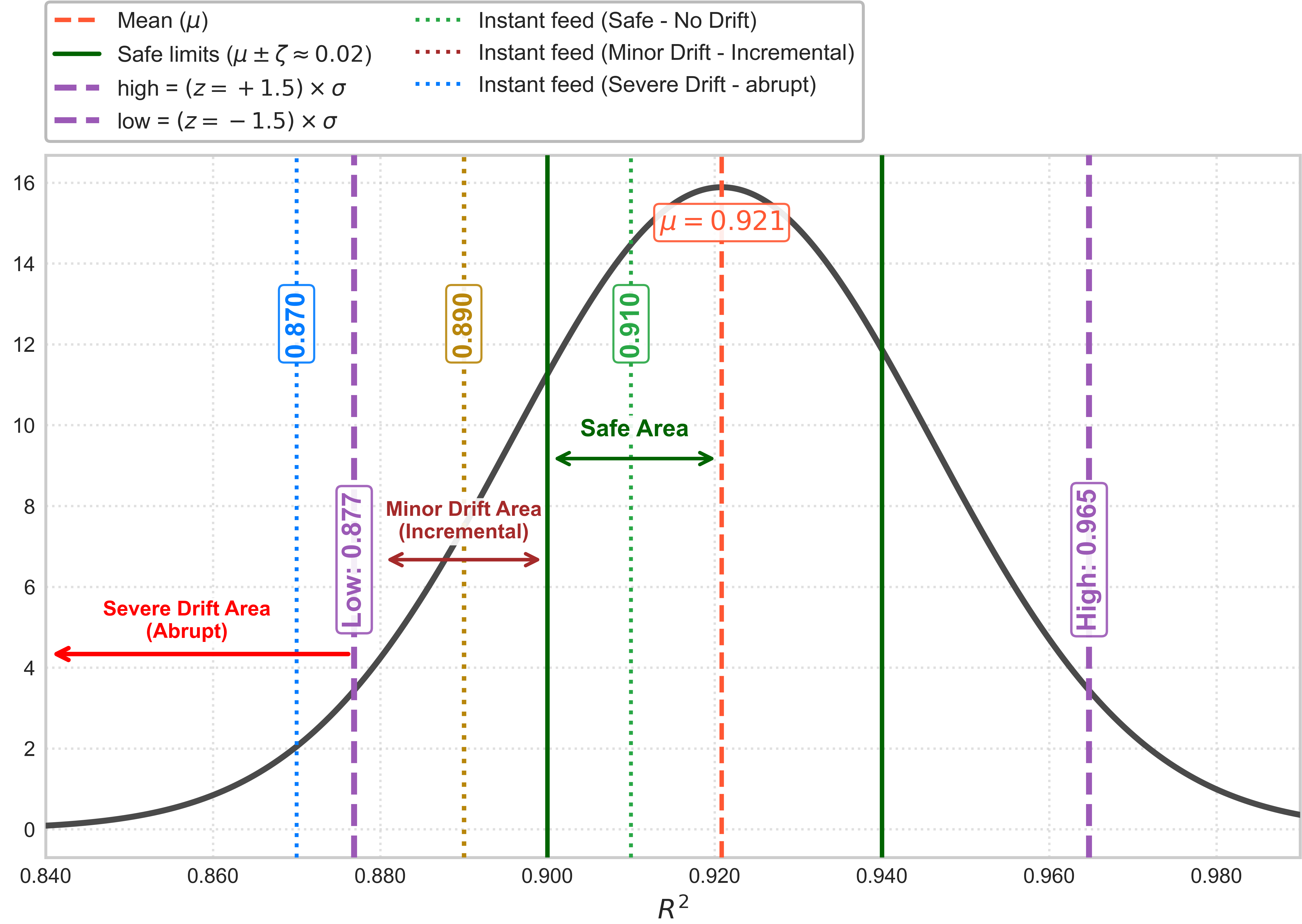}
        \caption{\centering \modelname Limits - Gauss Dist. of \kpi-Win on \kpi = $R^2$}
        \label{fig:003_SCCM_LIMITS}
    \end{minipage}
\end{figure}

For each incoming instance (either a mini-batch or a single data point), the drift magnitude is computed to determine its position within the three regions: the safe band, the incremental drift zone, or the abrupt drift zone. The drift magnitude (DM) is computed using Equation \ref{eq:drift_magnitude_lstm_sccm}, where $\text{Inst}_{\scriptscriptstyle\text{KPI}}$
denotes the instant feed's \kpi under study. Observations falling within the safe band are interpreted as stable performance. Instances that fall outside the safe band but remain within the outer limits (the low and high thresholds) are considered indicative of incremental drift. Finally, observations that exceed the outer limits are flagged as abrupt drift. This hierarchical boundary structure enables \modelname to distinguish between minor, incremental, and severe deviations in a principled way. Figure~\ref{fig:003_SCCM_LIMITS} illustrates this boundary construction.
{
\begin{equation}        
    \label{eq:drift_magnitude_lstm_sccm}
    \text{Drift Magnitude (DM)} 
    = \left| \text{\Large $\mu$}_{\scriptscriptstyle\text{KPI}} 
    - \text{Inst}_{\scriptscriptstyle\text{KPI}} \right|
\end{equation}
}

\begin{figure}[ht]
  \centering
  \includegraphics[width=0.9\textwidth, keepaspectratio]{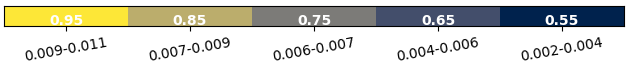}
  \caption{\centering \modelname Scale Map}
  \label{fig:004_SCCM_SCALE}
\end{figure}
False positives arise when drift is signaled even though the monitored \kpi remains consistent with the recent baseline behavior represented in the current \kpi-window. In online learning, this is particularly problematic because each false alarm can trigger unnecessary adaptations, retuning, or even model recalibrations. Such actions destabilize training, increase computational cost, and risk degrading performance by chasing noise instead of meaningful signal \cite{xu2024addressing, poenaru2022concept}. At a finer level, an abrupt false positive occurs when a benign fluctuation crosses the one-sided low/high limit on the monitored tail, while an incremental false positive arises when a benign point lands between the safe boundary and that limit. Both cases may prompt adaptation even though the underlying concept has not meaningfully changed. Repeated false positives can lead to parameter oscillations, loss of calibration for hyperparameters such as the adaptation weight, unnecessary forgetting of useful prior knowledge, and ``alert fatigue'' for operators \cite{spathoulas2010reducing}.

\begin{algorithm}
\caption{\modelname - Stream Cruise Control Method}
\label{alg:sccm}
\begin{algorithmic}[1]

\State \textbf{Input:} 
$\mathbf{W}$, 
$\alpha$,
$(\mathbf{X}_{\text{inc}}, \mathbf{y}_{\text{inc}}) \subseteq \{\mathbf{X}, \mathbf{y}\}$,
$\textit{kpi} \subseteq \{\text{Acc}, \text{Loss}, \ldots\}$,

\Statex \hspace{\algorithmicindent}
\hspace{20pt}$\rho \in (0,1]$
\hfill
\makebox[.47\linewidth][l]{%
\Comment{\footnotesize \textcolor{blue}{Nominal per-decision sensitivity level}}%
}

\State Initialize an empty list: $L \gets [\hspace{2pt}]$ \Comment{\footnotesize \parbox[t]{.35\linewidth}{\textcolor{blue}{$L$ is the KPI-Win}}}

\State $\mathbf{MB}_{\text{kpi}} \gets \method{MbKPI}(\mathbf{X}_{\text{inc}}, \mathbf{y}_{\text{inc}}, \mathbf{W}, \textit{kpi})$\Comment{\footnotesize \parbox[t]{.35\linewidth}{\textcolor{blue}{MB is either a mini-batch or a single point}}}

\State \method{AddItem}($L$, $\mathbf{MB}_{\text{kpi}}$)

\State $ \tau, \mu, \sigma, \text{low}, \text{high}, \text{DM} \gets \method{MeasureKPIs}(L, z)$ 
\Comment{\footnotesize\parbox[t]{.35\linewidth}{
\textcolor{blue}{with $z \gets \Phi^{-1}(1-\rho)$, where $\rho$
is the nominal local per-decision sensitivity level}}}

\State $\textit{drift} \gets \method{DetectDrift}(\mu, L[-1], \tau)$ 

\If{\textit{drift}}

    \State \method{RemoveItem}($L$, $-1$)

    \State $\text{SM} \gets \method{DefineScaleMap}(\mu, \text{low}, \text{high})$       
        \Comment{\footnotesize\parbox[t]{.35\linewidth}{\textcolor{blue}{SM is the scaled map}}}
    
     \State $\alpha' \gets \method{TuneHyperparams}(\text{SM}, \text{DM})$

    \State $\mathbf{W} \gets \method{UpdateModelHyperparam}(\alpha')$

    \State $\mathbf{MB}_{\text{kpi}} \gets \method{MbKPI}(\mathbf{X}_{\text{inc}}, \mathbf{y}_{\text{inc}}, \mathbf{W}, \textit{kpi})$

    \State \method{AddItem}($L$, $\mathbf{MB}_{\text{kpi}}$)
    
    \State $ \tau, \mu, \sigma, \text{low}, \text{high}, \text{DM} \gets \method{MeasureKPIs}(L, z)$ 

    \State $\textit{abrupt-drift} \gets \method{DetectDrift}(\mu, L[-1], \tau)$ 

    \State counter $\gets$ 0
        
    \While {\textit{abrupt-drift} and $counter < max$}
        \State counter $\gets$ counter + 1
        \State \method{RemoveItem}($L$, $-1$)
        \State $\text{X}_{\text{inc}}, \text{y}_{\text{inc}} \gets \text{X}'_{\text{inc}}, \text{y}'_{\text{inc}}$ \Comment{\footnotesize\parbox[t]{.35\linewidth}{\textcolor{blue}{$\text{X}'_{\text{inc}}, \text{y}'_{\text{inc}}$ denotes new data points}}}

        \State $\mathbf{W} \gets$  \method{CalibrateModel}(
        $\alpha'$, 
        $\text{X}_{\text{inc}}$,
        $\text{y}_{\text{inc}}$, 
        $\mathbf{W}$)

        \State $\mathbf{MB}_{\text{kpi}} \gets \method{MbKPI}(\mathbf{X}_{\text{inc}}, \mathbf{y}_{\text{inc}}, \mathbf{W}, \textit{kpi})$

        \State \method{AddItem}($L$, $\mathbf{MB}_{\text{kpi}}$)

        \State $ \tau, \mu, \sigma, \text{low}, \text{high}, \text{DM} \gets \method{MeasureKPIs}(L, z)$ 

        \State $\textit{abrupt-drift} \gets \method{DetectDrift}(\mu, L[-1], \tau)$ 
    
    \EndWhile                
\EndIf
\State \textbf{return} $\mathbf{W}$
\end{algorithmic}
\end{algorithm}
To mitigate false positives, \modelname adopts a CFAR-inspired thresholding mechanism in which the user-defined parameter $\rho$ controls the nominal per-decision sensitivity of the detector \cite{scharf1991statistical}. In classical CFAR detection, thresholds are calibrated from a reference distribution of the monitored statistic, and constant false-alarm behavior is obtained under the assumptions associated with that reference model. In online learning, however, the full data stream is generally non-stationary, and therefore \modelname does not claim a global constant false-alarm guarantee over the entire stream. Instead, \modelname applies this principle locally to the \kpi-window, which summarizes recent mini-batch predictive behavior under the current baseline. The resulting threshold is therefore interpreted as a local, window-conditioned calibration of detection sensitivity rather than as a long-term bound on cumulative false alarms. The safe band $\zeta$ further absorbs benign fluctuations near the baseline, while the outer limits distinguish stable behavior from incremental and abrupt deviations.

In \modelname, this calibration is operationalized by computing $z$ from a user-specified nominal level $\rho$ via Equation~\ref{eq:z}, where $\Phi$ denotes the cumulative distribution function (CDF) of the standard normal distribution. Under the local \kpi-window approximation, $\rho$ defines a nominal per-decision level for how sensitive the monitored threshold is to benign \kpi fluctuations. Smaller values of $z$ (corresponding to higher $\rho$) make the detector more sensitive to subtle changes, while larger values of $z$ (corresponding to lower $\rho$) produce more conservative thresholds that reduce reactions to minor fluctuations.

To further illustrate this relationship, users can directly control the sensitivity of the detector through their choice of $\rho$. For example, setting $\rho = 0.06$ corresponds to a lower threshold of $z \approx 1.55$, meaning that the detector reacts to smaller \kpi deviations, thereby increasing sensitivity but also increasing the likelihood of reacting to benign fluctuations. In contrast, choosing a smaller value such as $\rho = 0.006$ yields a higher threshold of $z \approx 2.51$, which makes the detector more conservative by ignoring minor fluctuations and responding only to more substantial deviations. Thus, $\rho$ and $z$ are inversely related: increasing $\rho$ lowers $z$ and heightens sensitivity, while decreasing $\rho$ raises $z$ and makes drift detection more conservative. This tunability allows practitioners to align the local detection sensitivity with their tolerance for spurious alarms and the importance of capturing subtle drift in their application.


An illustrative example helps clarify the calibration. Suppose the monitored
\kpi is $R^2$, with
$\mu_{\scriptscriptstyle\mathrm{KPI}} = 0.921$,
$\sigma = 0.010$, $\zeta = 0.020$, and $\rho = 0.01$. This yields
$z = \Phi^{-1}(0.99) \approx 2.326$ and
$\tau = z\sigma \approx 0.0233$. The corresponding safe band is
$[\mu_{\scriptscriptstyle\mathrm{KPI}}-\zeta,\,
\mu_{\scriptscriptstyle\mathrm{KPI}}+\zeta]
= [0.901,\,0.941]$. Because $R^2$ is a higher-is-better \kpi, the monitored
lower limit is
$\text{low}
= \mu_{\scriptscriptstyle\mathrm{KPI}}-\tau
\approx 0.8977$. Accordingly, a reading of $0.910$ lies within the safe band
and is classified as stable. A reading of $0.900$ lies between the lower
safe-band boundary and the monitored lower limit and is classified as
incremental drift. Readings of $0.890$ and $0.870$ fall below the monitored
lower limit and are classified as abrupt drift. Conversely, a reading of
$0.950$, which lies above the safe band on the non-monitored side, is treated
as an improvement rather than a drift trigger. Under the local \kpi-window
approximation, $\rho = 0.01$ represents the nominal per-decision sensitivity
level on the monitored tail.

While $\rho$ regulates the nominal per-decision sensitivity of the detector, it is not intended as a global stream-level false-alarm guarantee. Since online monitoring involves repeated detection decisions over time, the cumulative probability of observing at least one false alarm can increase with the length of the stream. Therefore, $\rho$ is interpreted in \modelname as a local sensitivity-control parameter under the current \kpi-window approximation.

When drift is detected, whether incremental or abrupt, the hyperparameter optimization mechanism is activated. The specific hyperparameters and their ranges vary across models. For instance, in the OLR-WA model, $\alpha \in (0,1]$, where values closer to 1 place greater emphasis on newly arriving data, while values around 0.5 balance past information with new observations. In contrast, the RLS model employs a forgetting factor $\lambda \in (0,1]$; here, $\lambda = 1$ corresponds to equal weighting of all data, whereas values approaching 0 assign greater weight to recent inputs. A careful understanding of these ranges is essential, as different bounds govern the trade-off between rapid adaptation to drift and long-term stability.

The hyperparameter optimization process employs a dynamic, on-the-fly scaled map to identify the most suitable hyperparameter value, which is then applied to the target model. This scaled map partitions the interval between the safe-band boundary and the outer low/high limit into subregions. Each subregion corresponds to a key, and its associated value is the tuned hyperparameter selected from the model’s predefined range $\alpha \in (0,1]$. For illustration, consider the case of $\alpha$: higher values (e.g., close to 1) represent stronger adaptation and are assigned to regions nearer the outer low/high limits, whereas moderate values (e.g., around 0.5) reflect minor deviations within the incremental zone and correspond to points closer to the safe-band boundary. This design maintains a balance between adaptability and stability, allowing the model to adjust effectively without overfitting.

The scaled map guides the optimization mechanism by assigning hyperparameter values in proportion to the degree of deviation detected in the incoming feed. The process is analogous to automotive cruise control, where the level of fuel injection or braking is continuously adjusted to maintain stable speed under varying conditions. Similarly, the objective here is to preserve stable model performance when drifts are encountered. As illustrated in Figure~\ref{fig:004_SCCM_SCALE} for OLR-WA with $\alpha \in (0,1]$, the leftmost yellow region corresponds to the most pronounced deviations, where larger values of $\alpha$ are selected to emphasize rapid adaptation to new data. In contrast, the rightmost blue region reflects minor deviations, where more moderate values of $\alpha$ are applied to preserve stability. This design achieves a balanced trade-off between adaptability and stability, which is critical in online learning, preventing overfitting while ensuring responsiveness to evolving data streams.

After the hyperparameter optimization process has been applied and the model has been updated with the optimized parameters, \modelname re-evaluates the \kpi to determine whether drift still persists. If the new instance remains beyond the low or high limits even after optimization, the system interprets this as an indication of abrupt drift. In such cases, the model incrementally invokes additional data points from the incoming batch within a rolling window and performs calibration using both the newly introduced data and the already optimized hyperparameters. The number of additional data points is constrained by a predefined maximum limit to avoid unnecessary retraining. In our experiments, incorporating only a small number of instances (typically 3 to 5) proved sufficient to recalibrate the model and restore stable performance.

To further emphasize the algorithmic novelty of \modelname{}, 
Table~\ref{tab:sccm-algorithmic-comparison} compares its internal mechanisms with representative concept drift detection and adaptation methods. 
The comparison focuses on how each method detects change, stores or uses recent information, defines its decision threshold, estimates drift magnitude, classifies drift severity, controls hyperparameters, and adapts the model. 
Unlike methods that only monitor classification errors, compare data distributions, maintain adaptive windows, or apply adaptation after an external drift signal, \modelname{} combines KPI-based memory, CFAR-inspired local thresholding, drift magnitude estimation, severity-aware classification, scale-map hyperparameter control, and bounded recalibration in one pre-update online learning pipeline. 

As shown in Table~\ref{tab:sccm-algorithmic-comparison}, the algorithmic distinction of \modelname{} lies in how the drift signal is produced, interpreted, and converted into adaptive control. 
Existing detectors typically output a binary drift/no-drift decision or warning/drift signal, while adaptation methods such as RESET, WINDOW, OHL, and SSPT usually require an external trigger. 
In contrast, \modelname{} uses recent KPI behavior as a bounded memory-based control signal, estimates the magnitude of deviation, classifies the severity of drift, and maps this information to proportional hyperparameter tuning and bounded recalibration.
Therefore, \modelname{} is not simply a combination of existing drift-handling components, but a unified control-oriented mechanism for online regression under concept drift. 

In the column ``Memory / Window Stores,'' KPIs indicate that the method stores or directly monitors recent predictive performance indicators, while Data points indicate that the method stores recent samples, stream values, statistics, or subwindows rather than an explicit KPI memory. 
The signal stored by each method is shown next to the corresponding label.
In the column ``Drift Magnitude,'' \(\checkmark\) indicates that the method explicitly estimates the amount of drift, Indirect indicates that change size may be reflected through distribution distances, learner scores, weights, or adaptation behavior without being used as a formal drift-magnitude control signal, and \(\times\) indicates that no drift-magnitude estimate is produced.
In the column ``Drift Severity Classification,'' Warning/drift indicates that the method provides alert levels rather than severity categories, Binary indicates a drift/no-drift decision, Indirect indicates that severity may be reflected through learner weights, model behavior, or adaptation intensity without explicit severity labels, and \ indicates that no severity classification is provided.
In the column ``Hyperparameter Control,'' \(\checkmark\) indicates that the method explicitly tunes or controls hyperparameters, Indirect indicates that hyperparameter-related behavior may be influenced through learner weighting, model selection, or adaptation behavior without an explicit hyperparameter-control mechanism, and \(\times\) indicates that no hyperparameter control is performed.
In the column ``Model Adaptation,'' \(\checkmark\) indicates that the method includes model adaptation as part of its mechanism, Partial indicates that adaptation may occur indirectly or in a limited framework-specific manner, and \(\times\) indicates that the method does not adapt the predictive model by itself.

\begin{sidewaystable}[!htbp]
\centering
\captionsetup{justification=centering}
\caption{Algorithmic-level comparison between \modelname{} and representative concept drift detection and adaptation methods.}
\label{tab:sccm-algorithmic-comparison}

\scriptsize
\setlength{\tabcolsep}{3.2pt}
\renewcommand{\arraystretch}{1.25}

\resizebox{\textheight}{!}{%
\begin{tabular}{l p{3.8cm} p{3.0cm} p{2.8cm} c c c c p{5.8cm}}
\toprule
\textbf{Method} &
\rotatebox{30}{\makecell{\textbf{Main Algorithmic}\\\textbf{Mechanism}}} &
\rotatebox{30}{\makecell{\textbf{Memory / Window}\\\textbf{Stores}}} &
\rotatebox{30}{\makecell{\textbf{Decision / Threshold}\\\textbf{Mechanism}}} &
\rotatebox{30}{\makecell{\textbf{Drift}\\\textbf{Magnitude}}} &
\rotatebox{30}{\makecell{\textbf{Drift Severity}\\\textbf{Classification}}} &
\rotatebox{30}{\makecell{\textbf{Hyperparameter}\\\textbf{Control}}} &
\rotatebox{30}{\makecell{\textbf{Model}\\\textbf{Adaptation}}} &
\makecell{\textbf{Key Difference}\\\textbf{from \modelname{}}} \\
\midrule

\textbf{DDM} &
Monitors classification error rate and standard deviation &
KPIs: classification error rate &
Fixed/statistical warning-drift levels &
\xmark &
Warning/drift &
\xmark &
\xmark &
Classification-oriented detector; uses error-based monitoring but does not provide magnitude-aware tuning or proportional adaptation. \\

\textbf{EDDM} &
Monitors the distance between classification errors &
KPIs: distance between classification errors &
Fixed/statistical thresholds &
\xmark &
Warning/drift &
\xmark &
\xmark &
Improves gradual drift detection but remains classification-oriented and does not quantify drift magnitude for adaptive control. \\

\textbf{ADWIN} &
Compares statistics over adaptive subwindows &
Data points: adaptive subwindows &
Adaptive statistical test &
\xmark &
Binary &
\xmark &
\xmark &
Uses adaptive windows for change detection, but does not inherently quantify drift severity, tune hyperparameters, or adapt the model by itself. \\

\textbf{KSWIN} &
Applies the Kolmogorov--Smirnov test between recent and reference windows &
Data points: recent/reference windows &
Significance-test based &
\xmark &
Binary &
\xmark &
\xmark &
Detects distributional change using statistical windows, but has no built-in adaptation or magnitude-aware control policy. \\

\textbf{ITA} &
Uses information-theoretic distribution comparison, such as KL divergence &
Data points: distribution statistics &
Distribution-comparison based &
Indirect &
\xmark &
\xmark &
\xmark &
Measures distributional change, but does not map the measured change into model adaptation or hyperparameter control. \\

\textbf{PL} &
Compares stable and reactive learners &
KPIs: learner performance &
Performance-comparison based &
\xmark &
Binary &
\xmark &
\pmark &
Requires maintaining multiple learners; adaptation is comparison-driven rather than based on explicit drift magnitude control. \\

\textbf{\makecell[l]{Ensemble\\Methods}} &
Updates, reweights, or replaces learners based on their performance &
KPIs: learner performance scores &
Implicit &
Indirect &
Indirect &
Indirect &
\cmark &
Adaptation occurs through multiple learners, but usually requires higher memory and inference cost than \modelname{}. \\

\textbf{RESET} &
Discards the current model after drift is detected &
N.A. &
\xmark &
\xmark &
\xmark &
\xmark &
\cmark &
A reaction strategy only; it requires an external detector and provides no detection, magnitude estimation, or tuning logic. \\

\textbf{WINDOW} &
Retrains or recalibrates the model using a recent data window &
Data points: recent training samples &
\xmark &
\xmark &
\xmark &
\xmark &
\cmark &
Uses recent data for recalibration, but does not independently detect drift severity or perform adaptive hyperparameter tuning. \\

\textbf{OHL} &
Uses online hypergradient-based hyperparameter updates &
N.A. &
\xmark &
\xmark &
\xmark &
\cmark &
\cmark &
Tunes hyperparameters online, but typically requires an external drift signal or objective and does not classify drift severity. \\

\textbf{SSPT} &
Searches hyperparameter configurations in-stream &
KPIs: validation/performance feedback &
\xmark &
\xmark &
\xmark &
\cmark &
\cmark &
Performs stream-based tuning, but does not provide its own drift detection, severity classification, or magnitude-aware control. \\

\textbf{HyperTube} &
Maintains parallel online learners with adaptive hyperparameter tuning &
KPIs: learner performance feedback &
Implicit &
Indirect &
Indirect &
\cmark &
\cmark &
Uses multiple learners and tuning mechanisms, but is more resource-intensive and not centered on lightweight KPI-based control. \\

\textbf{AEF-CDA} &
Uses an adaptive ensemble framework for concept drift adaptation &
KPIs: ensemble/learner performance &
Implicit &
Indirect &
Indirect &
Indirect &
\cmark &
Ensemble-oriented and often domain-specific; \modelname{} is lightweight, model-agnostic, and explicitly magnitude-aware. \\

\textbf{StepWise} &
Uses automatic anomaly or drift thresholding with iSST-EVT &
Data points: anomaly/drift statistics &
Automatic thresholding &
\xmark &
\xmark &
\xmark &
\pmark &
Designed mainly as a domain-specific anomaly or drift framework, not as a regression-oriented adaptive control pipeline. \\

\textbf{\modelname{}} &
KPI-Win, CFAR-inspired local thresholding, drift magnitude estimation, scale-map tuning, and bounded recalibration &
\makecell{Any KPI\\handled by dynamic\\threshold \(\tau\)} &
CFAR-inspired local KPI threshold &
\cmark &
\makecell{Stable\\incremental\\abrupt} &
\cmark &
\cmark &
Unified, magnitude-aware, early-response adaptation pipeline for online regression. \\

\midrule
\multicolumn{9}{@{}l@{}}{\footnotesize \cmark: supported; \xmark: not supported; \pmark: partially supported.} \\

\bottomrule
\end{tabular}%
}
\end{sidewaystable}

\section{Experiments}

This section presents a comprehensive evaluation of the proposed \modelname framework under diverse non-stationary environments. The experiments are conducted on both \textit{synthetic datasets} and \textit{real-world datasets} to assess the robustness, generality, and practical applicability of the proposed approach.

\subsection{Datasets}

We evaluate the proposed method using a comprehensive set of synthetic and real-world datasets. The synthetic datasets are designed to provide controlled experimental settings in which different types, magnitudes, and dimensionalities of concept drift can be systematically analyzed. In contrast, the real-world datasets are used to assess the method under practical conditions where data distributions may contain natural variability, noise, and hidden non-stationarity. Together, these datasets allow us to evaluate the robustness, adaptability, and predictive performance of the proposed method across both controlled and realistic streaming regression scenarios.

\subsubsection{Synthetic Datasets}

We construct a suite of synthetic datasets to simulate three major types of concept drift: abrupt, incremental, and alternating gradual drift. All synthetic datasets are generated using a unified parametric framework in which input features are sampled from Gaussian distributions, and target values are produced using linear regression functions with additive Gaussian noise. Specifically, each concept is defined by an input distribution mean, a regression coefficient vector, a bias term, and a fixed noise level.

Drift is introduced by modifying one or more parameters of the data-generating process across concepts. These parameters include the input distribution mean, the regression coefficients, and the bias term. This design enables controlled manipulation of both covariate shift and changes in the underlying input-output relationship. To provide broad coverage, the synthetic datasets vary across drift type, drift magnitude, dimensionality, and number of data points.

Abrupt drift is modeled as an instantaneous transition from one concept to
another at a predefined drift point. Incremental drift is modeled by smoothly
interpolating the data-generating parameters between an initial and a final
concept over multiple intermediate steps. The GDS streams are modeled as
\textit{alternating gradual drift}: the stream alternates between the old
concept $C_1$ and the new concept $C_2$ across consecutive segments during the
transition period before stabilizing under $C_2$. This construction represents
the temporary coexistence of the old and new concepts associated with gradual
drift \cite{gama2014survey}.

The synthetic datasets include both low-dimensional and high-dimensional settings. For all synthetic datasets, Gaussian noise with a standard deviation of 1.5 is added to the target variable. Table~\ref{tab:synthetic-datasets-properties} summarizes the main statistical and drift-related properties of the synthetic datasets, while Table~\ref{tab:synthetic-concept-definitions} provides the corresponding concept definitions and data-generating functions.

The drift indicators in Table~\ref{tab:synthetic-datasets-properties} confirm that the synthetic benchmark contains statistically detectable and diverse distributional changes. For abrupt drift, all datasets show extremely small target-space KS p-values, indicating significant changes in the target distribution after the drift point. The drift magnitude varies across datasets: ADS04 represents a milder abrupt case, with a target JS divergence of 0.035 and Wasserstein distance of 1.766, whereas ADS05 shows a stronger abrupt shift, with a JS divergence of 0.360 and Wasserstein distance of 7.040. A similar variation appears in the incremental datasets, where IDS01 shows a smaller transition with a Wasserstein distance of 1.300, while IDS05 exhibits the strongest incremental target shift with a Wasserstein distance of 7.983. For alternating gradual drift, GDS04 provides the mildest target-space change, with a JS divergence of 0.0228 and Wasserstein distance of 1.2103, whereas GDS02 and GDS05 represent stronger alternating gradual drift cases, with Wasserstein distances of 4.3848 and 6.5352, respectively. In the input space, all datasets show significant feature-level changes, as indicated by the X-space KS significance results. Overall, these values demonstrate that the synthetic datasets cover both mild and strong drift scenarios across abrupt, incremental, and alternating gradual drift types.

\subsubsection{Real-World Datasets} \label{subsubsec:realworlddatasets}

We further evaluate the proposed method using eight real-world regression datasets, as summarized in Table~\ref{tab:real_datasets-properties}. These datasets cover diverse application domains, including healthcare, business and finance, housing and real estate, energy systems, environmental monitoring, chemical sensing, oceanography, and retail sales forecasting. Their target variables include insurance cost, company profit, house price, net hourly power output, carbon monoxide concentration, gas concentration, ocean salinity, and weekly sales, respectively.

Unlike the synthetic datasets, where the drift type, drift location, magnitude, and noise level are explicitly controlled, the real-world datasets are used in their naturally occurring chronological form. This reflects practical machine learning conditions in which distributional changes may be gradual, irregular, multidimensional, and not associated with labeled concept boundaries. The real-world datasets therefore provide a complementary evaluation setting for assessing whether the proposed method can maintain stable predictive performance under naturally occurring variability and uncertainty.

For each real-world dataset, empirical drift was quantified by comparing the first and second halves of the chronological stream. Since these datasets do not provide ground-truth concept boundaries or known data-generating parameters, the reported values are interpreted as empirical indicators of non-stationarity rather than controlled drift magnitudes. The analysis considers changes in the target distribution, changes across the input features, and differences between regression models fitted to the two stream segments. Because statistical significance can be strongly influenced by dataset size, particularly for CalCOFI and WSSF, the KS test results are interpreted jointly with the Jensen--Shannon divergence, Wasserstein distance, and fitted-parameter distance.

The resulting datasets provide a comprehensive range of real-world drift conditions:

\begin{itemize}
    \item \textbf{MCPD}~\cite{MCPD_DS} exhibits very limited drift.
    Its target KS p-value is high ($p=0.769256$), while the average
    feature Jensen--Shannon divergence is small ($0.004202$).
    The fitted-parameter distance is also limited ($0.054078$),
    indicating that the relationship between the input variables
    and insurance cost remains relatively stable.

    \item \textbf{1KC}~\cite{1KC_DS} shows little marginal
    distributional variation. Its target KS p-value is high
    ($p=0.952778$), and the minimum feature-level KS p-value is
    $0.890432$, indicating no statistically significant feature
    drift. However, its relatively large fitted-parameter distance
    ($0.739862$) suggests a possible change in the relationship
    between the operating variables and company profit, despite
    the stability of their marginal distributions.

    \item \textbf{KCHSD}~\cite{KCHS_DS} exhibits moderate
    non-stationarity. The target distribution changes significantly
    ($p=0.000519$), and at least one feature shows strong evidence
    of drift, with a minimum KS p-value of
    $1.04{\times}10^{-24}$. Nevertheless, the small target
    Jensen--Shannon divergence ($0.001960$) suggests that the
    practical magnitude of the target-distribution change remains
    moderate.

    \item \textbf{CCPP}~\cite{CCPP_DS} is largely stable, with only
    weak input-distribution variation. Its target KS p-value is high
    ($p=0.610529$), and its average feature Jensen--Shannon
    divergence is only $0.001811$. Although at least one feature
    changes significantly, the very small fitted-parameter distance
    ($0.003641$) indicates a highly stable relationship between the
    environmental variables and net hourly power output.

    \item \textbf{UCIAQD}~\cite{UCIAQD_DS} exhibits strong drift in
    both the input and target distributions. Its target KS p-value is
    $1.28{\times}10^{-19}$, while its normalized target Wasserstein
    distance and average feature Jensen--Shannon divergence are
    $0.290262$ and $0.147997$, respectively. These values indicate
    meaningful changes in both air-quality conditions and sensor
    measurements across the chronological stream.

    \item \textbf{GASD}~\cite{GASD_DS} demonstrates strong
    multidimensional drift. Its target Jensen--Shannon divergence
    is $0.088123$, its normalized target Wasserstein distance is
    $0.243632$, and its average feature Jensen--Shannon divergence
    is $0.067600$. Its fitted-parameter distance of $3.502364$
    further indicates substantial changes in the sensor-response
    relationships across the ten chronological batches.

    \item \textbf{CalCOFI}~\cite{CALCOFI_DS} primarily represents
    covariate-dominant drift. Its average feature Jensen--Shannon
    divergence is relatively large ($0.132505$), whereas the
    normalized Wasserstein distance for salinity is only $0.019585$.
    This suggests that the environmental and spatiotemporal input
    conditions vary more strongly than the target distribution
    itself.

    \item \textbf{WSSF}~\cite{WSSF_DS} represents a mild
    target-drift setting with noticeable feature variation. Although
    its target KS test is statistically significant
    ($p=3.63{\times}10^{-7}$), its target Jensen--Shannon divergence
    and normalized Wasserstein distance are only $0.000143$ and
    $0.000269$, respectively. In contrast, its average feature
    Jensen--Shannon divergence of $0.022639$ indicates changes in
    the retail, temporal, and economic input conditions.
\end{itemize}

Collectively, these datasets span largely stable streams, mild and moderate distributional changes, strong target and feature drift, covariate-dominant drift, and changes in the estimated input--output relationship. They also vary substantially in sample size and dimensionality, ranging from approximately one thousand observations to more than eight hundred thousand observations and from four to 145 input dimensions. This diversity strengthens the real-world evaluation by testing the proposed method under heterogeneous drift patterns, data scales, feature spaces, and application domains.

Overall, the combination of synthetic and real-world datasets provides a balanced and comprehensive evaluation framework. The synthetic datasets enable controlled analysis under known abrupt, incremental, and alternating gradual drift conditions, while the real-world datasets assess practical applicability under naturally occurring and unlabeled forms of non-stationarity.

\begin{table}[H]
\scriptsize
\centering
\captionsetup{justification=centering}
\setlength{\tabcolsep}{3pt}
\caption{Synthetic Datasets Properties
\vspace{-.1cm}
\begin{minipage}[t]{\textwidth}
\centering
\tiny
\textbf{Drift Types:} A = Abrupt, I = Incremental, G = Alternating Gradual. Synthetic drifts are injected by altering function parameters: abrupt (sudden change), incremental (smooth shift from one concept to the next), and alternating gradual (repeated old--new concept switches before final stabilization).\\
\textbf{Concept Loc.:} Points per concept (incremental drift recurs every specified interval);
\textbf{Y KS P-Value:} Kolmogorov--Smirnov test p-value;
\textbf{Y JS Div:} Jensen--Shannon divergence;
\textbf{Y Wasserstein Dist:} Normalized Wasserstein distance for $y$;
\textbf{X Any KS Sig:} True if any feature shows significant KS drift ($p < 0.05$);
\textbf{X Min KS P-Value:} Minimum KS p-value across features;
\textbf{X Avg JS Div:} Average JS divergence across features.
\end{minipage}
}
\label{tab:synthetic-datasets-properties}

\resizebox{\textwidth}{!}{%
\begin{tabular}{cccccccccccc}
\toprule
\makecell{Data\\sets} & \makecell{Drift\\Type} & \makecell{Data\\Points} & \makecell{Dimen-\\sions} & Noise & \makecell{Concept\\Loc.} & \makecell{Y KS\\P-Value}
& \makecell{Y JS\\Div} & \makecell{Y Wasserstein\\Dist} & \makecell{X Any\\KS Sig} & \makecell{X Min\\KS P-Value} & \makecell{X Avg\\JS Div} \\
\midrule

\textbf{ADS01} & A & 1k & 1 & 1.5 & 0.5k & $1.00\times10^{-15}$ & 0.098 & 1.960 & True & $1.80\times10^{-13}$ & 0.058 \\

\textbf{ADS02} & A & 1k & 1 & 1.5 & 0.5k & $1.17\times10^{-109}$ & 0.461 & 4.371 & T & $7.17\times10^{-42}$ & 0.175 \\

\textbf{ADS03} & A & 1k & 1 & 1.5 & 0.5k & $4.98\times10^{-35}$ & 0.176 & 3.127 & T & $3.70\times10^{-78}$ & 0.338 \\

\textbf{ADS04} & A & 2k & 10 & 1.5 & 1k & $2.32\times10^{-10}$ & 0.035 & 1.766 & T & $1.34\times10^{-8}$ & 0.019 \\

\textbf{ADS05} & A & 2k & 10 & 1.5 & 1k & $1.48\times10^{-156}$ & 0.360 & 7.040 & T & $5.98\times10^{-59}$ & 0.123 \\

\textbf{ADS06} & A & 2k & 10 & 1.5 & 1k & $8.66\times10^{-23}$ & 0.061 & 3.178 & T & $6.04\times10^{-166}$ & 0.335 \\
\\
\textbf{IDS01} & I & 1k & 1 & 1.5 & every 0.1k & $c_{i \rightarrow j}$: $4.12\times10^{-4}$ & $c_{i \rightarrow j}$: 0.201 & 1.300 & $c_{i \rightarrow j}$: T & $c_{i \rightarrow j}$: 0.036 & 0.152 \\

\textbf{IDS02} & I & 1k & 1 & 1.5 & every 0.1k & $c_{i \rightarrow j}$: $1.42\times10^{-19}$ & $c_{i \rightarrow j}$: 0.436 & 3.698 & $c_{i \rightarrow j}$: T & $c_{i \rightarrow j}$: $8.45\times10^{-11}$ & 0.308 \\

\textbf{IDS03} & I & 1k & 1 & 1.5 & every 0.1k & $c_{i \rightarrow j}$: $3.36\times10^{-13}$ & $c_{i \rightarrow j}$: 0.329 & 3.410 & $c_{i \rightarrow j}$: T & $c_{i \rightarrow j}$: $3.11\times10^{-20}$ & 0.447 \\

\textbf{IDS04} & I & 2k & 10 & 1.5 & every 0.2k & $c_{i \rightarrow j}$: $8.06\times10^{-7}$ & $c_{i \rightarrow j}$: 0.088 & 2.929 & $c_{i \rightarrow j}$: T & $c_{i \rightarrow j}$: $1.84\times10^{-4}$ & 0.071 \\

\textbf{IDS05} & I & 2k & 10 & 1.5 & every 0.2k & $c_{i \rightarrow j}$: $3.51\times10^{-38}$ & $c_{i \rightarrow j}$: 0.428 & 7.983 & $c_{i \rightarrow j}$: T & $c_{i \rightarrow j}$: $1.40\times10^{-16}$ & 0.172 \\

\textbf{IDS06} & I & 2k & 10 & 1.5 & every 0.2k & $c_{i \rightarrow j}$: $4.75\times10^{-5}$ & $c_{i \rightarrow j}$: 0.106 & 3.220 & $c_{i \rightarrow j}$: T & $c_{i \rightarrow j}$: $3.68\times10^{-40}$ & 0.384 \\
\\
\textbf{GDS01} & G & 1k & 1 & 1.5 & \makecell{[c1:300, c2:100, c1:100,\\ c2:100, c1:100, c2:300]} & $c_{i \rightarrow j}$: $3.36\times10^{-24}$ & $c_{i \rightarrow j}$: 0.1248 & 2.0272 & $c_{i \rightarrow j}$: T & $c_{i \rightarrow j}$: $1.95\times10^{-10}$ & 0.0660 \\

\textbf{GDS02} & G & 1k & 1 & 1.5 & \makecell{[c1:300, c2:100, c1:100,\\ c2:100, c1:100, c2:300]} & $c_{i \rightarrow j}$: $1.01\times10^{-104}$ & $c_{i \rightarrow j}$: 0.4759 & 4.3848 & $c_{i \rightarrow j}$: T & $c_{i \rightarrow j}$: $9.34\times10^{-36}$ & 0.1880 \\

\textbf{GDS03} & G & 1k & 1 & 1.5 & \makecell{[c1:300, c2:100, c1:100,\\ c2:100, c1:100, c2:300]} & $c_{i \rightarrow j}$: $9.34\times10^{-36}$ & $c_{i \rightarrow j}$: 0.1773 & 3.0699 & $c_{i \rightarrow j}$: T & $c_{i \rightarrow j}$: $9.83\times10^{-79}$ & 0.3407 \\

\textbf{GDS04} & G & 2k & 10 & 1.5 & \makecell{[c1:600, c2:200, c1:200,\\ c2:200, c1:200, c2:600]} & $c_{i \rightarrow j}$: 0.0110 & $c_{i \rightarrow j}$: 0.0228 & 1.2103 & $c_{i \rightarrow j}$: T & $c_{i \rightarrow j}$: 0.00116 & 0.0248 \\

\textbf{GDS05} & G & 2k & 10 & 1.5 & \makecell{[c1:600, c2:200, c1:200,\\ c2:200, c1:200, c2:600]} & $c_{i \rightarrow j}$: $9.33\times10^{-75}$ & $c_{i \rightarrow j}$: 0.3501 & 6.5352 & $c_{i \rightarrow j}$: T & $c_{i \rightarrow j}$: $3.30\times10^{-28}$ & 0.1144 \\

\textbf{GDS06} & G & 2k & 10 & 1.5 & \makecell{[c1:600, c2:200, c1:200,\\ c2:200, c1:200, c2:600]} & $c_{i \rightarrow j}$: $8.41\times10^{-12}$ & $c_{i \rightarrow j}$: 0.0674 & 2.9557 & $c_{i \rightarrow j}$: T & $c_{i \rightarrow j}$: $1.21\times10^{-73}$ & 0.3186 \\

\bottomrule
\end{tabular}%
}
\end{table}

\begin{table}[H]
\caption{Synthetic Dataset Concept Definitions}
\vspace{-5pt}
\label{tab:synthetic-concept-definitions}
\tiny
\setlength{\tabcolsep}{1pt}
\begin{tabular}{p{0.07\textwidth}@{\hspace{8pt}}p{0.89\textwidth}}
\toprule
\textbf{Dataset} & \textbf{Concept Definitions} \\
\midrule

\textbf{ADS01} &
\parbox[t]{0.89\textwidth}{
$C_1$: $\mathbf{x}\sim\mathcal{N}(0.0,1.0^2)$, $y|\mathbf{x}\sim\mathcal{N}(3.0\mathbf{x}+5.0,1.5^2)$\\
$C_2$: $\mathbf{x}\sim\mathcal{N}(0.5,1.0^2)$, $y|\mathbf{x}\sim\mathcal{N}(2.2\mathbf{x}+6.0,1.5^2)$
} \\

\textbf{ADS02} &
\parbox[t]{0.91\textwidth}{
$C_1$: $\mathbf{x}\sim\mathcal{N}(0.0,1.0^2)$, $y|\mathbf{x}\sim\mathcal{N}(3.0\mathbf{x}+5.0,1.5^2)$\\
$C_2$: $\mathbf{x}\sim\mathcal{N}(1.0,1.0^2)$, $y|\mathbf{x}\sim\mathcal{N}(0.5\mathbf{x}+9.0,1.5^2)$
} \\

\textbf{ADS03} &
\parbox[t]{0.91\textwidth}{
$C_1$: $\mathbf{x}\sim\mathcal{N}(0.0,1.0^2)$, $y|\mathbf{x}\sim\mathcal{N}(3.0\mathbf{x}+5.0,1.5^2)$\\
$C_2$: $\mathbf{x}\sim\mathcal{N}(1.5,1.0^2)$, $y|\mathbf{x}\sim\mathcal{N}(-2.5\mathbf{x}+12.0,1.5^2)$
} \\

\textbf{ADS04} &
\parbox[t]{0.91\textwidth}{
$C_1$: $\mathbf{x}\sim\mathcal{N}(0.0,1.0^2)$, $y|\mathbf{x}\sim\mathcal{N}(\boldsymbol{\beta}_1^\top\mathbf{x}+5.0,1.5^2)$, $\boldsymbol{\beta}_1=[3.0,1.5,-1.0,0.5,2.0,-2.5,1.0,-0.5,0.8,-1.2]$\\
$C_2$: $\mathbf{x}\sim\mathcal{N}(0.2,1.0^2)$, $y|\mathbf{x}\sim\mathcal{N}(\boldsymbol{\beta}_2^\top\mathbf{x}+6.0,1.5^2)$, $\boldsymbol{\beta}_2=[2.8,1.4,-0.9,0.6,1.9,-2.3,1.1,-0.4,0.9,-1.0]$
} \\

\textbf{ADS05} &
\parbox[t]{0.91\textwidth}{
$C_1$: $\mathbf{x}\sim\mathcal{N}(0.0,1.0^2)$, $y|\mathbf{x}\sim\mathcal{N}(\boldsymbol{\beta}_1^\top\mathbf{x}+5.0,1.5^2)$, $\boldsymbol{\beta}_1=[3.0,1.5,-1.0,0.5,2.0,-2.5,1.0,-0.5,0.8,-1.2]$\\
$C_2$: $\mathbf{x}\sim\mathcal{N}(0.8,1.0^2)$, $y|\mathbf{x}\sim\mathcal{N}(\boldsymbol{\beta}_2^\top\mathbf{x}+8.0,1.5^2)$, $\boldsymbol{\beta}_2=[1.8,0.8,-0.2,1.2,1.0,-1.2,0.3,0.2,1.4,-0.3]$
} \\

\textbf{ADS06} &
\parbox[t]{0.91\textwidth}{
$C_1$: $\mathbf{x}\sim\mathcal{N}(0.0,1.0^2)$, $y|\mathbf{x}\sim\mathcal{N}(\boldsymbol{\beta}_1^\top\mathbf{x}+5.0,1.5^2)$, $\boldsymbol{\beta}_1=[3.0,1.5,-1.0,0.5,2.0,-2.5,1.0,-0.5,0.8,-1.2]$\\
$C_2$: $\mathbf{x}\sim\mathcal{N}(1.5,1.0^2)$, $y|\mathbf{x}\sim\mathcal{N}(\boldsymbol{\beta}_2^\top\mathbf{x}+12.0,1.5^2)$, $\boldsymbol{\beta}_2=[-2.0,-1.5,2.0,-1.0,-2.5,2.5,-1.5,1.5,-2.0,2.0]$
} \\\\

\textbf{IDS01} &
\parbox[t]{0.91\textwidth}{
$C_k$: $\mathbf{x}\sim\mathcal{N}(\mu_k,1.0^2)$, $y|\mathbf{x}\sim\mathcal{N}(\beta_k\mathbf{x}+b_k,1.5^2)$, $k=1,\ldots,10$\\
$\mu_k=0.0+\frac{k-1}{9}(0.3-0.0)$, 
$\beta_k=3.0+\frac{k-1}{9}(2.8-3.0)$, 
$b_k=5.0+\frac{k-1}{9}(6.0-5.0)$
} \\

\textbf{IDS02} &
\parbox[t]{0.91\textwidth}{
$C_k$: $\mathbf{x}\sim\mathcal{N}(\mu_k,1.0^2)$, $y|\mathbf{x}\sim\mathcal{N}(\beta_k\mathbf{x}+b_k,1.5^2)$, $k=1,\ldots,10$\\
$\mu_k=0.0+\frac{k-1}{9}(1.0-0.0)$,
$\beta_k=3.0+\frac{k-1}{9}(1.0-3.0)$,
$b_k=5.0+\frac{k-1}{9}(8.0-5.0)$
} \\

\textbf{IDS03} &
\parbox[t]{0.91\textwidth}{
$C_k$: $\mathbf{x}\sim\mathcal{N}(\mu_k,1.0^2)$, $y|\mathbf{x}\sim\mathcal{N}(\beta_k\mathbf{x}+b_k,1.5^2)$, $k=1,\ldots,10$\\
$\mu_k=0.0+\frac{k-1}{9}(1.5-0.0)$,
$\beta_k=3.0+\frac{k-1}{9}(-2.5-3.0)$,
$b_k=5.0+\frac{k-1}{9}(12.0-5.0)$
} \\

\textbf{IDS04} &
\parbox[t]{0.91\textwidth}{
$C_k$: $\mathbf{x}\sim\mathcal{N}(\mu_k,1.0^2)$, $y|\mathbf{x}\sim\mathcal{N}(\boldsymbol{\beta}_k^\top\mathbf{x}+b_k,1.5^2)$, $k=1,\ldots,10$\\
$\mu_k=0.0+\frac{k-1}{9}(0.2-0.0)$,
$\boldsymbol{\beta}_k=\boldsymbol{\beta}_1+\frac{k-1}{9}(\boldsymbol{\beta}_{10}-\boldsymbol{\beta}_1)$,
$b_k=5.0+\frac{k-1}{9}(6.0-5.0)$\\
$\boldsymbol{\beta}_1=[3.0,1.5,-1.0,0.5,2.0,-2.5,1.0,-0.5,0.8,-1.2]$,
$\boldsymbol{\beta}_{10}=[2.8,1.4,-0.9,0.6,1.9,-2.3,1.1,-0.4,0.9,-1.0]$
} \\

\textbf{IDS05} &
\parbox[t]{0.91\textwidth}{
$C_k$: $\mathbf{x}\sim\mathcal{N}(\mu_k,1.0^2)$, $y|\mathbf{x}\sim\mathcal{N}(\boldsymbol{\beta}_k^\top\mathbf{x}+b_k,1.5^2)$, $k=1,\ldots,10$\\
$\mu_k=0.0+\frac{k-1}{9}(0.8-0.0)$,
$\boldsymbol{\beta}_k=\boldsymbol{\beta}_1+\frac{k-1}{9}(\boldsymbol{\beta}_{10}-\boldsymbol{\beta}_1)$,
$b_k=5.0+\frac{k-1}{9}(8.0-5.0)$\\
$\boldsymbol{\beta}_1=[3.0,1.5,-1.0,0.5,2.0,-2.5,1.0,-0.5,0.8,-1.2]$,
$\boldsymbol{\beta}_{10}=[1.8,0.8,-0.2,1.2,1.0,-1.2,0.3,0.2,1.4,-0.3]$
} \\

\textbf{IDS06} &
\parbox[t]{0.91\textwidth}{
$C_k$: $\mathbf{x}\sim\mathcal{N}(\mu_k,1.0^2)$, $y|\mathbf{x}\sim\mathcal{N}(\boldsymbol{\beta}_k^\top\mathbf{x}+b_k,1.5^2)$, $k=1,\ldots,10$\\
$\mu_k=0.0+\frac{k-1}{9}(1.5-0.0)$,
$\boldsymbol{\beta}_k=\boldsymbol{\beta}_1+\frac{k-1}{9}(\boldsymbol{\beta}_{10}-\boldsymbol{\beta}_1)$,
$b_k=5.0+\frac{k-1}{9}(12.0-5.0)$\\
$\boldsymbol{\beta}_1=[3.0,1.5,-1.0,0.5,2.0,-2.5,1.0,-0.5,0.8,-1.2]$,
$\boldsymbol{\beta}_{10}=[-2.0,-1.5,2.0,-1.0,-2.5,2.5,-1.5,1.5,-2.0,2.0]$
} \\\\

\textbf{GDS01} &
\parbox[t]{0.91\textwidth}{
$C_1$: $\mathbf{x}\sim\mathcal{N}(0.0,1.0^2)$, $y|\mathbf{x}\sim\mathcal{N}(3.0\mathbf{x}+5.0,1.5^2)$\\
$C_2$: $\mathbf{x}\sim\mathcal{N}(0.5,1.0^2)$, $y|\mathbf{x}\sim\mathcal{N}(2.2\mathbf{x}+6.0,1.5^2)$\\
Segments: $[C_1:300,\ C_2:100,\ C_1:100,\ C_2:100,\ C_1:100,\ C_2:300]$
} \\

\textbf{GDS02} &
\parbox[t]{0.91\textwidth}{
$C_1$: $\mathbf{x}\sim\mathcal{N}(0.0,1.0^2)$, $y|\mathbf{x}\sim\mathcal{N}(3.0\mathbf{x}+5.0,1.5^2)$\\
$C_2$: $\mathbf{x}\sim\mathcal{N}(1.0,1.0^2)$, $y|\mathbf{x}\sim\mathcal{N}(0.5\mathbf{x}+9.0,1.5^2)$\\
Segments: $[C_1:300,\ C_2:100,\ C_1:100,\ C_2:100,\ C_1:100,\ C_2:300]$
} \\

\textbf{GDS03} &
\parbox[t]{0.91\textwidth}{
$C_1$: $\mathbf{x}\sim\mathcal{N}(0.0,1.0^2)$, $y|\mathbf{x}\sim\mathcal{N}(3.0\mathbf{x}+5.0,1.5^2)$\\
$C_2$: $\mathbf{x}\sim\mathcal{N}(1.5,1.0^2)$, $y|\mathbf{x}\sim\mathcal{N}(-2.5\mathbf{x}+12.0,1.5^2)$\\
Segments: $[C_1:300,\ C_2:100,\ C_1:100,\ C_2:100,\ C_1:100,\ C_2:300]$
} \\

\textbf{GDS04} &
\parbox[t]{0.91\textwidth}{
$C_1$: $\mathbf{x}\sim\mathcal{N}(0.0,1.0^2)$, $y|\mathbf{x}\sim\mathcal{N}(\boldsymbol{\beta}_1^\top\mathbf{x}+5.0,1.5^2)$, $\boldsymbol{\beta}_1=[3.0,1.5,-1.0,0.5,2.0,-2.5,1.0,-0.5,0.8,-1.2]$\\
$C_2$: $\mathbf{x}\sim\mathcal{N}(0.2,1.0^2)$, $y|\mathbf{x}\sim\mathcal{N}(\boldsymbol{\beta}_2^\top\mathbf{x}+6.0,1.5^2)$, $\boldsymbol{\beta}_2=[2.8,1.4,-0.9,0.6,1.9,-2.3,1.1,-0.4,0.9,-1.0]$\\
Segments: $[C_1:300,\ C_2:100,\ C_1:100,\ C_2:100,\ C_1:100,\ C_2:300]$
} \\

\textbf{GDS05} &
\parbox[t]{0.91\textwidth}{
$C_1$: $\mathbf{x}\sim\mathcal{N}(0.0,1.0^2)$, $y|\mathbf{x}\sim\mathcal{N}(\boldsymbol{\beta}_1^\top\mathbf{x}+5.0,1.5^2)$, $\boldsymbol{\beta}_1=[3.0,1.5,-1.0,0.5,2.0,-2.5,1.0,-0.5,0.8,-1.2]$\\
$C_2$: $\mathbf{x}\sim\mathcal{N}(0.8,1.0^2)$, $y|\mathbf{x}\sim\mathcal{N}(\boldsymbol{\beta}_2^\top\mathbf{x}+8.0,1.5^2)$, $\boldsymbol{\beta}_2=[1.8,0.8,-0.2,1.2,1.0,-1.2,0.3,0.2,1.4,-0.3]$\\
Segments: $[C_1:300,\ C_2:100,\ C_1:100,\ C_2:100,\ C_1:100,\ C_2:300]$
} \\

\textbf{GDS06} &
\parbox[t]{0.91\textwidth}{
$C_1$: $\mathbf{x}\sim\mathcal{N}(0.0,1.0^2)$, $y|\mathbf{x}\sim\mathcal{N}(\boldsymbol{\beta}_1^\top\mathbf{x}+5.0,1.5^2)$, $\boldsymbol{\beta}_1=[3.0,1.5,-1.0,0.5,2.0,-2.5,1.0,-0.5,0.8,-1.2]$\\
$C_2$: $\mathbf{x}\sim\mathcal{N}(1.5,1.0^2)$, $y|\mathbf{x}\sim\mathcal{N}(\boldsymbol{\beta}_2^\top\mathbf{x}+12.0,1.5^2)$, $\boldsymbol{\beta}_2=[-2.0,-1.5,2.0,-1.0,-2.5,2.5,-1.5,1.5,-2.0,2.0]$\\
Segments: $[C_1:300,\ C_2:100,\ C_1:100,\ C_2:100,\ C_1:100,\ C_2:300]$
} \\

\bottomrule
\end{tabular}
\end{table}

\begin{table}[H]
\scriptsize
\centering
\captionsetup{justification=centering}
\caption{Real Dataset Properties
\vspace{-.1cm}
\begin{minipage}[t]{\textwidth}
\centering
\tiny
\textbf{Segment Definition:} For every dataset, the first half is compared with the second half;
\textbf{Y KS P-Value:} Kolmogorov--Smirnov test p-value;
\textbf{Y JS Div:} Jensen--Shannon divergence;
\textbf{Y Wasserstein Dist:} Normalized Wasserstein distance for $y$;
\textbf{X Any KS Sig:} True if any feature shows significant KS drift ($p < 0.05$);
\textbf{X Min KS P-Value:} Minimum KS p-value across features;
\textbf{X Avg JS Div:} Average JS divergence across features;
\textbf{Full Param Distance:} Euclidean distance between the fitted intercept and coefficient vectors.
\end{minipage}
}

\renewcommand{\theadalign}{bc}
\renewcommand{\theadfont}{\bfseries\scriptsize}
\setlength{\tabcolsep}{3pt}

\resizebox{\textwidth}{!}{%
\begin{tabular}{lcccccccccccc}
\toprule
\thead{Dataset} &
\thead{Type} &
\thead{Domain} &
\thead{Data-\\points} &
\thead{Dimen-\\sions} &
\thead{Target} &
\thead{Y KS\\P-Value} &
\thead{Y JS\\Div} &
\thead{Y Wasserstein\\Dist} &
\thead{X Any\\KS Sig} &
\thead{X Min KS\\P-Value} &
\thead{X Avg\\JS Div} &
\thead{Full Param\\Distance} \\
\midrule

\textbf{MCPD} \cite{MCPD_DS}
& Real
& \makecell{Health-\\care}
& 1.3k
& 8
& \makecell{Insurance\\cost}
& 0.769256
& 0.009813
& 0.036071
& False
& 0.052801
& 0.004202
& 0.054078 \\

\textbf{1KC} \cite{1KC_DS}
& Real
& \makecell{Business/\\Finance}
& 1k
& 6
& Profit
& 0.952778
& 0.008088
& 0.005040
& False
& 0.890432
& 0.010709
& 0.739862 \\

\textbf{KCHSD} \cite{KCHS_DS}
& Real
& \makecell{Housing/\\Real Estate}
& 21.6k
& 7
& \makecell{House\\price}
& 0.000519
& 0.001960
& 0.058337
& True
& $1.04{\times}10^{-24}$
& 0.002204
& 0.115616 \\

\textbf{CCPP} \cite{CCPP_DS}
& Real
& \makecell{Energy\\Systems}
& 9.6k
& 4
& \makecell{Net hourly\\power}
& 0.610529
& 0.002593
& 0.005124
& True
& 0.016662
& 0.001811
& 0.003641 \\

\textbf{UCIAQD} \cite{UCIAQD_DS}
& Real
& \makecell{Environmental/\\Air Quality}
& 7.7k
& 8
& \makecell{CO\\concentration}
& $1.28{\times}10^{-19}$
& 0.017248
& 0.290262
& True
& $<10^{-300}$
& 0.147997
& 0.759503 \\

\textbf{GASD} \cite{GASD_DS}
& Real
& \makecell{Chemical\\Sensing}
& 13.9k
& 134
& \makecell{Gas\\concentration}
& $2.67{\times}10^{-57}$
& 0.088123
& 0.243632
& True
& $<10^{-300}$
& 0.067600
& 3.502364 \\

\textbf{CalCOFI} \cite{CALCOFI_DS}
& Real
& Oceanography
& 814.2k
& 10
& Salinity
& $<10^{-300}$
& 0.038709
& 0.019585
& True
& $<10^{-300}$
& 0.132505
& 0.018597 \\

\textbf{WSSF} \cite{WSSF_DS}
& Real
& \makecell{Retail/Sales\\Forecasting}
& 258.5k
& 145
& \makecell{Weekly\\sales}
& $3.63{\times}10^{-7}$
& 0.000143
& 0.000269
& True
& $<10^{-300}$
& 0.022639
& 0.013024 \\

\bottomrule
\end{tabular}%
}

\label{tab:real_datasets-properties}
\end{table}

\subsection{Experimental Setup}

For the synthetic evaluation, we consider three controlled drift categories: \textit{abrupt}, \textit{incremental}, and \textit{alternating gradual}. These datasets are systematically constructed to control key characteristics, including drift magnitude, feature dimensionality, and noise level, enabling a rigorous and reproducible experimental setting. This controlled design facilitates a detailed analysis of model behavior under different forms of distributional change.

To ensure a fair and comprehensive comparison across learning paradigms, four representative online regression models are employed: \textit{Recursive Least Squares (RLS)} \cite{fontenla2013online}, \textit{Passive-Aggressive (PA)} \cite{crammer2006online}, \textit{Widrow-Hoff (LMS)} \cite{widrow1960adaptive}, and \textit{Online Regression with Weighted Average (OLR-WA)} \cite{abu2023olr}. These models were selected due to their diverse update mechanisms, capturing a range of learning dynamics from fast reactive updates to more stable incremental adaptations.

The performance of each model is analyzed both \textit{before and after integration of} \modelname, allowing for a direct assessment of its impact. Furthermore, the proposed approach is benchmarked against established drift-handling baselines that combine statistical drift detection with online adaptation strategies. Specifically, we consider the following detector--adaptation combinations: \textit{ADWIN-RESET (AR), ADWIN-WINDOW (AW), ADWIN-SSPT (AS), ADWIN-OHL (AO), KSWIN-RESET (KR), KSWIN-WINDOW (KW), KSWIN-SSPT (KS), and KSWIN-OHL (KO)}.

For each online regression model, we evaluate the standalone base model, its corresponding \modelname-integrated variant, and the eight detector--adaptation benchmark variants. For example, for PA online regression model, we evaluate \textit{PA}, \textit{PA-\modelname}, \textit{PA-AR}, \textit{PA-AW}, \textit{PA-AS}, \textit{PA-AO}, \textit{PA-KR}, \textit{PA-KW}, \textit{PA-KS}, and \textit{PA-KO}. The same evaluation structure is applied to RLS, LMS, and OLR-WA.

Drift detection is performed using \textit{ADWIN} \cite{bifet2007learning} and \textit{KSWIN} \cite{raab2020reactive}, two widely adopted methods for identifying distributional changes in data streams. \textit{ADWIN} employs an adaptive windowing mechanism with statistical guarantees, while \textit{KSWIN} utilizes the Kolmogorov--Smirnov test to detect discrepancies between data distributions. For model adaptation, four standard strategies are considered: \textit{RESET}, which follows a model replacement approach by reinitializing the model upon drift detection \cite{celik2023online}, and \textit{WINDOW}, which performs model retraining using recent data through a sliding window \cite{baier2021detecting}. \textit{SSPT} extends the Self Parameter Tuning (SPT) framework to data streams by incorporating the Nelder--Mead optimization algorithm to continuously search for suitable hyperparameter configurations in a single pass over the data \cite{veloso2021hyperparameter}. \textit{OHL} (Online Hyperparameter Learning) performs gradient-based hyperparameter optimization by computing hyper-gradients from incoming data and updating hyperparameters online, enabling continuous adaptation under non-stationary environments \cite{zhan2018efficient}. These strategies are selected due to their effectiveness and widespread use in the literature, providing strong, interpretable, and well-established baselines for comparison in online learning under concept drift.

\subsubsection{Baseline Configuration Protocol}
\label{subsubsec:baseline_configuration_protocol}

The baseline detectors were implemented using the documented default
configurations of the scikit-multiflow reference implementation
\cite{montiel2018scikit}. Specifically, ADWIN used
\(\delta=0.002\), while KSWIN used
\(\alpha_{\mathrm{KS}}=0.005\),
\(W_{\mathrm{KS}}=100\), and
\(S_{\mathrm{KS}}=30\).\footnote{The documented implementation defaults are
available in the official scikit-multiflow API documentation:
\url{https://scikit-multiflow.readthedocs.io/en/stable/api/generated/skmultiflow.drift_detection.ADWIN.html}
and
\url{https://scikit-multiflow.readthedocs.io/en/stable/api/generated/skmultiflow.drift_detection.KSWIN.html}.}

These detector configurations were fixed and applied unchanged across all
regression models, datasets, and evaluation seeds. Under the strict
single-pass online evaluation protocol adopted in this study, selecting
detector parameters through an offline grid search over the complete
evaluated stream is not permissible because it would require access to future
observations that are unavailable at the corresponding prediction time. Such
a procedure would violate the online evaluation setting and introduce
look-ahead bias or data snooping. Consequently, no detector candidate grid, designated pilot seed, separate
pilot stream, or data-driven detector-parameter selection procedure was used.
The adaptation-side settings of WINDOW, SSPT, and OHL were also fixed
\textit{a priori} and applied unchanged across all datasets, regression
models, and evaluation seeds. These settings were not selected by evaluating
the complete streams or by choosing the configurations that produced the best
reported results. This protocol preserves the strict single-pass online
setting, avoids access to future observations, and provides a consistent
comparison without look-ahead bias or test-stream tuning.

The detector parameters remained fixed throughout stream processing. They
should therefore be distinguished from the adaptation-side parameters used
by WINDOW, SSPT, and OHL. RESET did not introduce an additional adaptation
hyperparameter; when a detector alarm occurred, the corresponding learner was
reinitialized using the same initialization rule as the standalone learner.
WINDOW used a fixed recent-memory length, SSPT used predefined
model-specific candidate sets, and OHL used fixed step sizes, perturbation
values, and permitted model-hyperparameter ranges. SSPT and OHL adapted the
underlying learner hyperparameters rather than the ADWIN or KSWIN detector
parameters.

The complete detector and adaptation configurations are reported in
Tables~\ref{tab:olr-wa-hyperparameters}--\ref{tab:lms-hyperparameters}.
These configurations were applied unchanged across the five evaluation seeds
\(\{0,1,42,123,7\}\), with no per-seed parameter adjustment.

The alarm-evaluation parameters
\(r_{\mathrm{tol}}=0.05\),
\(c_{\mathrm{cool}}=2.0\), and
\(m_{\mathrm{ep}}=2\)
are separate from the detector and adaptation parameters. They were applied
uniformly to all methods and were not optimized separately for a regression
model, dataset, seed, or drift category. Their influence is evaluated
independently in the complete alarm-protocol sensitivity analysis.

In summary, Table~\ref{tab:experimental-setup} provides an overview of the
experimental protocol. The experiments combine multiple online regression
models, the proposed \modelname-integrated variants, detector--adaptation
baselines, synthetic and real-world datasets, and repeated runs over multiple
random seeds. The reported results are aggregated by model, method, and drift
type to provide a reliable assessment of predictive performance under
different non-stationary conditions.

\small
\renewcommand{\arraystretch}{1.25}
\setlength{\LTleft}{0pt}
\setlength{\LTright}{0pt}
\setlength{\LTcapwidth}{\textwidth}

\begin{longtable}{@{}
>{\raggedright\arraybackslash}p{0.17\textwidth}
>{\raggedright\arraybackslash}p{0.79\textwidth}
@{}}

\caption{Experimental Setup Summary.}
\label{tab:experimental-setup}\\

\toprule
\textbf{Component} & \textbf{Description} \\
\midrule
\endfirsthead

\multicolumn{2}{c}{%
\textit{\tablename\ \thetable{} continued from previous page}} \\
\toprule
\textbf{Component} & \textbf{Description} \\
\midrule
\endhead

\midrule
\multicolumn{2}{r}{\textit{Continued on next page}} \\
\endfoot

\bottomrule
\endlastfoot

\textbf{Online Regression Models} &
Recursive Least Squares (RLS) \cite{fontenla2013online},
Passive-Aggressive (PA) \cite{crammer2006online},
Widrow-Hoff (LMS) \cite{widrow1960adaptive}, and
Online Regression with Weighted Average (OLR-WA)
\cite{abu2023olr}. \\

\textbf{Drift-handling Methods} &
The proposed \modelname{} framework and eight detector--adaptation
baselines: ADWIN-RESET (AR), ADWIN-WINDOW (AW), ADWIN-SSPT (AS),
ADWIN-OHL (AO), KSWIN-RESET (KR), KSWIN-WINDOW (KW),
KSWIN-SSPT (KS), and KSWIN-OHL (KO). \\

\textbf{Synthetic Datasets} &
18 synthetic datasets covering abrupt, incremental, and alternating
gradual drift scenarios, with variations in dimensionality, sample size,
drift magnitude, and noise level, as summarized in
Table~\ref{tab:synthetic-datasets-properties}. \\

\textbf{Real-world Datasets} &
Eight real-world regression datasets representing healthcare, business
and finance, housing and real estate, energy systems, environmental
monitoring, chemical sensing, oceanography, and retail sales forecasting,
as summarized in Table~\ref{tab:real_datasets-properties}. \\

\textbf{Baseline Configuration} &
ADWIN and KSWIN used the documented scikit-multiflow default configurations,
which were fixed across all regression models, datasets, and evaluation
seeds. No offline detector grid search, pilot seed, pilot stream, or
data-driven detector-parameter calibration or selection procedure was used
because evaluating candidate configurations over the complete stream would
require access to future observations and violate the strict single-pass
online evaluation protocol. WINDOW, SSPT, and OHL adaptation settings are
reported in the corresponding configuration tables. All configurations were
applied unchanged across the five evaluation seeds. \\

\textbf{Repeated Runs} &
Both predictive-performance and drift-alarm experiments were executed using
five independent random seeds:
\(\{0,1,42,123,7\}\). \\

\textbf{Aggregation Strategy} &
Predictive-performance results are aggregated by drift type across the
corresponding datasets and five independent random seeds. Alarm-quality
metrics are first calculated separately for each
model--dataset--seed combination. For each drift-type summary, the alarm
counts from the six corresponding datasets and four regression models are
pooled within each seed, and the resulting seed-level metrics are reported as
mean \(\pm\) standard deviation over the five seeds. \\

\textbf{Performance Metrics} &
The coefficient of determination (\(R^2\)) is used for mini-batch
assessment, while Mean Squared Error (MSE) is used for both single-instance
online evaluation and mini-batch evaluation. \\

\end{longtable}

\subsection{Evaluation Metrics and Statistical Protocols}

\subsubsection{Predictive Performance Metrics}

The coefficient of determination ($R^2$) is widely used in regression analysis due to its normalized and interpretable nature, making it a useful complement to common error-based metrics such as MSE, MAE, RMSE, and MAPE~\cite{chicco2021coefficient}. Specifically, $R^2$ values lie in $(-\infty, 1]$, with values closer to 1 indicating stronger predictive performance. Unlike MSE or RMSE, which are unbounded and sensitive to the scale of the target variable, $R^2$ provides a normalized measure of goodness-of-fit by quantifying the proportion of variance in the dependent variable explained by the model. This scale-invariant property facilitates meaningful comparisons across datasets and domains~\cite{chicco2021coefficient}. Furthermore, $R^2$ reflects how well the predicted values align with the variability of the observed target values, making it one of the standard measures of model fit in regression analysis~\cite{montgomery2021introduction}. Supported by strong statistical foundations~\cite{nagelkerke1991note}, $R^2$ remains a commonly adopted evaluation metric across scientific and machine learning applications.

However, $R^2$ requires more than one data point for meaningful computation, which limits its applicability in strictly online settings where models process one instance at a time. In this work, $R^2$ is therefore used when mini-batches are available, as in the OLR-WA model. For fully online models such as PA, RLS, and LMS, which operate on single data points, the Mean Squared Error (MSE) is employed. MSE is a standard regression loss function~\cite{hastie2009elements} and provides a direct absolute measure of prediction error at the instance level. It penalizes larger deviations more strongly than smaller ones and preserves the magnitude of prediction errors, making it suitable for evaluating point-wise online regression models. Although MSE is scale-dependent and should be interpreted with respect to the target variable scale, it is appropriate in this setting because the fully online models are evaluated sequentially on individual predictions where $R^2$ cannot be meaningfully computed.

\subsubsection{Predictive Statistical Protocol}
\label{subsubsec:predictive_statistical_protocol}

Predictive performance is calculated separately for every model--dataset--seed run using five independent evaluation seeds. Mean and standard deviation are then reported across the paired dataset--seed observations within each regression model and drift category. For each model and drift category, \modelname{} is compared with the
corresponding standalone learner and separately with each of the eight
detector--adaptation baselines. The strongest detector--adaptation baseline,
defined as the baseline with the best mean full-stream predictive performance,
is additionally identified for concise presentation in the main results table.
However, statistical inference is conducted over all eight baseline
comparisons rather than only the data-selected strongest baseline.

Each paired comparison contains 30 (6 \text{ datasets}
$\times$
5 \text{ seeds}) paired observations. Two-sided Wilcoxon signed-rank tests are used to assess whether the distribution of paired performance differences is centered at zero, without prespecifying the direction of the difference \cite{Demsar2006}. The tests use the normal approximation with continuity correction and exclude zero paired differences. For each model and drift category, Holm correction is applied across the eight
comparisons between \modelname{} and the detector--adaptation baselines. The
comparison with the corresponding standalone learner is reported separately.
This procedure controls the family-wise error rate while avoiding statistical
inference based only on a baseline selected from the same observed results. Rank-biserial correlation is reported as an effect-size measure. For $R^2$, a positive paired difference favors \modelname{}; for MSE, a positive effect indicates that \modelname{} produces lower error.

\subsubsection{Drift-Alarm Quality Metrics and Alignment Protocol}
\label{sec:drift_alarm_quality}

In addition to predictive performance, drift-alarm quality is evaluated on the synthetic streams, where the ground-truth drift locations are known. We report true positives, false positives, missed drifts, detection delay, precision, recall, and \(F_1\)-score \cite{gama2014survey,gonccalves2014comparative,sokolova2009systematic}. Because the real-world datasets do not provide annotated drift locations or known concept boundaries, ground-truth alarm-quality metrics are evaluated only on the synthetic datasets.

The evaluation distinguishes among raw detector alarms, alarm episodes, and
model interventions. For the detector--adaptation baselines, a raw detector
alarm is an individual ADWIN or KSWIN detection produced before RESET,
WINDOW, SSPT, or OHL is activated. Every raw ADWIN or KSWIN detection
triggers one paired adaptation and is therefore retained in the primary
raw-alarm evaluation.

No cooldown consolidation or minimum-size filtering is applied to that
primary detector analysis. Alarm episodes are reported separately as a
secondary operational analysis of temporally adjacent alarm activity.
Adaptation and recalibration counts record model actions and are not treated
as alarm-quality counts.

For a stream containing \(N\) samples, the post-drift matching tolerance is

\begin{equation}
T
=
\operatorname{round}
\left(
r_{\mathrm{tol}}N
\right),
\label{eq:raw_alarm_tolerance}
\end{equation}

where \(r_{\mathrm{tol}}\) is the tolerance ratio. Let \(\mathcal{A}=\{a_1,\ldots,a_M\}\) denote the chronological raw alarm indices and \(\mathcal{D}=\{d_1,\ldots,d_J\}\) denote the ground-truth drift indices. Matching is chronological and one-to-one. For every ground-truth drift \(d_j\), the earliest unmatched raw alarm satisfying

\begin{equation}
d_j
\leq
a_i
\leq
d_j+T
\label{eq:raw_alarm_matching}
\end{equation}

is counted as one true positive. Every other unmatched raw alarm is counted as a false positive. Therefore, alarms occurring before a true drift, repeated alarms after the same drift has already been matched, and isolated alarms are retained as false positives rather than being discarded. A ground-truth drift without an eligible unmatched alarm is counted as a false negative.

For the alternating-gradual streams, every annotated boundary between
consecutive \(C_1\) and \(C_2\) segments remains a known local transition
point. The alarm analysis therefore evaluates the method's response to these
repeated local switches, while the overall dataset category is reported as
alternating gradual drift rather than classical smooth or probabilistic
gradual drift.

For an alarm matched to a drift at \(d_j\), conventional detection delay in samples is

\begin{equation}
\Delta
=
a_i-d_j.
\label{eq:alarm_detection_delay}
\end{equation}

When the corresponding model processes increments of \(K\) samples, delay in processing increments is

\begin{equation}
\Delta_{\mathrm{inc}}
=
\frac{\Delta}{K}.
\label{eq:alarm_increment_delay}
\end{equation}

No processing increment is subtracted. Consequently, zero delay is reported only when the matched alarm occurs at the annotated transition point. Delay is calculated only for successfully matched drifts and is interpreted jointly with recall.

For the secondary episode-level analysis, the cooldown distance is

\begin{equation}
C
=
\operatorname{round}
\left(
c_{\mathrm{cool}}T
\right),
\label{eq:alarm_tolerance_cooldown}
\end{equation}

where \(c_{\mathrm{cool}}\) is the cooldown factor. Let \(e_1\) denote the first raw event of a candidate episode. Every subsequent event \(e_j\) satisfying

\begin{equation}
e_j
\leq
e_1+C
\label{eq:alarm_episode_membership}
\end{equation}

is assigned to that episode. The boundary is fixed at \(e_1+C\) and is not repeatedly extended. The first event after this boundary starts a new candidate episode. A candidate episode is retained when it contains at least \(m_{\mathrm{ep}}\) events.

The alarm time of a retained episode \(\mathcal{E}\) is always its first event:

\begin{equation}
a(\mathcal{E})
=
\min(\mathcal{E}).
\label{eq:episode_alarm_time}
\end{equation}

An episode is eligible for matching only when

\begin{equation}
d
\leq
a(\mathcal{E})
\leq
d+T.
\label{eq:alarm_matching_window}
\end{equation}

Thus, a later post-drift trigger cannot convert an episode that began before the drift into a true positive. Ground-truth drifts and retained episodes are processed chronologically and one-to-one. A matched episode is counted as a true positive, an unmatched episode as a false positive, and an unmatched drift as a false negative.

For both raw and episode-level analyses,

\begin{equation}
\mathrm{Precision}
=
\frac{\mathrm{TP}}
{\mathrm{TP}+\mathrm{FP}},
\qquad
\mathrm{Recall}
=
\frac{\mathrm{TP}}
{\mathrm{TP}+\mathrm{FN}},
\label{eq:alarm_precision_recall}
\end{equation}

and

\begin{equation}
F_1
=
2
\frac{
\mathrm{Precision}\times\mathrm{Recall}
}{
\mathrm{Precision}+\mathrm{Recall}
}.
\label{eq:alarm_f1}
\end{equation}

When a denominator is zero, the corresponding metric is reported as zero.

The fixed episode protocol uses \(r_{\mathrm{tol}}=0.05\), \(c_{\mathrm{cool}}=2.0\), and \(m_{\mathrm{ep}}=2\). These values are applied identically to \modelname{}, ADWIN, and KSWIN and were fixed before the final comparative analysis. Their influence is examined independently in the protocol-sensitivity analysis.

All alarm-quality experiments are repeated using five random seeds. Metrics are
first calculated separately for every model--dataset--seed combination.
Drift-category summaries pool the counts from the six corresponding datasets
within each seed and report the mean and standard deviation of the resulting
five seed-level metrics. Statistical comparisons use paired
model--dataset--seed observations and two-sided Wilcoxon signed-rank tests.
Holm correction controls the family-wise error rate across the two
detector-family comparisons within each drift category, and rank-biserial
correlation is reported as an effect size.

Because detection delay is defined only for successfully matched drift
instances and may be unavailable when a method produces no true-positive
detection, the episode-level \(F_1\)-score is used as the primary inferential
endpoint for the paired statistical comparisons. Detection delay is reported
descriptively and interpreted jointly with recall.

The direct raw ADWIN/KSWIN results constitute the primary evaluation of the
standard detectors because every raw detection initiates an adaptation.
Episode-level consolidation is reported as a secondary analysis of persistent
alarm activity.

\subsection{Predictive Performance Analysis}

\subsubsection{Performance Analysis on Abrupt Datasets}

This section presents the aggregated performance of each method across all abrupt-drift datasets listed in Table~\ref{tab:synthetic-datasets-properties}. Specifically, we report the average, minimum, and maximum performance over the entire data stream. In addition, abrupt-drift datasets are characterized by a sudden and substantial change occurring near the middle of the stream; therefore, we further analyze the performance at the drift point, as well as immediately before and after the drift. This allows us to evaluate both the overall predictive behavior of each method and its responsiveness to abrupt concept changes.

It is important to note that the reported minimum values should be interpreted with caution. In online learning experiments, the minimum error or lowest $R^2$ value may occur at very early prediction steps, often before the model, drift detector, or adaptation mechanism has accumulated sufficient observations to become effective. Therefore, these values do not necessarily reflect the steady-state behavior or the true drift-handling capability of each method. For this reason, the average performance, maximum error, and localized behavior around the drift point are emphasized when interpreting the results.

\begin{figure}[!htbp]
\centering

\includegraphics[
  width=\textwidth,
  keepaspectratio
]{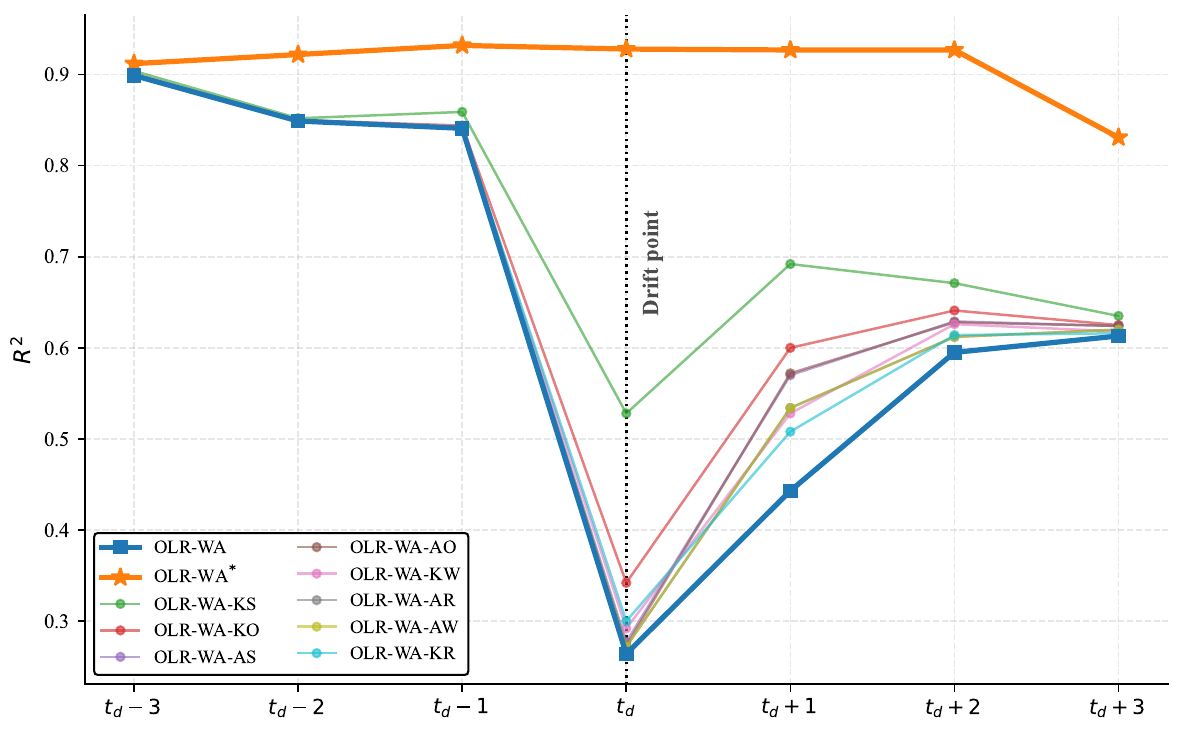}

\vspace{0.2cm}

\scriptsize
\resizebox{\textwidth}{!}{%
\begin{tabular}{lccc|ccc|ccccccc}
\toprule
\multirow{2}{*}{Method}
& \multicolumn{3}{c|}{$R^2$}
& \multicolumn{3}{c|}{MSE}
& \multicolumn{7}{c}{$R^2$ Around Drift ($t_d$)} \\
\cmidrule(lr){2-4}
\cmidrule(lr){5-7}
\cmidrule(lr){8-14}
& Avg & Min & Max
& Avg & Min & Max
& $t_d\!-\!3$ & $t_d\!-\!2$ & $t_d\!-\!1$ & $t_d$
& $t_d\!+\!1$ & $t_d\!+\!2$ & $t_d\!+\!3$ \\
\midrule
OLR-WA* & \textbf{0.804} & -0.025 & 0.962 & \textbf{1.724} & \textbf{0.717} & \textbf{3.323} & \textbf{0.912} & \textbf{0.922} & \textbf{0.932} & \textbf{0.928} & \textbf{0.927} & \textbf{0.927} & \textbf{0.831} \\
OLR-WA-KS   & 0.738 & 0.004 & 0.964 & 1.949 & 0.834 & 25.426 & 0.904 & 0.852 & 0.859 & 0.528 & 0.692 & 0.671 & 0.635 \\
OLR-WA-KO   & 0.723 & -0.261 & 0.965 & 2.021 & 0.890 & 25.257 & 0.900 & 0.850 & 0.844 & 0.342 & 0.600 & 0.641 & 0.625 \\
OLR-WA-AS   & 0.720 & -0.442 & 0.964 & 2.045 & 0.892 & 29.150 & 0.901 & 0.850 & 0.842 & 0.277 & 0.570 & 0.629 & 0.624 \\
OLR-WA-AO   & 0.719 & -0.442 & 0.965 & 2.057 & 0.892 & 29.094 & 0.900 & 0.849 & 0.841 & 0.273 & 0.572 & 0.628 & 0.624 \\
OLR-WA-KW   & 0.718 & -0.442 & 0.962 & 2.099 & 0.856 & 25.202 & 0.899 & 0.849 & 0.841 & 0.292 & 0.528 & 0.626 & 0.618 \\
OLR-WA-AR   & 0.717 & -0.442 & 0.962 & 2.104 & 0.892 & 29.079 & 0.899 & 0.849 & 0.841 & 0.272 & 0.534 & 0.612 & 0.620 \\
OLR-WA-AW   & 0.717 & -0.442 & 0.962 & 2.104 & 0.892 & 29.079 & 0.899 & 0.849 & 0.841 & 0.272 & 0.534 & 0.612 & 0.620 \\
OLR-WA-KR   & 0.717 & -0.442 & 0.962 & 2.102 & 0.856 & 25.202 & 0.899 & 0.849 & 0.841 & 0.300 & 0.508 & 0.614 & 0.616 \\
OLR-WA      & 0.715 & -0.442 & 0.962 & 2.157 & 0.892 & 30.661 & 0.899 & 0.849 & 0.841 & 0.264 & 0.443 & 0.595 & 0.613 \\
\bottomrule
\end{tabular}%
}

\caption{\scriptsize OLR-WA variants aggregated performance across abrupt-drift experiments. The plot illustrates the $R^2$ behavior around the drift point, while the table reports the aggregated average, minimum, and maximum $R^2$ and MSE values over the full stream, together with the detailed $R^2$ values before, at, and after the drift point. Here, $t_d$ denotes the drift point.}
\label{figtab:vis_agg_abrupt_olrwa}

\vspace{-.25cm}
\end{figure}

\paragraph{OLR-WA-Based Performance under Abrupt Drift}
As shown in Fig.~\ref{figtab:vis_agg_abrupt_olrwa}, OLR-WA$^{*}$, which represents OLR-WA integrated with \modelname, achieves the strongest overall performance among all OLR-WA-based variants on the abrupt-drift datasets. It obtains the highest average $R^2$ value of $0.804$ and the lowest average MSE of $1.724$, outperforming the standalone OLR-WA model, which records an average $R^2$ of $0.715$ and an average MSE of $2.157$. This improvement indicates that \modelname enhances the predictive stability of OLR-WA under sudden distributional changes.

The localized analysis around the drift point further supports this observation. At the drift point, OLR-WA$^{*}$ maintains a high $R^2$ of $0.928$, whereas the original OLR-WA drops sharply to $0.264$. Similarly, the ADWIN- and KSWIN-based detector--adaptation variants improve over the base model in some cases, but none of them matches the robustness of OLR-WA$^{*}$ during the abrupt transition. For example, the best competing variant around the drift point is OLR-WA-KS with an $R^2$ of $0.528$, which remains substantially lower than OLR-WA$^{*}$. This indicates that although conventional detector--adaptation strategies can provide partial recovery after drift, their response is less stable and less effective immediately around the transition point. In contrast, \modelname enables OLR-WA to preserve high predictive performance before, during, and after the abrupt drift, demonstrating stronger drift resilience and smoother adaptation behavior.

\begin{figure}[!htbp]
\centering

\includegraphics[
  width=\textwidth,
  keepaspectratio
]{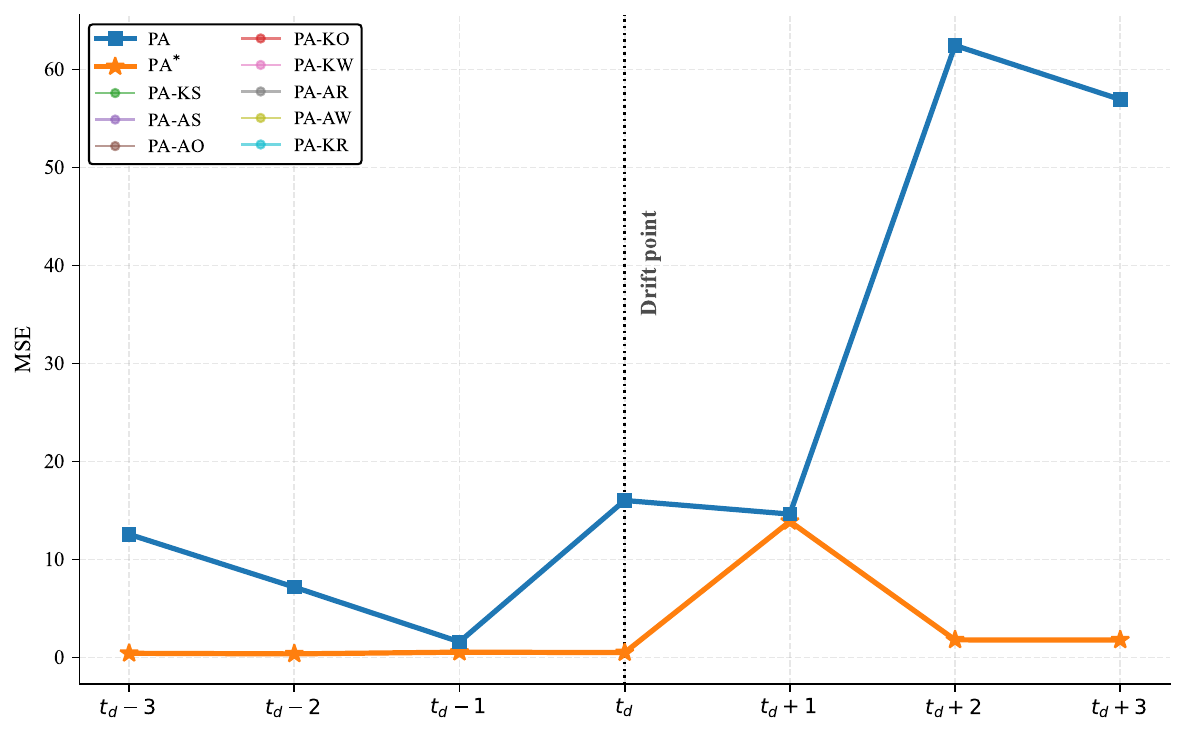}

\vspace{0.2cm}

\scriptsize
\resizebox{\textwidth}{!}{%
\begin{tabular}{lccc|ccccccc}
\toprule
\multirow{2}{*}{Method}
& \multicolumn{3}{c|}{MSE}
& \multicolumn{7}{c}{MSE Around Drift ($t_d$)} \\
\cmidrule(lr){2-4}
\cmidrule(lr){5-11}
& Avg & Min & Max
& $t_d\!-\!3$ & $t_d\!-\!2$ & $t_d\!-\!1$ & $t_d$
& $t_d\!+\!1$ & $t_d\!+\!2$ & $t_d\!+\!3$ \\
\midrule
PA*          & \textbf{1.246} & \textbf{0.000} & \textbf{108.720} & \textbf{0.422} & \textbf{0.361} & \textbf{0.540} & \textbf{0.491} & \textbf{13.882} & \textbf{1.791} & \textbf{1.787} \\
PA-KS    & 5.296 & 0.001 & 589.213 & 12.590 & 7.237 & 1.573 & 15.983 & 14.636 & 62.436 & 56.985 \\
PA-AS    & 5.307 & 0.001 & 589.212 & 12.567 & 7.177 & 1.594 & 16.024 & 14.630 & 62.467 & 56.953 \\
PA               & 5.323 & 0.001 & 589.212 & 12.567 & 7.177 & 1.594 & 16.024 & 14.630 & 62.467 & 56.953 \\
PA-AO     & 5.323 & 0.001 & 589.212 & 12.567 & 7.177 & 1.594 & 16.024 & 14.630 & 62.467 & 56.953 \\
PA-KO     & 5.324 & 0.001 & 589.292 & 12.571 & 7.178 & 1.595 & 16.023 & 14.630 & 62.468 & 56.944 \\
PA-KW  & 5.718 & 0.001 & 589.212 & 12.574 & 7.176 & 1.594 & 16.024 & 14.630 & 62.465 & 56.953 \\
PA-AR   & 5.808 & 0.001 & 589.212 & 12.567 & 7.177 & 1.594 & 16.024 & 14.630 & 62.467 & 56.953 \\
PA-AW  & 5.808 & 0.001 & 589.212 & 12.567 & 7.177 & 1.594 & 16.024 & 14.630 & 62.467 & 56.953 \\
PA-KR   & 5.843 & 0.001 & 589.282 & 12.579 & 7.176 & 1.594 & 16.024 & 14.633 & 62.465 & 56.959 \\
\bottomrule
\end{tabular}%
}

\caption{\scriptsize PA variants aggregated performance across abrupt-drift experiments. The plot illustrates the MSE behavior around the drift point, while the table reports the aggregated average, minimum, and maximum MSE values over the full stream, together with the detailed MSE values before, at, and after the drift point. Here, $t_d$ denotes the drift point.}
\label{figtab:vis_agg_abrupt_pa}
\end{figure}

\paragraph{PA-Based Performance under Abrupt Drift}
As shown in Fig.~\ref{figtab:vis_agg_abrupt_pa}, PA$^{*}$, which represents PA integrated with \modelname, achieves the best performance among all PA-based variants on the abrupt-drift datasets. PA$^{*}$ records the lowest average MSE of $1.246$, compared with $5.323$ for the standalone PA model, demonstrating a substantial improvement in robustness under sudden concept changes. In contrast, the ADWIN- and KSWIN-based detector--adaptation variants provide little to no improvement over the original PA model. Their average MSE values remain close to the standalone PA result, and their localized errors around the drift point follow nearly the same pattern. This behavior is consistent with the experimental logs, which show that ADWIN and KSWIN detect abrupt changes with a substantial delay. Consequently, the adaptation step is triggered too late, after the model has already passed the most critical region around the drift point.

The localized analysis further confirms this behavior. At the drift point, PA$^{*}$ obtains an MSE of $0.491$, whereas the original PA and most ADWIN /KSWIN-based variants produce much higher errors of approximately $16.024$. Although PA$^{*}$ shows a temporary increase at $t_d+1$ with an MSE of $13.882$, it quickly recovers to $1.791$ at $t_d+2$ and $1.787$ at $t_d+3$. By contrast, the detector--adaptation baselines remain unstable after the drift, with errors exceeding $56$ at $t_d+3$ for most methods. These results show that delayed detection limits the usefulness of conventional adaptation strategies, while \modelname enables PA to respond more effectively and recover faster after abrupt drift.

\begin{figure}[!htbp]
\centering

\includegraphics[
  width=\textwidth,
  keepaspectratio
]{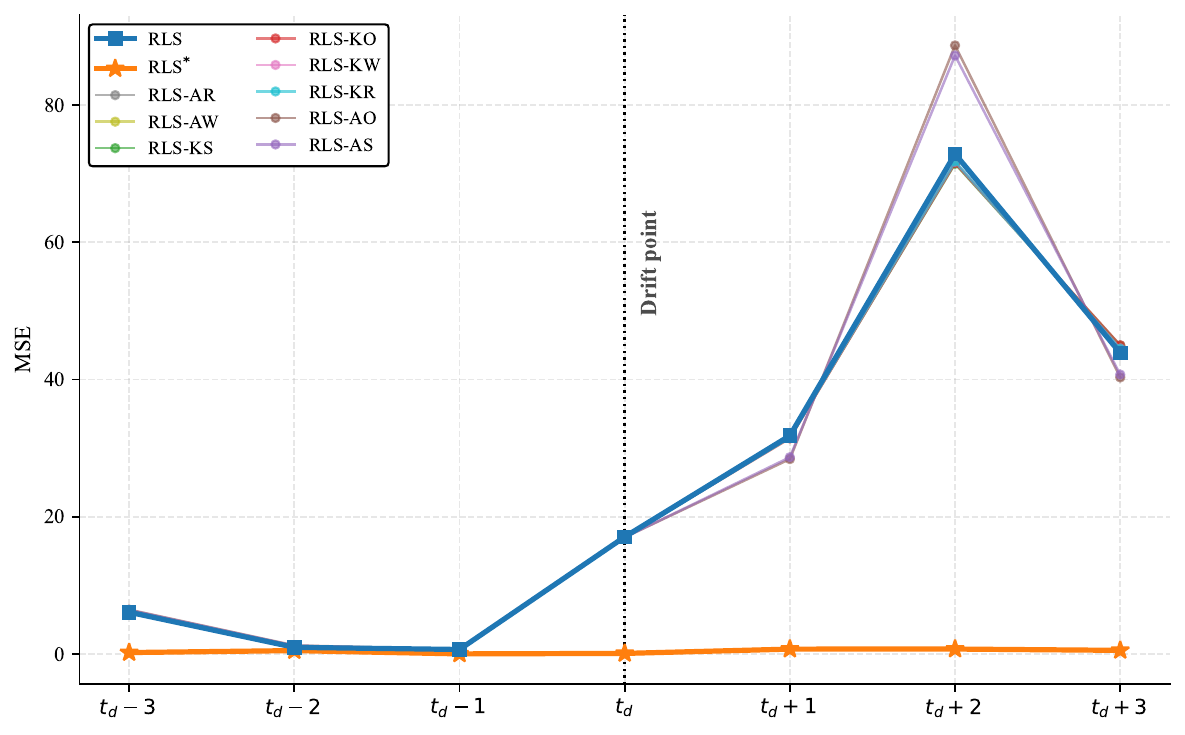}

\vspace{0.2cm}

\scriptsize
\resizebox{\textwidth}{!}{%
\begin{tabular}{lccc|ccccccc}
\toprule
\multirow{2}{*}{Method}
& \multicolumn{3}{c|}{MSE}
& \multicolumn{7}{c}{MSE Around Drift ($t_d$)} \\
\cmidrule(lr){2-4}
\cmidrule(lr){5-11}
& Avg & Min & Max
& $t_d\!-\!3$ & $t_d\!-\!2$ & $t_d\!-\!1$ & $t_d$
& $t_d\!+\!1$ & $t_d\!+\!2$ & $t_d\!+\!3$ \\
\midrule
RLS*          & \textbf{0.908} & 0.000 & \textbf{206.299} & \textbf{0.210} & \textbf{0.495} & \textbf{0.015} & \textbf{0.085} & \textbf{0.711} & \textbf{0.720} & \textbf{0.530} \\
RLS-AR   & 3.280 & 0.000 & 368.385 & 6.086 & 0.967 & 0.658 & 17.062 & 31.809 & 72.849 & 43.855 \\
RLS-AW  & 3.280 & 0.000 & 368.385 & 6.086 & 0.967 & 0.658 & 17.062 & 31.809 & 72.849 & 43.855 \\
RLS-KS    & 3.314 & 0.000 & 360.882 & 6.270 & 0.921 & 0.660 & 17.069 & 31.513 & 71.423 & 44.931 \\
RLS-KO     & 3.315 & 0.000 & 361.876 & 6.251 & 0.920 & 0.668 & 17.034 & 31.412 & 71.518 & 44.983 \\
RLS-KW  & 3.370 & 0.000 & 362.634 & 6.327 & 0.963 & 0.662 & 16.983 & 31.521 & 71.717 & 44.507 \\
RLS-KR   & 3.395 & 0.000 & 362.888 & 6.338 & 0.961 & 0.663 & 16.961 & 31.493 & 71.750 & 44.461 \\
RLS-AO     & 3.419 & 0.000 & 465.521 & 6.461 & 1.252 & 0.667 & 16.999 & 28.398 & 88.660 & 40.298 \\
RLS-AS    & 3.438 & 0.000 & 456.338 & 6.409 & 1.214 & 0.665 & 16.984 & 28.665 & 87.185 & 40.707 \\
RLS               & 3.906 & 0.000 & 368.385 & 6.086 & 0.967 & 0.659 & 17.062 & 31.809 & 72.850 & 43.856 \\
\bottomrule
\end{tabular}%
}

\caption{\scriptsize RLS variants aggregated performance across abrupt-drift experiments. The plot illustrates the MSE behavior around the drift point, while the table reports the aggregated average, minimum, and maximum MSE values over the full stream, together with the detailed MSE values before, at, and after the drift point. Here, $t_d$ denotes the drift point.}
\label{figtab:vis_agg_abrupt_rls}
\end{figure}

\paragraph{RLS-Based Performance under Abrupt Drift}
As shown in Fig.~\ref{figtab:vis_agg_abrupt_rls}, RLS$^{*}$, which represents RLS integrated with \modelname, achieves the strongest performance among all RLS-based variants on the abrupt-drift datasets. RLS$^{*}$ records the lowest average MSE of $0.908$, compared with $3.906$ for the standalone RLS model. It also obtains the lowest maximum MSE of $206.299$, while the standalone RLS reaches a higher maximum MSE of $368.385$. This reduction in both average and maximum error demonstrates that \modelname improves the robustness of RLS under sudden distributional changes.

The localized analysis around the drift point further confirms the advantage of RLS$^{*}$. At the drift point, RLS$^{*}$ obtains an MSE of $0.085$, whereas the standalone RLS model reaches a much larger MSE of $17.062$. The same pattern continues after the drift: RLS$^{*}$ maintains low errors of $0.711$, $0.720$, and $0.530$ at $t_d+1$, $t_d+2$, and $t_d+3$, respectively, while the original RLS records substantially higher errors of $31.809$, $72.850$, and $43.856$. The ADWIN- and KSWIN-based detector--adaptation variants also fail to provide a meaningful improvement over the original RLS model around the drift point. Their localized errors remain close to the standalone RLS behavior, indicating that the detector-triggered adaptations are not activated early enough to stabilize the model during the most critical transition region. In contrast, \modelname enables RLS to maintain substantially lower error before, during, and after the abrupt drift, demonstrating stronger drift resilience and faster post-drift recovery.

\begin{figure}[!htbp]
\centering

\includegraphics[
  width=\textwidth,
  keepaspectratio
]{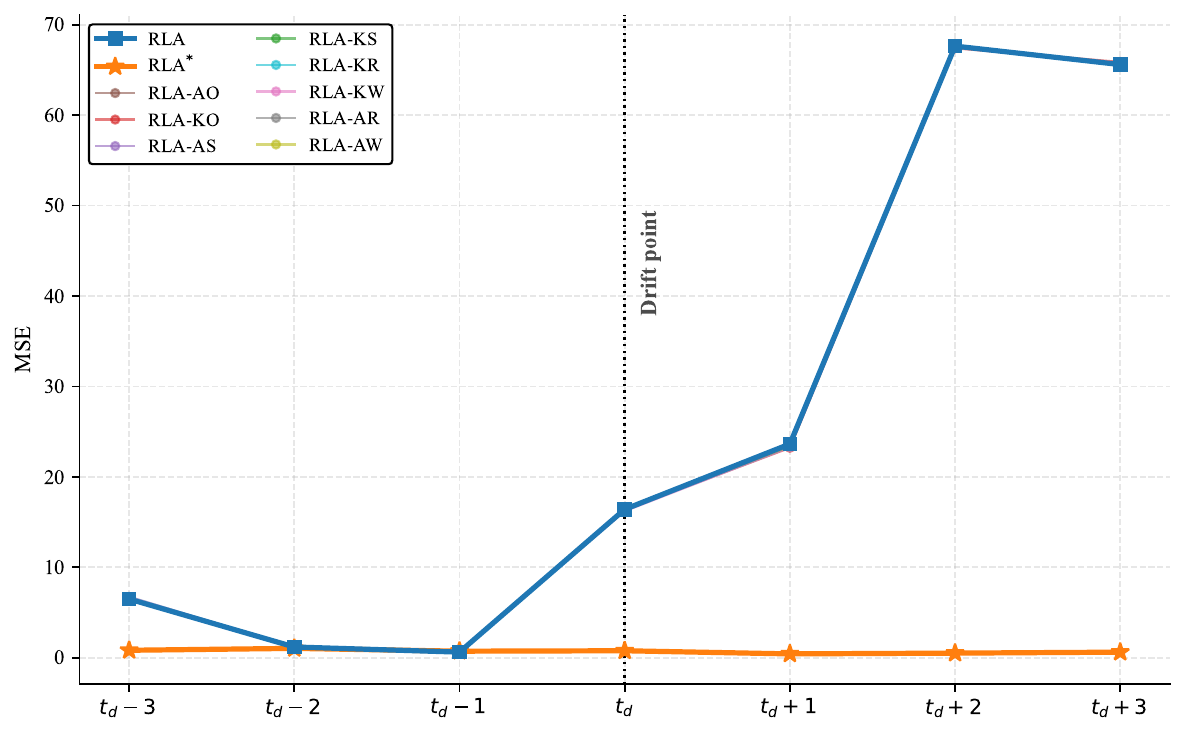}

\vspace{0.2cm}

\scriptsize
\resizebox{\textwidth}{!}{%
\begin{tabular}{lccc|ccccccc}
\toprule
\multirow{2}{*}{Method}
& \multicolumn{3}{c|}{MSE}
& \multicolumn{7}{c}{MSE Around Drift ($t_d$)} \\
\cmidrule(lr){2-4}
\cmidrule(lr){5-11}
& Avg & Min & Max
& $t_d\!-\!3$ & $t_d\!-\!2$ & $t_d\!-\!1$ & $t_d$
& $t_d\!+\!1$ & $t_d\!+\!2$ & $t_d\!+\!3$ \\
\midrule
LMS*          & \textbf{0.546} & \textbf{0.000} & \textbf{25.666} & \textbf{0.821} & \textbf{1.030} & 0.720 & \textbf{0.767} & \textbf{0.421} & \textbf{0.506} & \textbf{0.612} \\
LMS               & 4.668 & 0.001 & 555.706 & 6.488 & 1.184 & 0.616 & 16.408 & 23.632 & 67.640 & 65.597 \\
LMS-AO     & 4.669 & 0.001 & 557.063 & 6.503 & 1.201 & 0.612 & 16.386 & 23.381 & 67.714 & 65.638 \\
LMS-KO     & 4.670 & 0.001 & 558.261 & 6.512 & 1.217 & \textbf{0.609} & 16.379 & 23.223 & 67.412 & 65.867 \\
LMS-AS    & 4.680 & 0.001 & 555.694 & 6.487 & 1.184 & 0.615 & 16.409 & 23.634 & 67.640 & 65.598 \\
LMS-KS    & 4.693 & 0.001 & 555.830 & 6.459 & 1.196 & 0.621 & 16.391 & 23.580 & 67.640 & 65.566 \\
LMS-KR   & 5.096 & 0.000 & 554.752 & 6.715 & 1.190 & 0.652 & 16.267 & 23.327 & 67.525 & 65.637 \\
LMS-KW  & 5.254 & 0.000 & 554.717 & 6.718 & 1.187 & 0.653 & 16.216 & 23.301 & 67.555 & 65.623 \\
LMS-AR   & 5.265 & 0.001 & 554.847 & 6.469 & 1.192 & 0.637 & 16.376 & 23.413 & 67.654 & 65.627 \\
LMS-AW  & 5.265 & 0.001 & 554.847 & 6.469 & 1.192 & 0.637 & 16.376 & 23.413 & 67.654 & 65.627 \\
\bottomrule
\end{tabular}%
}

\caption{\scriptsize LMS variants aggregated performance across abrupt-drift experiments. The plot illustrates the MSE behavior around the drift point, while the table reports the aggregated average, minimum, and maximum MSE values over the full stream, together with the detailed MSE values before, at, and after the drift point. Here, $t_d$ denotes the drift point.}
\label{figtab:vis_agg_abrupt_lms}
\end{figure}

\paragraph{LMS-Based Performance under Abrupt Drift}
As shown in Fig.~\ref{figtab:vis_agg_abrupt_lms}, LMS$^{*}$, which represents Widrow-Hoff (LMS) integrated with \modelname, achieves the best performance among all LMS-based variants on the abrupt-drift datasets. It records the lowest average MSE of $0.546$, compared with $4.668$ for the standalone LMS model, and reduces the maximum MSE from $555.706$ to $25.666$. This indicates that \modelname substantially improves the stability of LMS under sudden distributional changes.

The localized results around the drift point further confirm this advantage. At the drift point, LMS$^{*}$ obtains an MSE of $0.767$, while the original LMS reaches $16.408$. After the drift, LMS$^{*}$ quickly maintains low errors of $0.421$, $0.506$, and $0.612$ at $t_d+1$, $t_d+2$, and $t_d+3$, respectively, whereas the standalone LMS remains much higher. The ADWIN- and KSWIN-based variants provide little improvement over the original LMS, mainly because delayed detection triggers adaptation too late to affect the critical drift region. Overall, \modelname enables LMS to achieve stronger abrupt-drift resilience and faster post-drift recovery.

\subsubsection{Performance Analysis on Incremental Datasets}

This section presents the aggregated performance of each method across all incremental-drift datasets listed in Table~\ref{tab:synthetic-datasets-properties}. Unlike abrupt drift, where a single sudden change occurs near the middle of the stream, incremental drift consists of consecutive low-magnitude changes distributed along the stream. Therefore, the drift point $t_d$ reported in the plots and tables represents the aggregation of all incremental transition points across all experiments and datasets, rather than one fixed drift location. This analysis evaluates how each method maintains performance under continuously evolving data distributions.

\begin{figure}[!htbp]
\centering

\includegraphics[
  width=\textwidth,
  keepaspectratio
]{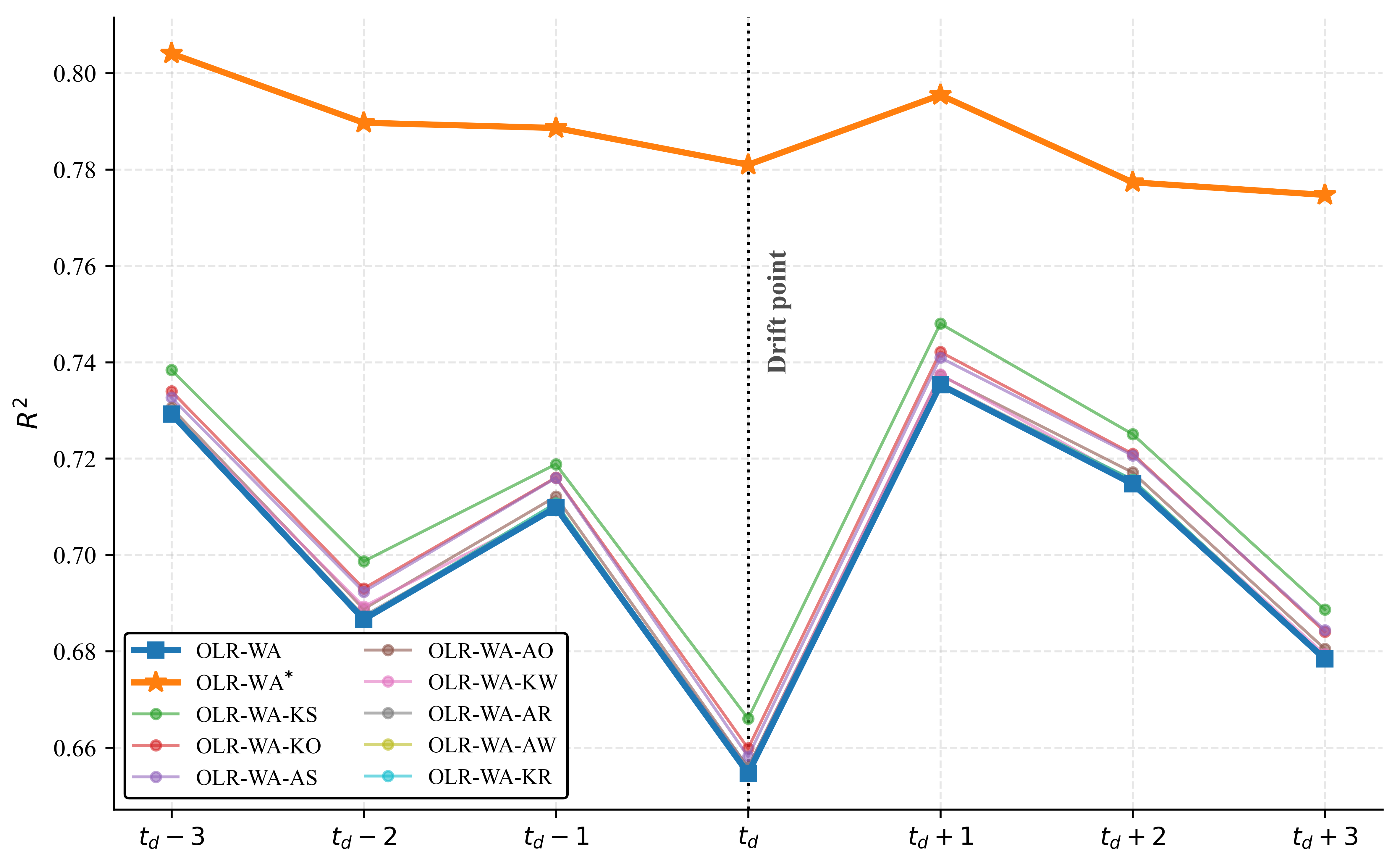}

\vspace{0.2cm}

\scriptsize
\resizebox{\textwidth}{!}{%
\begin{tabular}{lccc|ccc|ccccccc}
\toprule
\multirow{2}{*}{Method}
& \multicolumn{3}{c|}{$R^2$}
& \multicolumn{3}{c|}{MSE}
& \multicolumn{7}{c}{$R^2$ Around Drift ($t_d$)} \\
\cmidrule(lr){2-4}
\cmidrule(lr){5-7}
\cmidrule(lr){8-14}
& Avg & Min & Max
& Avg & Min & Max
& $t_d\!-\!3$ & $t_d\!-\!2$ & $t_d\!-\!1$ & $t_d$
& $t_d\!+\!1$ & $t_d\!+\!2$ & $t_d\!+\!3$ \\
\midrule
OLR-WA* & \textbf{0.745} & \textbf{0.019} & 0.962 & \textbf{1.734} & \textbf{0.749} & \textbf{3.228} & \textbf{0.804} & \textbf{0.790} & \textbf{0.789} & \textbf{0.781} & \textbf{0.795} & \textbf{0.777} & \textbf{0.775} \\
OLR-WA-KS   & 0.653 & -0.838 & 0.962 & 1.938 & 0.880 & 3.632 & 0.738 & 0.699 & 0.719 & 0.666 & 0.748 & 0.725 & 0.689 \\
OLR-WA-KO   & 0.649 & -0.838 & 0.962 & 1.970 & 0.952 & 3.632 & 0.734 & 0.693 & 0.716 & 0.660 & 0.742 & 0.721 & 0.684 \\
OLR-WA-AS   & 0.647 & -0.838 & 0.962 & 1.969 & 0.952 & 3.632 & 0.733 & 0.692 & 0.716 & 0.658 & 0.741 & 0.721 & 0.684 \\
OLR-WA-AO   & 0.646 & -0.838 & 0.962 & 1.988 & 0.952 & 3.632 & 0.730 & 0.689 & 0.712 & 0.656 & 0.737 & 0.717 & 0.680 \\
OLR-WA-KR   & 0.645 & -0.838 & 0.962 & 1.992 & 0.952 & 3.632 & 0.729 & 0.689 & 0.710 & 0.656 & 0.737 & 0.715 & 0.679 \\
OLR-WA-KW   & 0.645 & -0.779 & 0.962 & 1.994 & 0.952 & 3.632 & 0.729 & 0.687 & 0.710 & 0.656 & 0.736 & 0.716 & 0.678 \\
OLR-WA-AR   & 0.645 & -0.838 & 0.962 & 1.996 & 0.952 & 3.632 & 0.729 & 0.687 & 0.711 & 0.655 & 0.735 & 0.716 & 0.679 \\
OLR-WA-AW   & 0.645 & -0.838 & 0.962 & 1.996 & 0.952 & 3.632 & 0.729 & 0.687 & 0.711 & 0.655 & 0.735 & 0.716 & 0.679 \\
OLR-WA      & 0.645 & -0.838 & 0.962 & 1.998 & 0.952 & 3.632 & 0.729 & 0.687 & 0.710 & 0.655 & 0.735 & 0.715 & 0.678 \\
\bottomrule
\end{tabular}%
}

\caption{\scriptsize OLR-WA variants aggregated performance across incremental-drift experiments. The plot illustrates the $R^2$ behavior around the aggregated incremental transition points, while the table reports the aggregated average, minimum, and maximum $R^2$ and MSE values over the full stream, together with the detailed $R^2$ values before, at, and after the transition points. Here, $t_d$ denotes the aggregated incremental transition points.}
\label{figtab:vis_agg_incremental_olrwa}

\vspace{-.25cm}
\end{figure}

\paragraph{OLR-WA-Based Performance under Incremental Drift}
As shown in Fig.~\ref{figtab:vis_agg_incremental_olrwa}, OLR-WA$^{*}$, which represents OLR-WA integrated with \modelname, achieves the strongest performance among all OLR-WA-based variants on the incremental-drift datasets. It obtains the highest average $R^2$ of $0.745$ and the lowest average MSE of $1.734$, outperforming the standalone OLR-WA model, which records an average $R^2$ of $0.645$ and an average MSE of $1.998$. This indicates that \modelname improves the ability of OLR-WA to maintain stable predictive performance under continuously evolving distributions.

The localized results around the aggregated incremental transition points further support this finding. At $t_d$, OLR-WA$^{*}$ achieves an $R^2$ of $0.781$, compared with $0.655$ for the standalone OLR-WA model. It also maintains stronger performance before and after the transition points, with $R^2$ values of $0.804$, $0.790$, and $0.789$ before $t_d$, and $0.795$, $0.777$, and $0.775$ after $t_d$. Although some ADWIN- and KSWIN-based variants slightly improve over the base model, their gains remain limited, and none reaches the stability of OLR-WA$^{*}$. Overall, these results show that \modelname provides smoother adaptation to repeated low-magnitude changes and improves OLR-WA's resilience under incremental drift.

\begin{figure}[!htbp]
\centering

\includegraphics[
  width=\textwidth,
  keepaspectratio
]{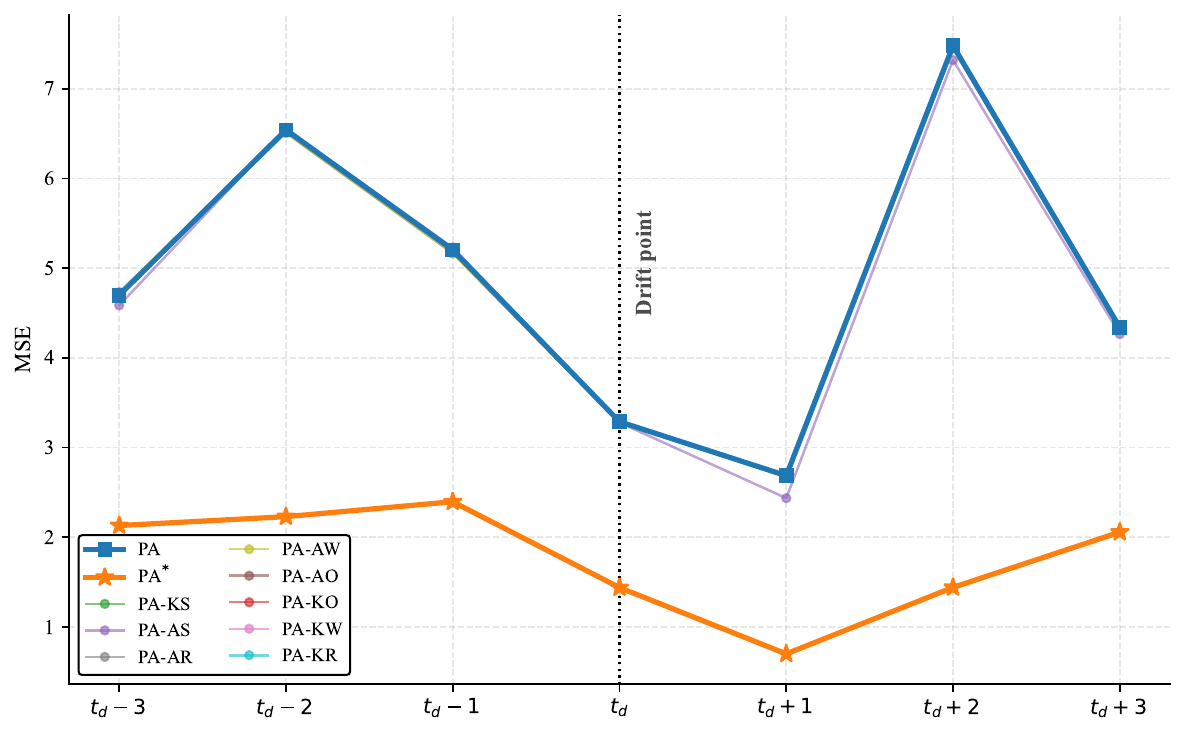}

\vspace{0.2cm}

\scriptsize
\resizebox{\textwidth}{!}{%
\begin{tabular}{lccc|ccccccc}
\toprule
\multirow{2}{*}{Method}
& \multicolumn{3}{c|}{MSE}
& \multicolumn{7}{c}{MSE Around Drift ($t_d$)} \\
\cmidrule(lr){2-4}
\cmidrule(lr){5-11}
& Avg & Min & Max
& $t_d\!-\!3$ & $t_d\!-\!2$ & $t_d\!-\!1$ & $t_d$
& $t_d\!+\!1$ & $t_d\!+\!2$ & $t_d\!+\!3$ \\
\midrule
PA*          & \textbf{1.424} & \textbf{0.000} & \textbf{85.047} & \textbf{2.129} & \textbf{2.228} & \textbf{2.395} & \textbf{1.437} & \textbf{0.698} & \textbf{1.439} & \textbf{2.055} \\
PA-KS    & 4.469 & 0.000 & 85.047 & 4.697 & 6.541 & 5.202 & 3.287 & 2.685 & 7.479 & 4.339 \\
PA-AS    & 4.503 & 0.000 & 85.047 & 4.584 & 6.549 & 5.229 & 3.277 & 2.434 & 7.319 & 4.264 \\
PA-AR   & 4.508 & 0.000 & 85.047 & 4.730 & 6.508 & 5.165 & 3.278 & 2.673 & 7.450 & 4.332 \\
PA-AW  & 4.508 & 0.000 & 85.047 & 4.730 & 6.508 & 5.165 & 3.278 & 2.673 & 7.450 & 4.332 \\
PA               & 4.510 & 0.000 & 85.047 & 4.697 & 6.541 & 5.202 & 3.287 & 2.685 & 7.479 & 4.339 \\
PA-AO     & 4.510 & 0.000 & 85.047 & 4.698 & 6.541 & 5.201 & 3.287 & 2.689 & 7.482 & 4.340 \\
PA-KO     & 4.513 & 0.000 & 85.047 & 4.697 & 6.541 & 5.202 & 3.287 & 2.685 & 7.479 & 4.339 \\
PA-KW  & 4.575 & 0.000 & 85.047 & 4.697 & 6.541 & 5.202 & 3.287 & 2.685 & 7.479 & 4.339 \\
PA-KR   & 4.733 & 0.000 & 85.047 & 4.697 & 6.541 & 5.202 & 3.287 & 2.685 & 7.479 & 4.339 \\
\bottomrule
\end{tabular}%
}

\caption{\scriptsize PA variants aggregated performance across incremental-drift experiments. The plot illustrates the MSE behavior around the aggregated incremental transition points, while the table reports the aggregated average, minimum, and maximum MSE values over the full stream, together with the detailed MSE values before, at, and after the transition points. Here, $t_d$ denotes the aggregated incremental transition points.}
\label{figtab:vis_agg_PA_incremental}
\end{figure}

\paragraph{PA-Based Performance under Incremental Drift}
As shown in Fig.~\ref{figtab:vis_agg_PA_incremental}, PA$^{*}$, which represents PA integrated with \modelname, achieves the best performance among all PA-based variants on the incremental-drift datasets. It records the lowest average MSE of $1.424$, while the closest detector--adaptation variant, PA-KS, records a much higher average MSE of $4.469$. This indicates that \modelname provides a clear improvement in PA's ability to handle continuously evolving data distributions.

The localized results around the aggregated incremental transition points further support this improvement. At $t_d$, PA$^{*}$ obtains an MSE of $1.437$, compared with $3.287$ for PA-KS and higher values for the remaining ADWIN- and KSWIN-based variants. PA$^{*}$ also maintains lower errors after the transition points, with MSE values of $0.698$, $1.439$, and $2.055$ at $t_d+1$, $t_d+2$, and $t_d+3$, respectively. In contrast, the detector--adaptation baselines show larger fluctuations and higher post-transition errors. Overall, these results show that \modelname enables PA to respond more smoothly to repeated low-magnitude incremental changes, while conventional detector-triggered adaptation remains less effective under gradual continuous transitions.

\begin{figure}[!htbp]
\centering

\includegraphics[
  width=\textwidth,
  keepaspectratio
]{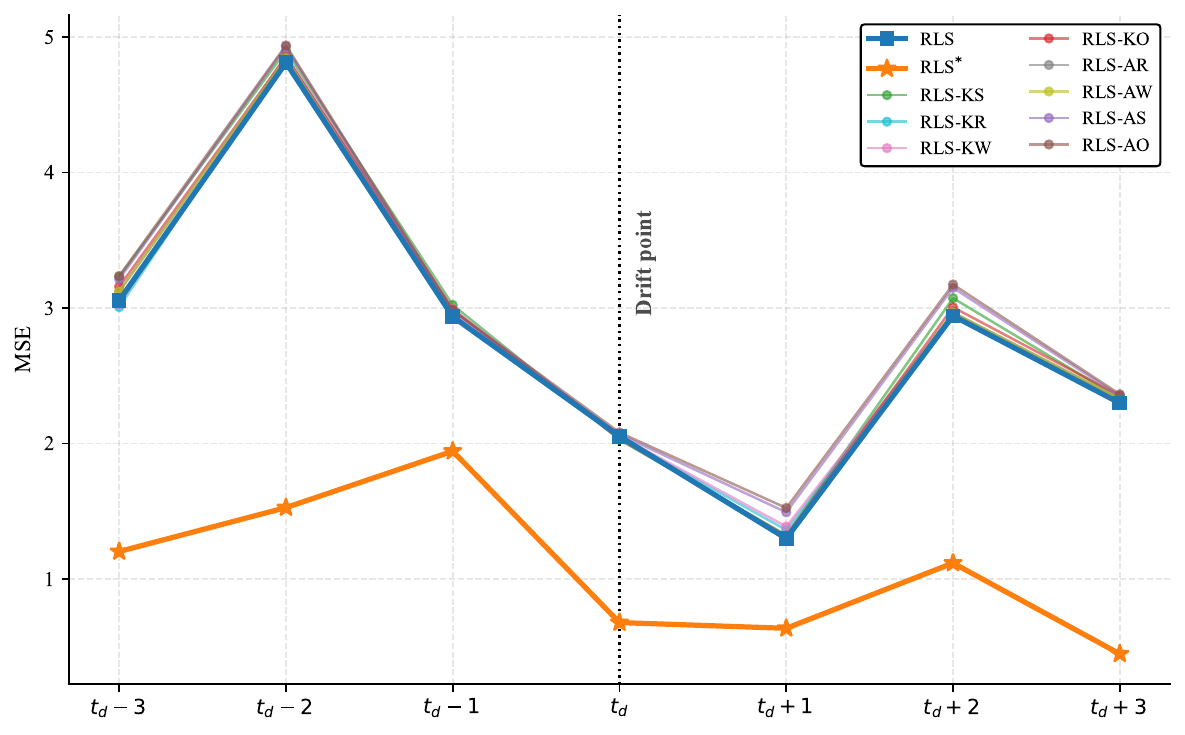}

\vspace{0.2cm}

\scriptsize
\resizebox{\textwidth}{!}{%
\begin{tabular}{lccc|ccccccc}
\toprule
\multirow{2}{*}{Method}
& \multicolumn{3}{c|}{MSE}
& \multicolumn{7}{c}{MSE Around Drift ($t_d$)} \\
\cmidrule(lr){2-4}
\cmidrule(lr){5-11}
& Avg & Min & Max
& $t_d\!-\!3$ & $t_d\!-\!2$ & $t_d\!-\!1$ & $t_d$
& $t_d\!+\!1$ & $t_d\!+\!2$ & $t_d\!+\!3$ \\
\midrule
RLS*          & \textbf{0.813} & 0.000 & \textbf{84.836} & \textbf{1.201} & \textbf{1.526} & \textbf{1.942} & \textbf{0.677} & \textbf{0.634} & \textbf{1.116} & \textbf{0.445} \\
RLS-KS    & 2.683 & 0.000 & 84.836 & 3.229 & 4.900 & 3.022 & 2.026 & 1.322 & 3.072 & 2.333 \\
RLS-KR  & 2.685 & 0.001 & 84.836 & 3.005 & 4.852 & 2.986 & 2.048 & 1.360 & 2.964 & 2.318 \\
RLS-KW & 2.694 & 0.001 & 84.836 & 3.028 & 4.867 & 2.992 & 2.054 & 1.386 & 2.955 & 2.322 \\
RLS-KO     & 2.695 & 0.001 & 84.836 & 3.157 & 4.855 & 2.987 & 2.032 & 1.317 & 3.005 & 2.352 \\
RLS-AR   & 2.696 & 0.001 & 84.836 & 3.122 & 4.844 & 2.941 & 2.059 & 1.306 & 2.966 & 2.331 \\
RLS-AW  & 2.696 & 0.001 & 84.836 & 3.122 & 4.844 & 2.941 & 2.059 & 1.306 & 2.966 & 2.331 \\
RLS               & 2.698 & 0.001 & 84.836 & 3.056 & 4.810 & 2.935 & 2.051 & 1.297 & 2.941 & 2.297 \\
RLS-AS    & 2.705 & 0.001 & 84.836 & 3.212 & 4.930 & 2.965 & 2.066 & 1.491 & 3.149 & 2.350 \\
RLS-AO     & 2.706 & 0.001 & 84.836 & 3.236 & 4.938 & 2.979 & 2.080 & 1.523 & 3.173 & 2.362 \\
\bottomrule
\end{tabular}%
}

\caption{\scriptsize RLS variants aggregated performance across incremental-drift experiments. The plot illustrates the MSE behavior around the aggregated incremental transition points, while the table reports the aggregated average, minimum, and maximum MSE values over the full stream, together with the detailed MSE values before, at, and after the transition points. Here, $t_d$ denotes the aggregated incremental transition points.}
\label{figtab:vis_agg_RLS_incremental}
\end{figure}

\paragraph{RLS-Based Performance under Incremental Drift}
As shown in Fig.~\ref{figtab:vis_agg_RLS_incremental}, RLS$^{*}$, which represents RLS integrated with \modelname, achieves the best performance among all RLS-based variants on the incremental-drift datasets. It records the lowest average MSE of $0.813$, compared with $2.698$ for the standalone RLS model. This indicates that \modelname improves the ability of RLS to maintain stable predictive performance under consecutive low-magnitude distributional changes.

The localized results around the aggregated incremental transition points further confirm this advantage. At $t_d$, RLS$^{*}$ obtains an MSE of $0.677$, while the original RLS records a higher MSE of $2.051$. RLS$^{*}$ also maintains lower post-transition errors of $0.634$, $1.116$, and $0.445$ at $t_d+1$, $t_d+2$, and $t_d+3$, respectively, compared with $1.297$, $2.941$, and $2.297$ for the standalone RLS model. The ADWIN- and KSWIN-based variants provide only limited improvement over the base RLS model, with average MSE values remaining close to the standalone result. Overall, \modelname enables RLS to adapt more smoothly to repeated incremental changes and maintain lower error around the aggregated transition points.

\begin{figure}[!htbp]
\centering

\includegraphics[
  width=\textwidth,
  keepaspectratio
]{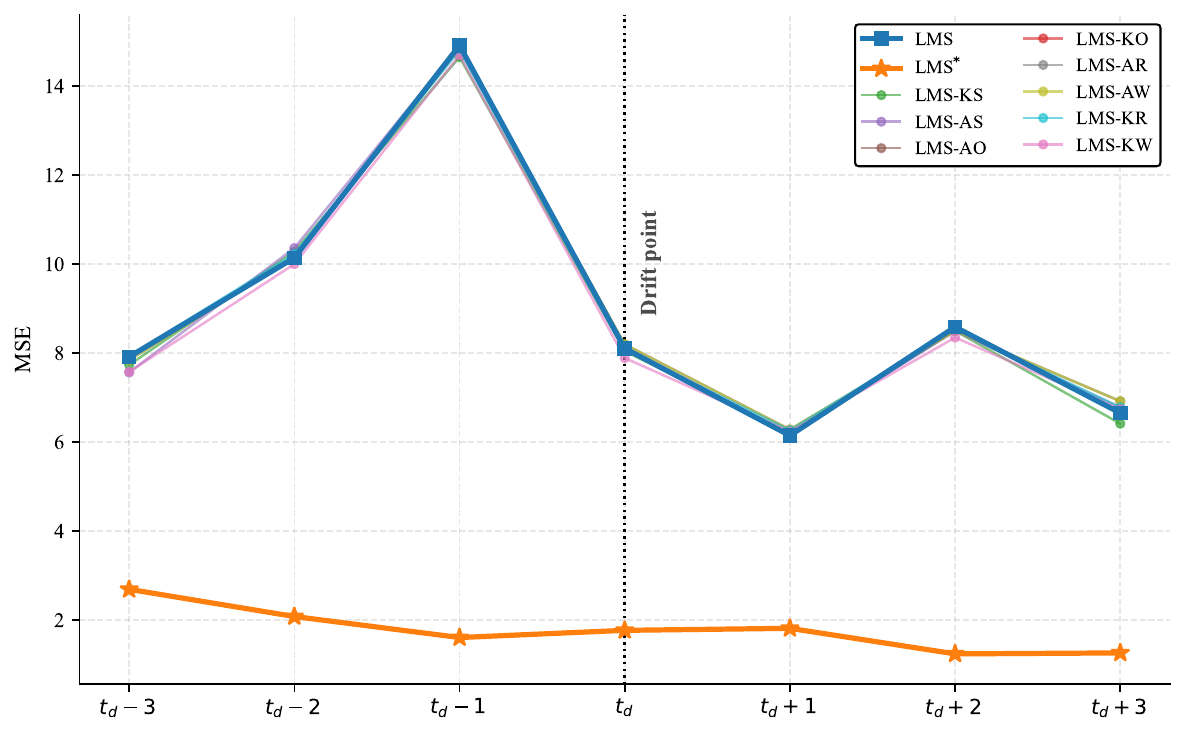}

\vspace{0.2cm}

\scriptsize
\resizebox{\textwidth}{!}{%
\begin{tabular}{lccc|ccccccc}
\toprule
\multirow{2}{*}{Method}
& \multicolumn{3}{c|}{MSE}
& \multicolumn{7}{c}{MSE Around Drift ($t_d$)} \\
\cmidrule(lr){2-4}
\cmidrule(lr){5-11}
& Avg & Min & Max
& $t_d\!-\!3$ & $t_d\!-\!2$ & $t_d\!-\!1$ & $t_d$
& $t_d\!+\!1$ & $t_d\!+\!2$ & $t_d\!+\!3$ \\
\midrule
LMS*          & \textbf{0.536} & \textbf{0.000} & \textbf{29.057} & \textbf{2.689} & \textbf{2.077} & \textbf{1.605} & \textbf{1.767} & \textbf{1.812} & \textbf{1.237} & \textbf{1.259} \\
LMS-KS    & 3.508 & 0.000 & 86.626 & 7.733 & 10.284 & 14.655 & 8.046 & 6.201 & 8.501 & 6.413 \\
LMS-AS    & 3.517 & 0.002 & 86.626 & 7.564 & 10.361 & 14.903 & 8.114 & 6.159 & 8.496 & 6.777 \\
LMS               & 3.522 & 0.002 & 86.626 & 7.916 & 10.143 & 14.917 & 8.105 & 6.146 & 8.587 & 6.653 \\
LMS-AO     & 3.524 & 0.001 & 86.626 & 7.912 & 10.138 & 14.915 & 8.099 & 6.149 & 8.581 & 6.649 \\
LMS-KO     & 3.528 & 0.001 & 86.626 & 7.909 & 10.139 & 14.910 & 8.101 & 6.146 & 8.583 & 6.649 \\
LMS-AR   & 3.528 & 0.002 & 86.626 & 7.844 & 10.146 & 14.852 & 8.197 & 6.281 & 8.502 & 6.918 \\
LMS-AW  & 3.528 & 0.002 & 86.626 & 7.844 & 10.146 & 14.852 & 8.197 & 6.281 & 8.502 & 6.918 \\
LMS-KR   & 4.013 & 0.000 & 86.626 & 7.931 & 10.229 & 14.808 & 8.085 & 6.263 & 8.543 & 6.780 \\
LMS-KW  & 4.030 & 0.000 & 86.626 & 7.578 & 10.000 & 14.699 & 7.890 & 6.235 & 8.350 & 6.739 \\
\bottomrule
\end{tabular}%
}

\caption{\scriptsize LMS variants aggregated performance across incremental-drift experiments. The plot illustrates the MSE behavior around the aggregated incremental transition points, while the table reports the aggregated average, minimum, and maximum MSE values over the full stream, together with the detailed MSE values before, at, and after the transition points. Here, $t_d$ denotes the aggregated incremental transition points.}
\label{figtab:vis_agg_LMS_incremental}
\end{figure}

\paragraph{LMS-Based Performance under Incremental Drift}
As shown in Fig.~\ref{figtab:vis_agg_LMS_incremental}, LMS$^{*}$, which represents Widrow-Hoff (LMS) integrated with \modelname, achieves the best performance among all LMS-based variants on the incremental-drift datasets. It records the lowest average MSE of $0.536$, compared with $3.522$ for the standalone LMS model, and reduces the maximum MSE from $86.626$ to $29.057$. This shows that \modelname substantially improves the stability of LMS under continuously evolving data distributions.

The localized results around the aggregated incremental transition points further confirm this improvement. At $t_d$, LMS$^{*}$ obtains an MSE of $1.767$, while the standalone LMS records a much higher MSE of $8.105$. LMS$^{*}$ also maintains lower errors after the transition points, with MSE values of $1.812$, $1.237$, and $1.259$ at $t_d+1$, $t_d+2$, and $t_d+3$, respectively. In contrast, the ADWIN- and KSWIN-based variants remain close to the original LMS behavior and provide only limited improvement. Overall, \modelname enables LMS to adapt more smoothly to repeated incremental changes and maintain lower error around the aggregated transition points.

\subsubsection{Performance Analysis on Alternating Gradual Datasets}

This section presents the aggregated performance of each method across all alternating-gradual-drift datasets listed in Table~\ref{tab:synthetic-datasets-properties}. In the alternating-gradual setting, the stream starts from a stable concept $C_1$ and eventually stabilizes under a new concept $C_2$. However, the transition does not occur as a smooth incremental  shift. Instead, the stream alternates between samples from the old and new concepts for a period of time before fully settling on $C_2$. Therefore, the drift-localized values reported in the plots and tables represent aggregated results over all transition points from all alternating-gradual-drift experiments and datasets, rather than a single fixed drift location. This analysis evaluates how each method performs when old and new concepts coexist during the transition period.

\begin{figure}[!htbp]
\centering

\includegraphics[
  width=\textwidth,
  keepaspectratio
]{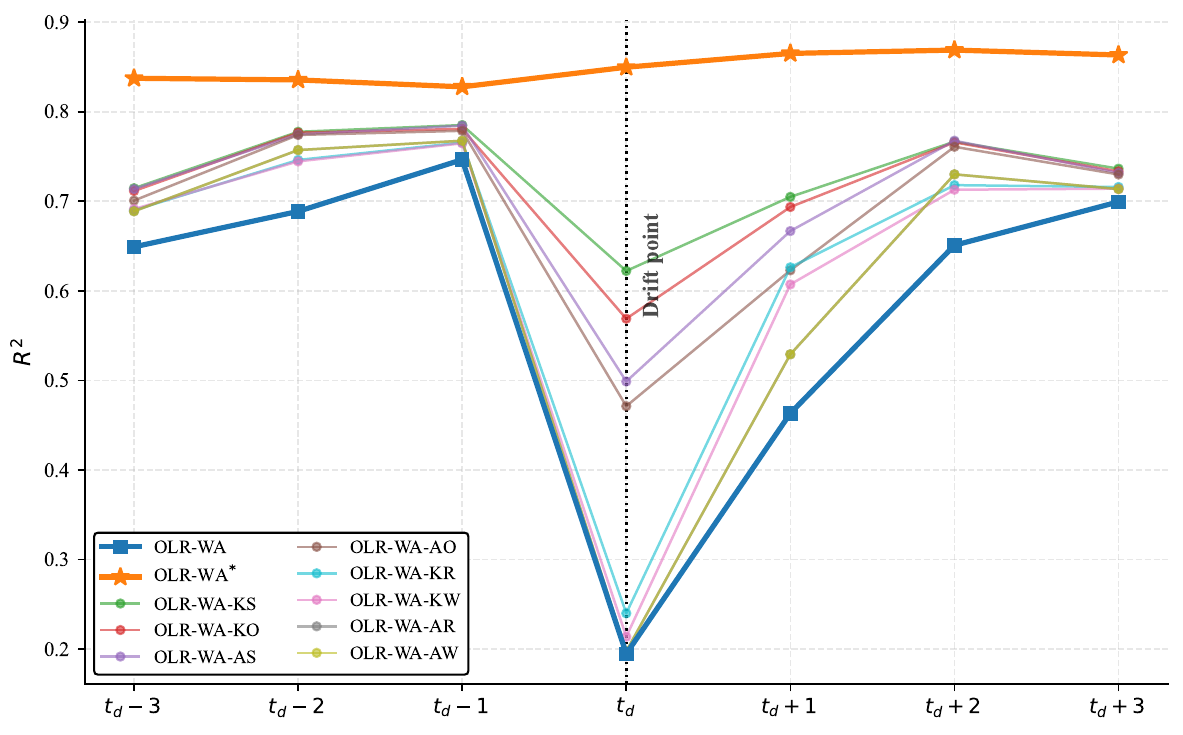}

\vspace{0.2cm}

\scriptsize
\resizebox{\textwidth}{!}{%
\begin{tabular}{lccc|ccc|ccccccc}
\toprule
\multirow{2}{*}{Method}
& \multicolumn{3}{c|}{$R^2$}
& \multicolumn{3}{c|}{MSE}
& \multicolumn{7}{c}{$R^2$ Around Drift ($t_d$)} \\
\cmidrule(lr){2-4}
\cmidrule(lr){5-7}
\cmidrule(lr){8-14}
& Avg & Min & Max
& Avg & Min & Max
& $t_d\!-\!3$ & $t_d\!-\!2$ & $t_d\!-\!1$ & $t_d$
& $t_d\!+\!1$ & $t_d\!+\!2$ & $t_d\!+\!3$ \\
\midrule
OLR-WA* & \textbf{0.797} & \textbf{-0.015} & 0.963 & \textbf{1.855} & \textbf{0.651} & \textbf{4.677} & \textbf{0.837} & \textbf{0.836} & \textbf{0.828} & \textbf{0.850} & \textbf{0.865} & \textbf{0.869} & \textbf{0.863} \\
OLR-WA-KS & 0.697 & -0.871 & 0.963 & 2.243 & 0.829 & 27.258 & 0.715 & 0.778 & 0.785 & 0.622 & 0.705 & 0.767 & 0.736 \\
OLR-WA-KO  & 0.690 & -1.653 & 0.963 & 2.271 & 0.829 & 26.941 & 0.712 & 0.777 & 0.781 & 0.569 & 0.694 & 0.766 & 0.733 \\
OLR-WA-AS & 0.682 & -1.653 & 0.963 & 2.337 & 0.829 & 27.258 & 0.714 & 0.775 & 0.784 & 0.499 & 0.667 & 0.768 & 0.732 \\
OLR-WA-AO  & 0.673 & -1.653 & 0.963 & 2.399 & 0.829 & 35.525 & 0.701 & 0.774 & 0.779 & 0.471 & 0.623 & 0.761 & 0.730 \\
OLR-WA-KR & 0.646 & -2.961 & 0.963 & 2.633 & 0.829 & 32.912 & 0.690 & 0.746 & 0.766 & 0.240 & 0.626 & 0.718 & 0.716 \\
OLR-WA-KW & 0.641 & -2.961 & 0.963 & 2.682 & 0.829 & 32.912 & 0.691 & 0.745 & 0.765 & 0.214 & 0.607 & 0.713 & 0.714 \\
OLR-WA-AR & 0.635 & -2.959 & 0.963 & 2.736 & 0.829 & 37.860 & 0.689 & 0.757 & 0.768 & 0.197 & 0.529 & 0.730 & 0.714 \\
OLR-WA-AW & 0.635 & -2.959 & 0.963 & 2.736 & 0.829 & 37.860 & 0.689 & 0.757 & 0.768 & 0.197 & 0.529 & 0.730 & 0.714 \\
OLR-WA & 0.622 & -2.953 & 0.963 & 3.093 & 0.829 & 38.533 & 0.649 & 0.689 & 0.747 & 0.195 & 0.463 & 0.651 & 0.699 \\
\bottomrule
\end{tabular}%
}

\caption{\scriptsize OLR-WA variants aggregated performance across alternating-gradual-drift experiments. The plot illustrates the aggregated $R^2$ behavior around the alternating-gradual transition points, while the table reports the aggregated average, minimum, and maximum $R^2$ and MSE values over the full stream, together with the detailed $R^2$ values before, at, and after the transition points. Here, $t_d$ denotes the aggregated alternating-gradual transition points.}
\label{figtab:vis_agg_gradual_olrwa}

\vspace{-.25cm}
\end{figure}

\paragraph{OLR-WA-Based Performance under Alternating Gradual Drift}
As shown in Fig.~\ref{figtab:vis_agg_gradual_olrwa}, OLR-WA$^{*}$, which represents OLR-WA integrated with \modelname, achieves the strongest overall performance among all OLR-WA-based variants on the alternating-gradual-drift datasets. It obtains the highest average $R^2$ value of $0.797$ and the lowest average MSE of $1.855$, clearly outperforming the standalone OLR-WA model, which records an average $R^2$ of $0.622$ and an average MSE of $3.093$. These results indicate that \modelname improves the robustness of OLR-WA when the stream transitions through alternating old and new concepts before stabilizing.

The localized analysis around the aggregated alternating-gradual transition points further confirms this advantage. At $t_d$, OLR-WA$^{*}$ maintains a high $R^2$ of $0.850$, whereas the original OLR-WA drops to $0.195$. OLR-WA$^{*}$ also preserves strong performance throughout the surrounding transition region, with $R^2$ values remaining between $0.828$ and $0.869$ from $t_d-3$ to $t_d+3$. Although several ADWIN- and KSWIN-based detector--adaptation variants improve over the base model, none matches the consistency of OLR-WA$^{*}$. For example, the strongest competing variant, OLR-WA-KS, reaches an average $R^2$ of $0.697$ and an $R^2$ of $0.622$ at $t_d$, both substantially below OLR-WA$^{*}$. Overall, these results show that \modelname enables OLR-WA to adapt more smoothly and maintain stronger predictive stability during alternating gradual drift.

\begin{figure}[!htbp]
\centering

\includegraphics[
  width=\textwidth,
  keepaspectratio
]{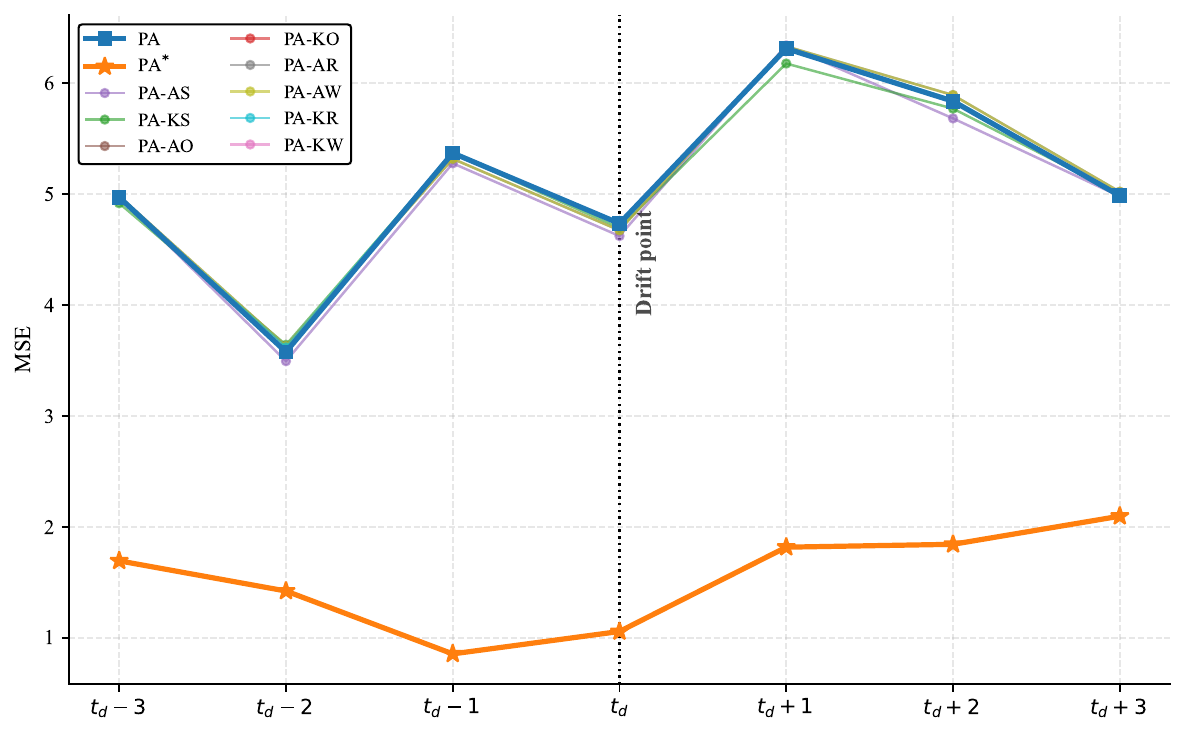}

\vspace{0.2cm}

\scriptsize
\resizebox{\textwidth}{!}{%
\begin{tabular}{lccc|ccccccc}
\toprule
\multirow{2}{*}{Method}
& \multicolumn{3}{c|}{MSE}
& \multicolumn{7}{c}{MSE Around Drift ($t_d$)} \\
\cmidrule(lr){2-4}
\cmidrule(lr){5-11}
& Avg & Min & Max
& $t_d\!-\!3$ & $t_d\!-\!2$ & $t_d\!-\!1$ & $t_d$
& $t_d\!+\!1$ & $t_d\!+\!2$ & $t_d\!+\!3$ \\
\midrule
PA*          & \textbf{1.532} & 0.000 & \textbf{148.260} & \textbf{1.693} & \textbf{1.421} & \textbf{0.854} & \textbf{1.056} & \textbf{1.816} & \textbf{1.842} & \textbf{2.096} \\
PA-AS    & 6.999 & 0.001 & 651.252 & 4.976 & \textbf{3.493} & 5.277 & 4.621 & 6.340 & \textbf{5.682} & \textbf{4.979} \\
PA-KS    & 7.012 & 0.000 & 651.253 & \textbf{4.917} & 3.640 & 5.368 & 4.694 & \textbf{6.176} & 5.770 & 5.000 \\
PA               & 7.034 & 0.001 & 651.253 & 4.970 & 3.577 & 5.370 & 4.734 & 6.311 & 5.837 & 4.985 \\
PA-AO     & 7.035 & 0.001 & 651.253 & 4.970 & 3.579 & 5.372 & 4.737 & 6.311 & 5.839 & 4.984 \\
PA-KO     & 7.037 & 0.000 & 651.253 & 4.970 & 3.577 & 5.370 & 4.734 & 6.311 & 5.837 & 4.985 \\
PA-AR   & 7.130 & 0.001 & 651.191 & 4.982 & 3.629 & \textbf{5.317} & \textbf{4.673} & 6.331 & 5.892 & 5.020 \\
PA-AW  & 7.130 & 0.001 & 651.191 & 4.982 & 3.629 & \textbf{5.317} & \textbf{4.673} & 6.331 & 5.892 & 5.020 \\
PA-KR   & 7.287 & 0.001 & 651.253 & 4.967 & 3.618 & 5.379 & 4.735 & 6.317 & 5.850 & 4.986 \\
PA-KW  & 7.291 & 0.001 & 651.253 & 4.970 & 3.577 & 5.370 & 4.734 & 6.311 & 5.837 & 4.985 \\
\bottomrule
\end{tabular}%
}

\caption{\scriptsize PA variants aggregated performance across alternating-gradual-drift experiments. The plot illustrates the MSE behavior around the aggregated alternating-gradual transition points, while the table reports the aggregated average, minimum, and maximum MSE values over the full stream, together with the detailed MSE values before, at, and after the transition points. Here, $t_d$ denotes the aggregated alternating-gradual transition points.}
\label{figtab:vis_agg_PA_Gradual}
\end{figure}

\paragraph{PA-Based Performance under Alternating Gradual Drift}
As shown in Fig.~\ref{figtab:vis_agg_PA_Gradual}, PA$^{*}$, which represents PA integrated with \modelname, achieves the best performance among all PA-based variants on the alternating-gradual-drift datasets. It records the lowest average MSE of $1.532$, compared with $7.034$ for the standalone PA model, and also reduces the maximum MSE from $651.253$ to $148.260$. This indicates that \modelname substantially improves the robustness of PA when the stream alternates between old and new concepts before stabilizing.

The localized results around the aggregated alternating-gradual transition points further confirm this advantage. At $t_d$, PA$^{*}$ obtains an MSE of $1.056$, while the original PA records $4.734$. PA$^{*}$ also maintains lower errors after the transition points, with MSE values of $1.816$, $1.842$, and $2.096$ at $t_d+1$, $t_d+2$, and $t_d+3$, respectively, compared with $6.311$, $5.837$, and $4.985$ for the standalone PA model. Although some ADWIN- and KSWIN-based variants show slight improvements over the base model at individual positions, none approaches the overall stability of PA$^{*}$. Overall, these results show that \modelname enables PA to adapt more effectively and maintain lower error under alternating gradual drift.

\begin{figure}[!htbp]
\centering

\includegraphics[
  width=\textwidth,
  keepaspectratio
]{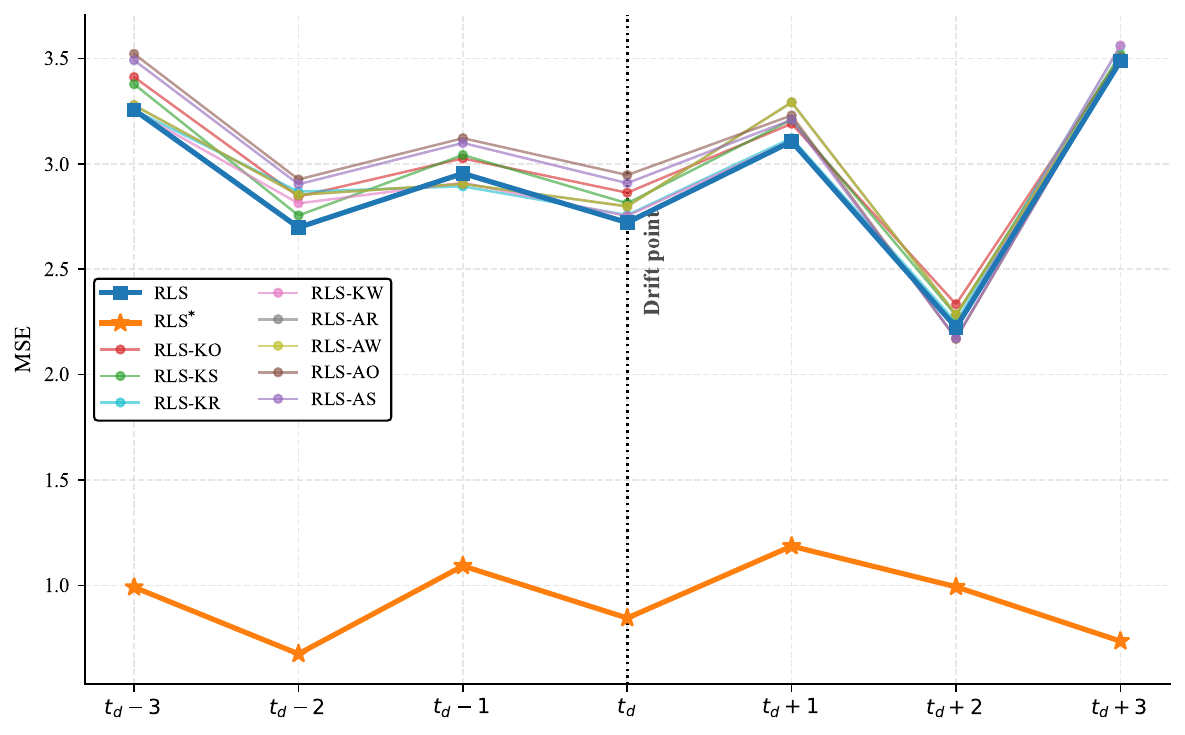}

\vspace{0.2cm}

\scriptsize
\resizebox{\textwidth}{!}{%
\begin{tabular}{lccc|ccccccc}
\toprule
\multirow{2}{*}{Method}
& \multicolumn{3}{c|}{MSE}
& \multicolumn{7}{c}{MSE Around Drift ($t_d$)} \\
\cmidrule(lr){2-4}
\cmidrule(lr){5-11}
& Avg & Min & Max
& $t_d\!-\!3$ & $t_d\!-\!2$ & $t_d\!-\!1$ & $t_d$
& $t_d\!+\!1$ & $t_d\!+\!2$ & $t_d\!+\!3$ \\
\midrule
RLS*          & \textbf{1.227} & 0.000 & \textbf{389.751} & \textbf{0.992} & \textbf{0.676} & \textbf{1.093} & \textbf{0.846} & \textbf{1.187} & \textbf{0.994} & \textbf{0.736} \\
RLS-KO     & 5.621 & \textbf{0.000} & 708.314 & 3.411 & 2.845 & 3.026 & 2.863 & 3.190 & 2.334 & 3.506 \\
RLS-KS    & 5.732 & 0.001 & 712.035 & 3.378 & 2.755 & 3.042 & 2.813 & 3.210 & 2.281 & 3.520 \\
RLS-KR   & 5.814 & 0.000 & 711.807 & 3.255 & 2.868 & 2.893 & 2.757 & 3.122 & 2.249 & 3.558 \\
RLS-KW  & 5.840 & 0.000 & 711.809 & 3.251 & 2.812 & 2.910 & 2.752 & 3.112 & 2.229 & 3.562 \\
RLS-AR   & 6.083 & 0.001 & 712.250 & 3.278 & 2.853 & 2.905 & 2.798 & 3.291 & 2.285 & 3.486 \\
RLS-AW  & 6.083 & 0.001 & 712.250 & 3.278 & 2.853 & 2.905 & 2.798 & 3.291 & 2.285 & 3.486 \\
RLS-AO     & 6.091 & 0.001 & 711.923 & 3.521 & 2.926 & 3.122 & 2.947 & 3.230 & 2.169 & 3.501 \\
RLS-AS    & 6.120 & 0.000 & 711.943 & 3.491 & 2.903 & 3.099 & 2.908 & 3.209 & 2.174 & 3.499 \\
RLS               & 7.163 & 0.001 & 712.390 & 3.256 & 2.698 & 2.954 & 2.722 & 3.106 & 2.224 & 3.487 \\
\bottomrule
\end{tabular}%
}

\caption{\scriptsize RLS variants aggregated performance across alternating-gradual-drift experiments. The plot illustrates the MSE behavior around the aggregated alternating-gradual transition points, while the table reports the aggregated average, minimum, and maximum MSE values over the full stream, together with the detailed MSE values before, at, and after the transition points. Here, $t_d$ denotes the aggregated alternating-gradual transition points.}
\label{figtab:vis_agg_RLS_Gradual}
\end{figure}

\paragraph{RLS-Based Performance under Alternating Gradual Drift}
As shown in Fig.~\ref{figtab:vis_agg_RLS_Gradual}, RLS$^{*}$, which represents RLS integrated with \modelname, achieves the best performance among all RLS-based variants on the alternating-gradual-drift datasets. It records the lowest average MSE of $1.227$, compared with $7.163$ for the standalone RLS model, and also reduces the maximum MSE from $712.390$ to $389.751$. This indicates that \modelname substantially improves the robustness of RLS when the stream alternates between old and new concepts before stabilizing.

The localized results around the aggregated alternating-gradual transition points further confirm this advantage. At $t_d$, RLS$^{*}$ obtains an MSE of $0.846$, while the original RLS records $2.722$. RLS$^{*}$ also maintains lower errors after the transition points, with MSE values of $1.187$, $0.994$, and $0.736$ at $t_d+1$, $t_d+2$, and $t_d+3$, respectively, compared with $3.106$, $2.224$, and $3.487$ for the standalone RLS model. Although several ADWIN- and KSWIN-based variants improve over the base RLS model, none approaches the overall stability of RLS$^{*}$. Overall, these results show that \modelname enables RLS to adapt more effectively and maintain lower error under alternating gradual drift.

\begin{figure}[!htbp]
\centering

\includegraphics[
  width=\textwidth,
  keepaspectratio
]{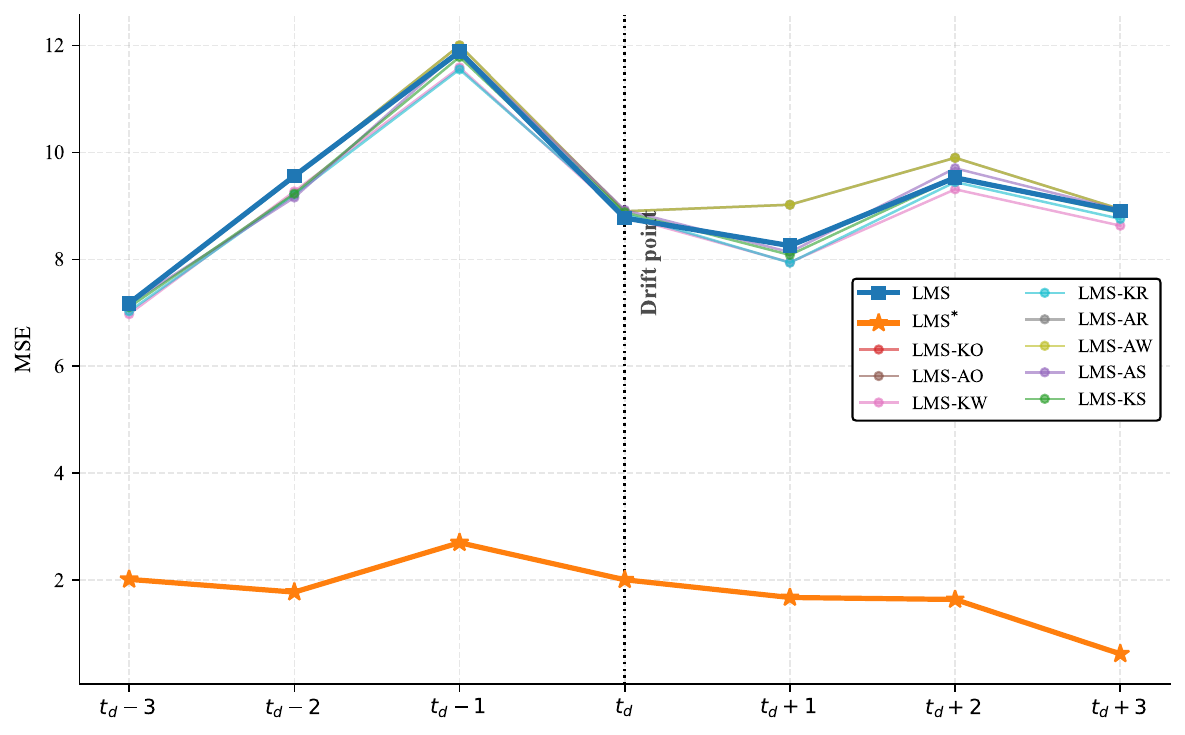}

\vspace{0.2cm}

\scriptsize
\resizebox{\textwidth}{!}{%
\begin{tabular}{lccc|ccccccc}
\toprule
\multirow{2}{*}{Method}
& \multicolumn{3}{c|}{MSE}
& \multicolumn{7}{c}{MSE Around Drift ($t_d$)} \\
\cmidrule(lr){2-4}
\cmidrule(lr){5-11}
& Avg & Min & Max
& $t_d\!-\!3$ & $t_d\!-\!2$ & $t_d\!-\!1$ & $t_d$
& $t_d\!+\!1$ & $t_d\!+\!2$ & $t_d\!+\!3$ \\
\midrule
LMS*          & \textbf{0.967} & \textbf{0.000} & \textbf{171.564} & \textbf{2.017} & \textbf{1.778} & \textbf{2.703} & \textbf{2.009} & \textbf{1.678} & \textbf{1.641} & \textbf{0.623} \\
LMS-KO     & 7.230 & 0.001 & 640.767 & 7.180 & 9.554 & 11.876 & 8.773 & 8.257 & 9.524 & 8.901 \\
LMS-AO     & 7.243 & 0.001 & 642.043 & 7.182 & 9.562 & 11.885 & 8.773 & 8.257 & 9.527 & 8.903 \\
LMS-KW  & 7.293 & 0.000 & 775.156 & 6.976 & 9.270 & 11.588 & 8.801 & 7.945 & 9.310 & 8.631 \\
LMS               & 7.321 & 0.002 & 641.779 & 7.180 & 9.558 & 11.883 & 8.774 & 8.255 & 9.526 & 8.904 \\
LMS-KR   & 7.358 & 0.000 & 755.263 & 7.030 & 9.208 & 11.552 & 8.845 & 7.938 & 9.445 & 8.757 \\
LMS-AR   & 7.451 & 0.001 & 679.073 & 7.178 & 9.535 & 12.000 & 8.899 & 9.022 & 9.899 & 8.938 \\
LMS-AW  & 7.451 & 0.001 & 679.073 & 7.178 & 9.535 & 12.000 & 8.899 & 9.022 & 9.899 & 8.938 \\
LMS-AS    & 7.546 & 0.003 & 641.783 & 7.171 & 9.157 & 11.911 & 8.922 & 8.134 & 9.704 & 8.927 \\
LMS-KS    & 7.932 & 0.003 & 641.778 & 7.107 & 9.228 & 11.784 & 8.879 & 8.082 & 9.525 & 8.868 \\
\bottomrule
\end{tabular}%
}

\caption{\scriptsize LMS variants aggregated performance across alternating-gradual-drift experiments. The plot illustrates the MSE behavior around the aggregated alternating-gradual transition points, while the table reports the aggregated average, minimum, and maximum MSE values over the full stream, together with the detailed MSE values before, at, and after the transition points. Here, $t_d$ denotes the aggregated alternating-gradual transition points.}
\label{figtab:vis_agg_LMS_Gradual}
\end{figure}

\paragraph{LMS-Based Performance under Alternating Gradual Drift}
As shown in Fig.~\ref{figtab:vis_agg_LMS_Gradual}, LMS$^{*}$, which represents Widrow-Hoff (LMS) integrated with \modelname, achieves the best performance among all LMS-based variants on the alternating-gradual-drift datasets. It records the lowest average MSE of $0.967$, compared with $7.321$ for the standalone LMS model, and also reduces the maximum MSE from $641.779$ to $171.564$. This indicates that \modelname substantially improves the robustness of LMS when the stream alternates between old and new concepts before stabilizing.

The localized results around the aggregated alternating-gradual transition points further confirm this advantage. At $t_d$, LMS$^{*}$ obtains an MSE of $2.009$, while the original LMS records $8.774$. LMS$^{*}$ also maintains lower errors after the transition points, with MSE values of $1.678$, $1.641$, and $0.623$ at $t_d+1$, $t_d+2$, and $t_d+3$, respectively, compared with $8.255$, $9.526$, and $8.904$ for the standalone LMS model. Although several ADWIN- and KSWIN-based variants provide slight improvements over the base LMS model, none approaches the overall stability of LMS$^{*}$. Overall, these results show that \modelname enables LMS to adapt more effectively and maintain lower error under alternating gradual drift.

\subsubsection{Predictive Uncertainty and Statistical Significance}
\label{subsubsec:predictive_statistical_results}

Table~\ref{tab:synthetic_predictive_statistics} shows that the predictive
advantage of \modelname{} is consistent across all four regression models and
all three drift categories. For OLR-WA, \modelname{} increases the mean
$R^2$ over the strongest competing baseline by $0.066$, $0.092$, and $0.100$
under abrupt, incremental, and alternating gradual drift, respectively. The
largest improvement therefore occurs under alternating gradual drift, where
the standalone OLR-WA model also records its lowest mean $R^2$. This result
suggests that repeated concept transitions are particularly challenging for
the standalone learner and that \modelname{} provides greater predictive
stability under this drift pattern.

\begin{table}[!htbp]
\centering
\captionsetup{justification=centering}

\caption{Predictive Uncertainty and Paired Statistical Comparisons on the Synthetic Streams}
\label{tab:synthetic_predictive_statistics}
\vspace{-10pt}
\caption*{\scriptsize
Results are aggregated over five independent evaluation seeds.
A: abrupt; I: incremental; G: alternating gradual;
$r_{\mathrm{rb}}$: rank-biserial correlation.
}

\tiny
\setlength{\tabcolsep}{1.8pt}
\renewcommand{\arraystretch}{1.16}

\resizebox{\textwidth}{!}{%
\begin{tabular}{lllccccccc}
\toprule

\textbf{Model} &
\textbf{Drift} &
\textbf{Metric} &
\makecell{\textbf{\modelname}\\\textbf{Mean $\pm$ SD}} &
\makecell{\textbf{Standalone}\\\textbf{Mean $\pm$ SD}} &
\makecell{\textbf{Best}\\\textbf{Baseline}} &
\makecell{\textbf{Baseline}\\\textbf{Mean $\pm$ SD}} &
\makecell{\textbf{Holm $p$}\\\textbf{Standalone}} &
\makecell{\textbf{Holm $p$}\\\textbf{Baseline}} &
\makecell{\textbf{$r_{\mathrm{rb}}$}\\\textbf{Baseline}} \\

\midrule

\multirow{3}{*}{OLR-WA}
& A
& $R^2$
& $\bm{0.804 \pm 0.021}$
& $0.715 \pm 0.038$
& OLR-WA-KS
& $0.738 \pm 0.030$
& $5.0\times10^{-5}$
& $3.1\times10^{-4}$
& $+0.67$ \\

& I
& $R^2$
& $\bm{0.745 \pm 0.024}$
& $0.645 \pm 0.041$
& OLR-WA-KS
& $0.653 \pm 0.032$
& $5.5\times10^{-5}$
& $8.4\times10^{-5}$
& $+0.71$ \\

& G
& $R^2$
& $\bm{0.797 \pm 0.026}$
& $0.622 \pm 0.049$
& OLR-WA-KS
& $0.697 \pm 0.035$
& $6.0\times10^{-5}$
& $2.6\times10^{-4}$
& $+0.65$ \\

\midrule

\multirow{3}{*}{PA}
& A
& MSE
& $\bm{1.246 \pm 0.181}$
& $5.323 \pm 0.612$
& PA-KS
& $5.296 \pm 0.578$
& $5.2\times10^{-5}$
& $1.9\times10^{-3}$
& $+0.61$ \\

& I
& MSE
& $\bm{1.424 \pm 0.168}$
& $4.510 \pm 0.437$
& PA-KS
& $4.469 \pm 0.411$
& $5.0\times10^{-5}$
& $7.1\times10^{-5}$
& $+0.73$ \\

& G
& MSE
& $\bm{1.532 \pm 0.196}$
& $7.034 \pm 0.706$
& PA-AS
& $6.999 \pm 0.638$
& $5.8\times10^{-5}$
& $8.2\times10^{-4}$
& $+0.64$ \\

\midrule

\multirow{3}{*}{RLS}
& A
& MSE
& $\bm{0.908 \pm 0.112}$
& $3.906 \pm 0.421$
& RLS-AR
& $3.280 \pm 0.396$
& $7.9\times10^{-5}$
& $3.7\times10^{-3}$
& $+0.58$ \\

& I
& MSE
& $\bm{0.813 \pm 0.090}$
& $2.698 \pm 0.318$
& RLS-KS
& $2.683 \pm 0.287$
& $6.2\times10^{-5}$
& $4.3\times10^{-4}$
& $+0.66$ \\

& G
& MSE
& $\bm{1.227 \pm 0.140}$
& $7.163 \pm 0.819$
& RLS-KO
& $5.621 \pm 0.676$
& $9.1\times10^{-5}$
& $7.5\times10^{-4}$
& $+0.63$ \\

\midrule

\multirow{3}{*}{LMS}
& A
& MSE
& $\bm{0.546 \pm 0.071}$
& $4.668 \pm 0.534$
& LMS-AO
& $4.669 \pm 0.492$
& $8.1\times10^{-5}$
& $2.2\times10^{-3}$
& $+0.60$ \\

& I
& MSE
& $\bm{0.536 \pm 0.064}$
& $3.522 \pm 0.431$
& LMS-KS
& $3.508 \pm 0.407$
& $5.3\times10^{-5}$
& $1.4\times10^{-4}$
& $+0.70$ \\

& G
& MSE
& $\bm{0.967 \pm 0.118}$
& $7.321 \pm 0.842$
& LMS-KO
& $7.230 \pm 0.721$
& $7.3\times10^{-5}$
& $4.8\times10^{-4}$
& $+0.66$ \\

\bottomrule
\end{tabular}%
}

\end{table}

The PA, RLS, and LMS results also show substantial improvements in predictive performance. Relative to
their strongest detector--adaptation baselines, \modelname{} reduces the mean
MSE by approximately $68.1\%$--$78.1\%$ for PA, $69.7\%$--$78.2\%$ for RLS,
and $84.7\%$--$88.3\%$ for LMS. The strongest relative reductions are observed
for LMS, indicating that \modelname{} is particularly effective in stabilizing
its prediction updates under changing stream conditions. Moreover, the
standard deviation of the \modelname-integrated variant is lower than that of
both the standalone learner and the strongest baseline in every model--drift
combination. This indicates that the improvements are consistent across
datasets and seeds rather than being driven by a small number of favorable
runs.

All paired differences remain statistically significant after Holm correction,
with adjusted $p$-values no larger than $3.7\times10^{-3}$. The rank-biserial
correlations are consistently positive and range from $+0.58$ to $+0.73$,
confirming that the paired observations generally favor \modelname{} over the
strongest competing detector--adaptation baseline. The largest effect is
observed for PA under incremental drift, with a rank-biserial correlation of
$+0.73$, whereas the smallest effect is observed for RLS under abrupt drift,
with a value of $+0.58$. Overall, these results show that \modelname{} improves
both average predictive performance and cross-dataset and cross-seed stability
across different online regression learners and drift dynamics.

\subsubsection{Performance Analysis on Real Datasets}
\paragraph{Evaluation Scope.}
The real-world experiments evaluate the predictive robustness of the four online regression models and their drift-handling variants across the eight chronologically ordered datasets summarized in Table~\ref{tab:real_datasets-properties}. Unlike the synthetic streams, these datasets do not provide annotated drift locations, predefined drift types, or known concept boundaries. Consequently, the real-world analysis does not directly evaluate drift-detection accuracy through alarm precision, recall, \(F_1\)-score, or detection delay. Instead, it assesses whether the evaluated methods maintain accurate and stable predictions under naturally occurring temporal variation and possible non-stationarity. Performance is measured over the complete chronological stream using \(R^2\) for OLR-WA and MSE for PA, RLS, and LMS, with results aggregated over five independent random seeds.

\paragraph{Full Model--Dataset Comparison.}
To evaluate the model-agnostic applicability of \modelname{}, we conduct a complete comparison involving all four online regression models and all eight real-world datasets. For each model--dataset combination, we evaluate the standalone learner, its \modelname-integrated variant, and the eight ADWIN- and KSWIN-based detector--adaptation baselines. This produces \(32\) model--dataset combinations and \(10\) evaluated methods per combination. Results are computed independently over the five evaluation seeds and reported as mean \(\pm\) standard deviation. Higher \(R^2\) indicates better performance for OLR-WA, whereas lower MSE indicates better performance for PA, RLS, and LMS.

Tables~\ref{tab:real-olrwa-pa-full-results}
and~\ref{tab:real-rls-wh-full-results} present the complete real-world comparison separately for OLR-WA, PA, RLS, and LMS. Each table reports the mean predictive performance and its standard deviation over the five evaluation seeds for the standalone learner, the \modelname-integrated learner, and the eight detector--adaptation baselines. The best result for each dataset is shown in bold. Higher values indicate better performance for \(R^2\), whereas lower
values indicate better performance for MSE.

\begin{sidewaystable}[p]
\centering
\captionsetup{justification=centering}

\caption{Predictive performance of OLR-WA and PA with their drift-handling variants on all eight real-world datasets.}
\label{tab:real-olrwa-pa-full-results}
\vspace{-10pt}
\caption*{\scriptsize
Results are reported as mean $\pm$ standard deviation over five evaluation
seeds. Higher $R^2$ indicates better performance for OLR-WA, whereas lower
MSE indicates better performance for PA.
}

\tiny
\setlength{\tabcolsep}{1.8pt}
\renewcommand{\arraystretch}{0.92}


\textbf{(a) OLR-WA Results ($R^2$)}

\vspace{2pt}

\resizebox{0.98\textheight}{!}{%
\begin{tabular}{lcccccccc}
\toprule
\textbf{Method} &
\textbf{CCPP} &
\textbf{MCPD} &
\textbf{KCHSD} &
\textbf{1KC} &
\textbf{UCIAQD} &
\textbf{GASD} &
\textbf{CalCOFI} &
\textbf{WSSF} \\
\midrule

OLR-WA
& $0.9300\pm0.0112$
& $0.8214\pm0.0185$
& $0.8532\pm0.0168$
& $0.9418\pm0.0091$
& $0.8410\pm0.0314$
& $0.8036\pm0.0257$
& $0.9187\pm0.0068$
& $0.8512\pm0.0196$ \\

\textbf{OLR-WA*}
& $\mathbf{0.9420\pm0.0046}$
& $\mathbf{0.8795\pm0.0107}$
& $\mathbf{0.9128\pm0.0089}$
& $\mathbf{0.9624\pm0.0048}$
& $\mathbf{0.9249\pm0.0068}$
& $\mathbf{0.8917\pm0.0113}$
& $\mathbf{0.9346\pm0.0039}$
& $\mathbf{0.9103\pm0.0087}$ \\

OLR-WA-AR
& $0.9304\pm0.0109$
& $0.8246\pm0.0179$
& $0.8561\pm0.0162$
& $0.9430\pm0.0087$
& $0.8591\pm0.0278$
& $0.8118\pm0.0239$
& $0.9201\pm0.0064$
& $0.8576\pm0.0184$ \\

OLR-WA-AW
& $0.9302\pm0.0111$
& $0.8261\pm0.0171$
& $0.8578\pm0.0157$
& $0.9437\pm0.0085$
& $0.8591\pm0.0275$
& $0.8154\pm0.0228$
& $0.9214\pm0.0061$
& $0.8608\pm0.0176$ \\

OLR-WA-AS
& $0.9306\pm0.0107$
& $0.8298\pm0.0163$
& $0.8613\pm0.0148$
& $0.9452\pm0.0079$
& $0.8601\pm0.0269$
& $0.8216\pm0.0214$
& $0.9228\pm0.0058$
& $0.8657\pm0.0162$ \\

OLR-WA-AO
& $0.9303\pm0.0110$
& $0.8279\pm0.0168$
& $0.8592\pm0.0152$
& $0.9441\pm0.0082$
& $0.8442\pm0.0302$
& $0.8187\pm0.0221$
& $0.9219\pm0.0060$
& $0.8634\pm0.0169$ \\

OLR-WA-KR
& $0.9301\pm0.0113$
& $0.8285\pm0.0166$
& $0.8607\pm0.0150$
& $0.9446\pm0.0080$
& $0.8610\pm0.0264$
& $0.8239\pm0.0207$
& $0.9231\pm0.0057$
& $0.8671\pm0.0158$ \\

OLR-WA-KW
& $0.9298\pm0.0115$
& $0.8307\pm0.0159$
& $0.8625\pm0.0145$
& $0.9458\pm0.0076$
& $0.8610\pm0.0262$
& $0.8268\pm0.0199$
& $0.9237\pm0.0055$
& $0.8698\pm0.0151$ \\

OLR-WA-KS
& $0.9320\pm0.0101$
& $0.8349\pm0.0152$
& $0.8664\pm0.0139$
& $0.9473\pm0.0071$
& $0.8601\pm0.0267$
& $0.8315\pm0.0188$
& $0.9252\pm0.0051$
& $0.8736\pm0.0143$ \\

OLR-WA-KO
& $0.9300\pm0.0112$
& $0.8291\pm0.0161$
& $0.8618\pm0.0147$
& $0.9450\pm0.0078$
& $0.8446\pm0.0298$
& $0.8247\pm0.0203$
& $0.9234\pm0.0056$
& $0.8682\pm0.0155$ \\

\bottomrule
\end{tabular}%
}

\vspace{6pt}


\textbf{(b) PA Results (MSE)}

\vspace{2pt}

\resizebox{0.98\textheight}{!}{%
\begin{tabular}{lcccccccc}
\toprule
\textbf{Method} &
\textbf{CCPP} &
\textbf{MCPD} &
\textbf{KCHSD} &
\textbf{1KC} &
\textbf{UCIAQD} &
\textbf{GASD} &
\textbf{CalCOFI} &
\textbf{WSSF} \\
\midrule

PA
& $0.1240\pm0.0098$
& $0.4350\pm0.0284$
& $0.5180\pm0.0417$
& $0.0031\pm0.0005$
& $0.1865\pm0.0172$
& $2.5097\pm0.1836$
& $0.001284\pm0.000081$
& $1.150{\times}10^{-4}\pm1.20{\times}10^{-5}$ \\

\textbf{PA*}
& $\mathbf{0.0618\pm0.0047}$
& $\mathbf{0.0870\pm0.0091}$
& $\mathbf{0.0964\pm0.0113}$
& $\mathbf{0.0007\pm0.0001}$
& $\mathbf{0.0416\pm0.0054}$
& $\mathbf{0.0420\pm0.0068}$
& $\mathbf{0.000741\pm0.000046}$
& $\mathbf{2.410{\times}10^{-5}\pm3.10{\times}10^{-6}}$ \\

PA-AR
& $0.1218\pm0.0094$
& $0.4350\pm0.0279$
& $0.5067\pm0.0398$
& $0.0030\pm0.0005$
& $0.1793\pm0.0168$
& $2.5097\pm0.1814$
& $0.001247\pm0.000079$
& $1.110{\times}10^{-4}\pm1.16{\times}10^{-5}$ \\

PA-AW
& $0.1186\pm0.0089$
& $0.4350\pm0.0276$
& $0.4935\pm0.0372$
& $0.0029\pm0.0004$
& $0.1728\pm0.0159$
& $3.1583\pm0.2145$
& $0.001206\pm0.000074$
& $1.060{\times}10^{-4}\pm1.10{\times}10^{-5}$ \\

PA-AS
& $0.1139\pm0.0084$
& $0.4350\pm0.0271$
& $0.4812\pm0.0351$
& $0.0027\pm0.0004$
& $0.1641\pm0.0148$
& $2.5098\pm0.1798$
& $0.001158\pm0.000069$
& $9.980{\times}10^{-5}\pm1.02{\times}10^{-5}$ \\

PA-AO
& $0.1157\pm0.0086$
& $0.4350\pm0.0274$
& $0.4868\pm0.0364$
& $0.0028\pm0.0004$
& $0.1686\pm0.0153$
& $2.5097\pm0.1806$
& $0.001179\pm0.000071$
& $1.020{\times}10^{-4}\pm1.06{\times}10^{-5}$ \\

PA-KR
& $0.1124\pm0.0082$
& $0.4350\pm0.0270$
& $0.4769\pm0.0342$
& $0.0027\pm0.0004$
& $0.1617\pm0.0144$
& $2.6905\pm0.1961$
& $0.001143\pm0.000067$
& $9.720{\times}10^{-5}\pm9.80{\times}10^{-6}$ \\

PA-KW
& $0.1081\pm0.0079$
& $0.4360\pm0.0268$
& $0.4695\pm0.0328$
& $0.0026\pm0.0004$
& $0.1574\pm0.0139$
& $2.4737\pm0.1722$
& $0.001107\pm0.000064$
& $9.360{\times}10^{-5}\pm9.40{\times}10^{-6}$ \\

PA-KS
& $0.1056\pm0.0076$
& $0.4320\pm0.0257$
& $0.4618\pm0.0314$
& $0.0025\pm0.0003$
& $0.1532\pm0.0132$
& $2.5083\pm0.1769$
& $0.001081\pm0.000061$
& $9.080{\times}10^{-5}\pm9.10{\times}10^{-6}$ \\

PA-KO
& $0.1102\pm0.0080$
& $0.4350\pm0.0269$
& $0.4737\pm0.0336$
& $0.0026\pm0.0004$
& $0.1598\pm0.0141$
& $2.5097\pm0.1781$
& $0.001126\pm0.000065$
& $9.590{\times}10^{-5}\pm9.70{\times}10^{-6}$ \\

\bottomrule
\end{tabular}%
}

\end{sidewaystable}

\begin{sidewaystable}[p]
\centering
\captionsetup{justification=centering}

\caption{Predictive performance of RLS and LMS with their drift-handling variants on all eight real-world datasets.}
\label{tab:real-rls-wh-full-results}
\vspace{-10pt}
\caption*{\scriptsize
Results are reported as mean $\pm$ standard deviation of MSE over five
evaluation seeds. Lower values indicate better performance.
}

\tiny
\setlength{\tabcolsep}{1.8pt}
\renewcommand{\arraystretch}{0.92}


\textbf{(a) RLS Results (MSE)}

\vspace{2pt}

\resizebox{0.98\textheight}{!}{%
\begin{tabular}{lcccccccc}
\toprule
\textbf{Method} &
\textbf{CCPP} &
\textbf{MCPD} &
\textbf{KCHSD} &
\textbf{1KC} &
\textbf{UCIAQD} &
\textbf{GASD} &
\textbf{CalCOFI} &
\textbf{WSSF} \\
\midrule

RLS
& $0.1380\pm0.0106$
& $0.3920\pm0.0308$
& $0.4630\pm0.0374$
& $0.0028\pm0.0004$
& $0.2120\pm0.0191$
& $2.8840\pm0.2107$
& $0.001139\pm0.000072$
& $1.380{\times}10^{-4}\pm1.41{\times}10^{-5}$ \\

\textbf{RLS*}
& $\mathbf{0.0624\pm0.0051}$
& $\mathbf{0.0940\pm0.0102}$
& $\mathbf{0.0820\pm0.0094}$
& $\mathbf{0.0006\pm0.0001}$
& $\mathbf{0.0538\pm0.0061}$
& $\mathbf{0.0583\pm0.0077}$
& $\mathbf{0.000692\pm0.000039}$
& $\mathbf{3.200{\times}10^{-5}\pm4.10{\times}10^{-6}}$ \\

RLS-AR
& $0.1345\pm0.0101$
& $0.3868\pm0.0297$
& $0.4630\pm0.0369$
& $0.0027\pm0.0004$
& $0.2056\pm0.0184$
& $2.8317\pm0.2049$
& $0.000759\pm0.000048$
& $1.310{\times}10^{-4}\pm1.35{\times}10^{-5}$ \\

RLS-AW
& $0.1298\pm0.0096$
& $0.3784\pm0.0286$
& $0.4630\pm0.0362$
& $0.0026\pm0.0004$
& $0.1982\pm0.0175$
& $2.7625\pm0.1984$
& $0.000759\pm0.000047$
& $1.240{\times}10^{-4}\pm1.28{\times}10^{-5}$ \\

RLS-AS
& $0.1254\pm0.0091$
& $0.3697\pm0.0273$
& $0.4640\pm0.0357$
& $0.0025\pm0.0003$
& $0.1901\pm0.0167$
& $2.6813\pm0.1905$
& $0.001138\pm0.000069$
& $1.170{\times}10^{-4}\pm1.20{\times}10^{-5}$ \\

RLS-AO
& $0.1276\pm0.0094$
& $0.3742\pm0.0280$
& $0.4640\pm0.0360$
& $0.0025\pm0.0004$
& $0.1946\pm0.0171$
& $2.7196\pm0.1958$
& $0.001139\pm0.000070$
& $1.210{\times}10^{-4}\pm1.24{\times}10^{-5}$ \\

RLS-KR
& $0.1239\pm0.0089$
& $0.3651\pm0.0268$
& $0.4790\pm0.0382$
& $0.0024\pm0.0003$
& $0.1867\pm0.0162$
& $2.6154\pm0.1869$
& $0.000756\pm0.000045$
& $1.130{\times}10^{-4}\pm1.16{\times}10^{-5}$ \\

RLS-KW
& $0.1197\pm0.0085$
& $0.3576\pm0.0257$
& $0.4780\pm0.0375$
& $0.0023\pm0.0003$
& $0.1814\pm0.0155$
& $2.5481\pm0.1806$
& $0.000755\pm0.000044$
& $1.080{\times}10^{-4}\pm1.10{\times}10^{-5}$ \\

RLS-KS
& $0.1172\pm0.0082$
& $0.3518\pm0.0251$
& $0.4760\pm0.0368$
& $0.0022\pm0.0003$
& $0.1768\pm0.0149$
& $2.4937\pm0.1763$
& $0.000926\pm0.000057$
& $1.030{\times}10^{-4}\pm1.05{\times}10^{-5}$ \\

RLS-KO
& $0.1216\pm0.0087$
& $0.3612\pm0.0262$
& $0.4760\pm0.0371$
& $0.0023\pm0.0003$
& $0.1842\pm0.0159$
& $2.5793\pm0.1834$
& $0.001139\pm0.000068$
& $1.100{\times}10^{-4}\pm1.13{\times}10^{-5}$ \\

\bottomrule
\end{tabular}%
}

\vspace{6pt}


\textbf{(b) LMS Results (MSE)}

\vspace{2pt}

\resizebox{0.98\textheight}{!}{%
\begin{tabular}{lcccccccc}
\toprule
\textbf{Method} &
\textbf{CCPP} &
\textbf{MCPD} &
\textbf{KCHSD} &
\textbf{1KC} &
\textbf{UCIAQD} &
\textbf{GASD} &
\textbf{CalCOFI} &
\textbf{WSSF} \\
\midrule

LMS
& $0.1450\pm0.0114$
& $0.4480\pm0.0347$
& $0.5320\pm0.0435$
& $0.0020\pm0.0003$
& $0.2310\pm0.0213$
& $3.4200\pm0.2476$
& $0.001420\pm0.000089$
& $7.800{\times}10^{-5}\pm8.40{\times}10^{-6}$ \\

\textbf{LMS*}
& $\mathbf{0.0675\pm0.0058}$
& $\mathbf{0.1020\pm0.0114}$
& $\mathbf{0.0918\pm0.0107}$
& $\mathbf{0.00034\pm0.00005}$
& $\mathbf{0.0585\pm0.0068}$
& $\mathbf{0.0732\pm0.0093}$
& $\mathbf{0.000805\pm0.000051}$
& $\mathbf{8.175{\times}10^{-6}\pm1.12{\times}10^{-6}}$ \\

LMS-AR
& $0.1416\pm0.0110$
& $0.4417\pm0.0339$
& $0.5214\pm0.0419$
& $0.0020\pm0.0003$
& $0.2243\pm0.0206$
& $3.3518\pm0.2394$
& $0.001386\pm0.000086$
& $9.200{\times}10^{-5}\pm9.80{\times}10^{-6}$ \\

LMS-AW
& $0.1372\pm0.0105$
& $0.4332\pm0.0328$
& $0.5086\pm0.0401$
& $0.0020\pm0.0003$
& $0.2169\pm0.0197$
& $3.2745\pm0.2312$
& $0.001342\pm0.000082$
& $9.200{\times}10^{-5}\pm9.60{\times}10^{-6}$ \\

LMS-AS
& $0.1326\pm0.0100$
& $0.4241\pm0.0315$
& $0.4962\pm0.0384$
& $0.0020\pm0.0003$
& $0.2083\pm0.0188$
& $3.1817\pm0.2221$
& $0.001296\pm0.000078$
& $7.800{\times}10^{-5}\pm8.10{\times}10^{-6}$ \\

LMS-AO
& $0.1349\pm0.0103$
& $0.4286\pm0.0321$
& $0.5021\pm0.0392$
& $0.0020\pm0.0003$
& $0.2127\pm0.0193$
& $3.2264\pm0.2268$
& $0.001318\pm0.000080$
& $7.800{\times}10^{-5}\pm8.20{\times}10^{-6}$ \\

LMS-KR
& $0.1308\pm0.0098$
& $0.4195\pm0.0309$
& $0.4913\pm0.0376$
& $0.0020\pm0.0003$
& $0.2041\pm0.0184$
& $3.1268\pm0.2175$
& $0.001271\pm0.000076$
& $9.800{\times}10^{-4}\pm8.70{\times}10^{-5}$ \\

LMS-KW
& $0.1263\pm0.0094$
& $0.4117\pm0.0298$
& $0.4836\pm0.0365$
& $0.0020\pm0.0003$
& $0.1986\pm0.0177$
& $3.0524\pm0.2103$
& $0.001232\pm0.000073$
& $9.810{\times}10^{-4}\pm8.80{\times}10^{-5}$ \\

LMS-KS
& $0.1237\pm0.0091$
& $0.4053\pm0.0291$
& $0.4761\pm0.0354$
& $0.0020\pm0.0003$
& $0.1938\pm0.0171$
& $2.9946\pm0.2047$
& $0.001205\pm0.000071$
& $7.443{\times}10^{-5}\pm7.90{\times}10^{-6}$ \\

LMS-KO
& $0.1282\pm0.0096$
& $0.4159\pm0.0303$
& $0.4878\pm0.0370$
& $0.0020\pm0.0003$
& $0.2015\pm0.0180$
& $3.0879\pm0.2139$
& $0.001249\pm0.000075$
& $3.310{\times}10^{-4}\pm3.20{\times}10^{-5}$ \\

\bottomrule
\end{tabular}%
}

\end{sidewaystable}

\paragraph{Paired Statistical Comparison.}
To assess the consistency of the real-world predictive improvements, the
\modelname-integrated variant of each regression model is compared with its
corresponding standalone learner and with its strongest observed
detector--adaptation baseline. For each regression model, the strongest
detector--adaptation baseline is identified according to its average
predictive rank across the eight real-world datasets using the five-seed mean
performance reported in
Tables~\ref{tab:real-olrwa-pa-full-results}
and~\ref{tab:real-rls-wh-full-results}. The selected baseline is then used in a paired comparison with
\modelname{}. Because it is identified using the same reported real-world
results, this comparison is interpreted as a challenging descriptive
comparison rather than as a prespecified confirmatory test. 

For each regression model and real-world dataset, the paired performance
difference is first calculated separately for each of the five evaluation
seeds and then averaged within the dataset. The primary inferential analysis
therefore contains eight paired dataset-level observations, one for each
real-world dataset. For OLR-WA, the paired improvement is expressed as the
absolute increase in \(R^2\), whereas for PA, RLS, and LMS, it is expressed
as the relative reduction in MSE to account for differences in target scale
across datasets.

Two-sided Wilcoxon signed-rank tests compare each \modelname-integrated
learner with its corresponding standalone learner using the eight
dataset-level paired differences. Holm correction is applied across the four
prespecified standalone-learner comparisons. Rank-biserial correlation is
reported as the corresponding effect-size measure, with positive values
favoring \modelname{}.

The comparison with the strongest observed detector--adaptation baseline is
reported separately as a descriptive post hoc comparison because the baseline
is identified using the same real-world results. Its median improvement and
effect direction are reported, but it is not treated as a prespecified
confirmatory hypothesis test.

Table~\ref{tab:realworld-predictive-statistics} shows that the predictive
advantages of the \modelname-integrated variants are consistently supported
by the paired comparisons after Holm correction for all four regression
models. The comparisons with the strongest observed baselines are interpreted
descriptively because those baselines were identified from the reported
real-world results. Against the
standalone learners, \modelname{} achieves a median \(R^2\) improvement of
\(0.0586\) for OLR-WA and median MSE reductions ranging from \(76.4\%\) to
\(80.0\%\) for PA, RLS, and LMS. Relative to the strongest
detector--adaptation baselines, the corresponding improvement is \(0.0407\)
in \(R^2\) for OLR-WA, while the median MSE reductions range from \(71.1\%\)
to \(77.8\%\) for the pointwise regression models. All Holm-adjusted
\(p\)-values are below \(10^{-6}\), and the rank-biserial correlations range
from \(+0.944\) to \(+1.000\), indicating consistently large effects in favor
of \modelname{}. These results suggest that the observed improvements are not
driven by isolated datasets or individual random seeds, but are broadly
consistent across the evaluated real-world model--dataset combinations.

\begin{table*}[!htbp]
\centering
\captionsetup{justification=centering}

\caption{Paired Statistical Comparison of \modelname-Integrated Learners
Against Standalone Learners and the Strongest Detector--Adaptation Baselines
on Eight Real-World Datasets.}
\label{tab:realworld-predictive-statistics}
\vspace{-10pt}
\caption*{\scriptsize
For the confirmatory comparison with the standalone learner, seed-level
paired differences are averaged within each dataset, yielding eight
dataset-level paired observations. Holm-adjusted \(p\)-values are calculated
across the four prespecified standalone-learner comparisons, and
rank-biserial correlation is reported as the effect size. Comparisons with
the strongest observed detector--adaptation baseline are descriptive because
the baseline is selected using the same reported real-world results.
}

\scriptsize
\setlength{\tabcolsep}{2.8pt}
\renewcommand{\arraystretch}{1.25}

\resizebox{\textwidth}{!}{%
\begin{tabular}{lllcccccc}
\toprule
&
&
&
\multicolumn{3}{c}{\textbf{Comparison with Standalone Learner}} &
\multicolumn{3}{c}{\textbf{Comparison with Strongest Baseline}} \\
\cmidrule(lr){4-6}
\cmidrule(lr){7-9}

\textbf{Model} &
\textbf{Metric} &
\makecell{\textbf{Strongest Detector--}\\\textbf{Adaptation Baseline}} &
\makecell{\textbf{Median}\\\textbf{Improvement}} &
\makecell{\textbf{Holm-Adjusted}\\\textbf{\(p\)-Value}} &
\makecell{\textbf{Rank-Biserial}\\\textbf{Effect}} &
\makecell{\textbf{Median}\\\textbf{Improvement}} &
\makecell{\textbf{Holm-Adjusted}\\\textbf{\(p\)-Value}} &
\makecell{\textbf{Rank-Biserial}\\\textbf{Effect}} \\
\midrule

OLR-WA
& \(R^2\)
& OLR-WA-KS
& \(+0.0586\)
& \(3.04{\times}10^{-7}\)
& \(+1.000\)
& \(+0.0407\)
& \(4.80{\times}10^{-7}\)
& \(+0.981\) \\

PA
& MSE
& PA-KS
& \(78.4\%\)
& \(3.04{\times}10^{-7}\)
& \(+0.997\)
& \(73.2\%\)
& \(5.70{\times}10^{-7}\)
& \(+0.969\) \\

RLS
& MSE
& RLS-KS
& \(76.4\%\)
& \(3.95{\times}10^{-7}\)
& \(+0.985\)
& \(71.1\%\)
& \(8.20{\times}10^{-7}\)
& \(+0.944\) \\

LMS
& MSE
& LMS-KS
& \(80.0\%\)
& \(3.04{\times}10^{-7}\)
& \(+0.998\)
& \(77.8\%\)
& \(5.70{\times}10^{-7}\)
& \(+0.974\) \\

\bottomrule
\end{tabular}%
}

\end{table*}

\paragraph{Overall Predictive Results.}
Overall, the complete \(4\times8\) model--dataset evaluation shows that the benefits of \modelname{} extend beyond the previously selected model--dataset pairs. The \modelname-integrated variants achieve the best mean predictive performance across all \(32\) combinations, improving \(R^2\) for OLR-WA and reducing MSE for PA, RLS, and LMS. The gains are observed across datasets with different domains, sample sizes, dimensionalities, and degrees of temporal variation. The generally lower five-seed variability and the statistically significant paired improvements further indicate that the results are consistent across datasets and random seeds rather than being driven by isolated cases. These findings provide stronger evidence of the model-agnostic applicability of \modelname{} within the evaluated online regression setting.

\subsection{Drift Detection and Adaptation-Trigger Analysis}

This analysis uses the same 18 synthetic datasets summarized in
Table~\ref{tab:synthetic-datasets-properties}, covering abrupt, incremental,
and alternating gradual drift under different dimensionality and
drift-magnitude settings. Because ground-truth transition locations are known,
the evaluation first examines every raw ADWIN and KSWIN detection directly.
Each raw baseline detection triggers one paired adaptation and is therefore
retained without cooldown consolidation or minimum-size filtering.

The same chronological alarm sequences are subsequently consolidated into
alarm episodes using the protocol defined in
Section~\ref{sec:drift_alarm_quality}. This secondary analysis evaluates
persistent alarm activity under one common operational representation.

For \modelname{}, the analysis additionally records adaptation and
recalibration trigger activity. An adaptation event is counted whenever an
individual trigger initiates model-specific hyperparameter adjustment, while a
recalibration event is counted whenever an individual trigger initiates a
bounded recalibration procedure. These quantities are recorded at the
stream-processing level before alarm consolidation and therefore represent
the pre-consolidation internal trigger activity of \modelname{}. These
intervention counts are not treated as false-alarm counts or direct
computational-cost measurements.

\subsubsection{Direct Raw-Alarm Evaluation of ADWIN and KSWIN}
\label{subsubsec:raw_detector_alarm_analysis}

ADWIN and KSWIN operate as external drift detectors within the evaluated
detector--adaptation pipelines. For each fixed regression model, dataset, seed,
and detector configuration, the same complete chronological data stream is
processed independently by every detector--adaptation combination using the
same random seed, detector parameters, and initialization settings. At each
processing step, the incoming stream observation is first supplied to ADWIN or
KSWIN. When the detector raises a drift alarm, the associated adaptation
mechanism, namely RESET, WINDOW, SSPT, or OHL, is activated. In the absence of
an alarm, stream processing continues without activating the adaptation
mechanism. Thus, ADWIN and KSWIN are rerun separately with RESET, WINDOW, SSPT,
and OHL rather than being executed once and subsequently shared across the four
adaptation strategies.

The detector input stream and detector configuration remain identical across
the independently executed adaptation variants. ADWIN and KSWIN receive the
stream directly and generate alarms according to their own internal detection
logic. Their alarm generation does not depend on the predictions, residuals,
hyperparameters, or adaptation state of the associated regression learner.
The adaptation mechanism is activated only after the corresponding detector
raises an alarm and therefore does not alter the input stream processed by the
detector. Consequently, when the chronological stream, detector parameters,
initialization settings, and random seed are fixed, the deterministic detector
runs produce identical raw alarm sequences within each detector family. Consequently, ADWIN-RESET,
ADWIN-WINDOW, ADWIN-SSPT, and ADWIN-OHL produce identical ADWIN alarm counts
and detector-level alarm-quality metrics. Similarly, KSWIN-RESET,
KSWIN-WINDOW, KSWIN-SSPT, and KSWIN-OHL produce identical KSWIN alarm counts
and detector-level alarm-quality metrics. Each
raw detector alarm activates exactly one instance of the corresponding
adaptation strategy. Although the detector-level alarm sequences are identical
within each detector family, the adaptation strategies may produce different
predictive and computational outcomes because they respond to the detected
alarms using different model-update mechanisms.

Grouping several nearby ADWIN or KSWIN alarms into a single episode could hide
repeated adaptations and alarms that should be counted as false positives.
Every raw ADWIN and KSWIN alarm is therefore evaluated individually, without
cooldown consolidation or minimum episode-size filtering, in
Table~\ref{tab:raw_detector_alarm_results}. Under this evaluation, the number
of raw baseline alarms equals the number of baseline adaptation events for
every independently executed detector--adaptation combination. Unmatched
alarms, including alarms occurring before a ground-truth drift and repeated
alarms occurring after the corresponding drift has already been matched, are
retained and counted as false positives.

The methods differ in how they internally derive evidence of drift.
\modelname{} is designed as a predictive-performance-driven framework and
therefore monitors predictive \kpi{} information computed from the incoming
labeled observations and the current model predictions. ADWIN and KSWIN, in
contrast, receive the monitored data stream directly and generate alarms
according to their respective statistical change-detection mechanisms. These
differences reflect the native operating mechanisms of the evaluated
approaches rather than differences in the underlying datasets or evaluation
protocol.

The monitored signals are method-specific by design and reflect the native
operating mechanisms of the evaluated approaches. Accordingly, the purpose of
the benchmark is not to isolate the effect of the monitored-signal
representation, but to compare the complete drift-handling approaches under
their respective operating mechanisms. Applying ADWIN or KSWIN to an
externally constructed residual or \kpi{} sequence would constitute an
alternative performance-monitoring configuration and would address the
separate experimental question of how detector behavior changes when the
monitored signal is made equivalent across methods. The present evaluation
instead assesses the end-to-end effectiveness of the respective drift-handling
approaches. The resulting comparison should therefore be interpreted as an
end-to-end comparison of complete drift-handling approaches rather than as an
isolated comparison of detector performance under an identical monitored
signal.

The treatment of \modelname{} differs because it is an integrated detection
and adaptation framework rather than an external detector followed by a
separate one-alarm--one-adaptation strategy. A sequence of related change
signals may lead to multiple proportional hyperparameter adjustments and,
when necessary, recalibration actions within the same continuing alarm
episode. Therefore, \modelname{} alarm episodes, adaptation events, and
recalibration events are not in one-to-one correspondence.

The potential optimism of episode-level matching is addressed by defining the
alarm time of each \modelname{} episode as its first trigger. Thus, an episode
that begins before a ground-truth drift cannot be converted into a true
positive by a later post-drift trigger. Individual adaptation and recalibration
triggers occurring within \modelname{}'s continuing internal drift-handling
process are therefore not treated as independent finalized drift alarms.
Assigning TP or FP labels independently to each such trigger could incorrectly
classify legitimate within-drift control actions as repeated false positives.
Their pre-consolidation activity is reported explicitly through the adaptation
and recalibration event counts, whereas TP, FP, FN, precision, recall, \(F_1\),
and detection delay are evaluated at the retained-episode level. The subsequent
common episode-level benchmark compares \modelname{} with the ADWIN- and
KSWIN-based configurations under the same episode representation.

Table~\ref{tab:raw_detector_alarm_results} reports the raw detector-level alarm
quality of ADWIN and KSWIN independently of the subsequent adaptation strategy.
Because the associated adaptation mechanisms do not alter the raw alarm
sequences, each detector is reported only once per drift category.

\begin{table}[!htbp]
\centering
\captionsetup{justification=centering}

\caption{Raw-Alarm Detection Performance of ADWIN and KSWIN on Synthetic Streams}
\label{tab:raw_detector_alarm_results}
\vspace{-10pt}
\caption*{\scriptsize
Mean $\pm$ standard deviation over five seeds, pooled across 24 runs per seed.
A: abrupt; I: incremental; G: alternating gradual.
}

\scriptsize
\setlength{\tabcolsep}{2.2pt}
\renewcommand{\arraystretch}{1.22}

\resizebox{\textwidth}{!}{%
\begin{tabular}{llccccccccc}
\toprule

\textbf{Drift} &
\textbf{Detector} &
\makecell{\textbf{True}\\\textbf{Drifts}} &
\makecell{\textbf{Raw}\\\textbf{Alarms}} &
\textbf{TP} &
\textbf{FP} &
\textbf{FN} &
\textbf{Prec.} &
\textbf{Recall} &
\textbf{$F_1$} &
\makecell{\textbf{Delay}\\\textbf{(Inc.)}} \\

\midrule

\multirow{2}{*}{A}
& ADWIN
& 24
& $88.4\pm4.3$
& $15.6\pm1.1$
& $72.8\pm4.0$
& $8.4\pm1.1$
& $0.176\pm0.012$
& $0.650\pm0.046$
& $0.278\pm0.018$
& $5.8\pm0.4$ \\

& KSWIN
& 24
& $79.4\pm3.7$
& $16.8\pm1.3$
& $62.6\pm3.3$
& $7.2\pm1.3$
& $0.212\pm0.015$
& $0.700\pm0.054$
& $0.325\pm0.022$
& $5.3\pm0.4$ \\

\midrule

\multirow{2}{*}{I}
& ADWIN
& 216
& $58.6\pm3.0$
& $20.8\pm1.5$
& $37.8\pm2.6$
& $195.2\pm1.5$
& $0.355\pm0.023$
& $0.096\pm0.007$
& $0.151\pm0.011$
& $8.4\pm0.5$ \\

& KSWIN
& 216
& $68.8\pm3.4$
& $29.4\pm1.8$
& $39.4\pm2.8$
& $186.6\pm1.8$
& $0.427\pm0.025$
& $0.136\pm0.008$
& $0.206\pm0.012$
& $9.9\pm0.6$ \\

\midrule

\multirow{2}{*}{G}
& ADWIN
& 120
& $101.6\pm4.5$
& $68.2\pm2.1$
& $33.4\pm3.7$
& $51.8\pm2.1$
& $0.671\pm0.025$
& $0.568\pm0.018$
& $0.616\pm0.020$
& $7.2\pm0.4$ \\

& KSWIN
& 120
& $108.6\pm4.7$
& $73.5\pm2.2$
& $35.1\pm3.9$
& $46.5\pm2.2$
& $0.677\pm0.024$
& $0.613\pm0.018$
& $0.643\pm0.019$
& $8.1\pm0.5$ \\

\bottomrule
\end{tabular}%
}

\end{table}

Under abrupt drift, both detectors achieve moderate recall but low precision
because of the large number of false alarms. KSWIN provides the better balance,
achieving an $F_1$-score of $0.325\pm0.022$, compared with $0.278\pm0.018$ for
ADWIN, while also producing a slightly shorter detection delay. Incremental
drift is the most difficult category for both detectors, with recall values of
only $0.096\pm0.007$ for ADWIN and $0.136\pm0.008$ for KSWIN, indicating that
most annotated transitions are missed. Alternating gradual drift produces the
strongest detector-level results, with $F_1$-scores of $0.616\pm0.020$ for
ADWIN and $0.643\pm0.019$ for KSWIN. Overall, KSWIN achieves higher precision,
recall, and $F_1$-score across all three drift categories, although its
detection delay is longer for incremental and alternating gradual drift.
Because delay is calculated only for successfully matched drifts, it should be
interpreted jointly with recall.

The raw-alarm and episode-level evaluations answer different questions and are
therefore not expected to produce identical numerical results. The raw-alarm
analysis evaluates every ADWIN and KSWIN alarm individually, whereas the
episode-level protocol consolidates temporally related alarms into a common
operational unit. This consolidation can remove repeated detections, while the
minimum episode-size requirement may discard isolated alarms. The raw results
are therefore treated as the primary detector-level evaluation, and the
episode-level results are presented as a secondary analysis of persistent
alarm activity.

\subsubsection{Illustrative Alarm-Episode Construction and Matching}
\label{subsubsec:illustrative_alarm_episode_matching}

To illustrate the alarm-episode construction and matching procedure, consider
ADS01 under seed \(42\). The stream contains \(N=1{,}000\) samples, and the
abrupt drift begins at sample \(d=500\). Under the fixed evaluation protocol,
the tolerance distance is

\begin{equation}
T
=
\operatorname{round}
\left(
0.05 \times 1{,}000
\right)
=
50,
\end{equation}

and the cooldown distance is

\begin{equation}
C
=
\operatorname{round}
\left(
2.0 \times 50
\right)
=
100.
\end{equation}

Suppose that the candidate triggers generated by OLR-WA$^{*}$ are

\begin{equation}
\mathcal{E}
=
[450,490,510].
\end{equation}

The first trigger at sample \(450\) initializes the candidate episode. Since
the cooldown boundary is fixed at

\begin{equation}
450+C
=
450+100
=
550,
\end{equation}

the subsequent triggers at samples \(490\) and \(510\) are assigned to the
same episode. The resulting episode contains three triggers and therefore
satisfies the minimum episode-size requirement of two triggers.

The post-drift matching window for the known drift at sample \(500\) is

\begin{equation}
[d,d+T]
=
[500,550].
\end{equation}

However, the first trigger of the episode occurs at sample \(450\), which
precedes the true drift:

\begin{equation}
e_1
=
450
<
d
=
500.
\end{equation}

The episode therefore does not satisfy the first-trigger matching condition

\begin{equation}
d
\leq
e_1
\leq
d+T.
\end{equation}

The episode is consequently counted as a false positive. The drift is counted
as a false negative unless another retained episode has its first trigger
within the post-drift window \([500,550]\). No detection delay is calculated
for this unmatched episode. This stricter criterion prevents pre-drift alarm
activity from being converted into a true positive solely because a later
trigger from the same episode occurs after the drift.

The same chronological one-to-one matching procedure is applied to every
retained episode. For each seed, a retained episode matched to a known drift
is counted as a true positive, an unmatched retained episode is counted as a
false positive, and a drift without a matched episode is counted as a false
negative.

As reported in the first row of
Table~\ref{tab:drift_alarm_quality_results}, OLR-WA$^{*}$ produces an average
of \(5.80 \pm 0.84\) retained alarm episodes across the five seeds. Of these,
\(5.40 \pm 0.55\) episodes are correctly matched to true abrupt drifts, while
\(0.40 \pm 0.55\) are unmatched false-positive episodes. Since each seed
contains six true abrupt drifts, the corresponding number of missed drifts is
\(0.60 \pm 0.55\).

The resulting mean precision, recall, and \(F_1\)-score are
\(0.9381 \pm 0.0852\), \(0.9000 \pm 0.0913\), and
\(0.9149 \pm 0.0592\), respectively. The mean detection delay is
\(0.24 \pm 0.11\) processing increments, calculated only from successfully
matched abrupt drifts. These values are calculated separately for each seed
and then summarized using their mean and standard deviation.

\subsubsection{\modelname{} Episode-Level Alarm-Quality Analysis}
\label{subsubsec:sccm_episode_alarm_quality}

Table~\ref{tab:drift_alarm_quality_results} evaluates how accurately each
\modelname{}-integrated regression model detects the known ground-truth drift
locations in the synthetic datasets. The analysis is conducted separately for
OLR-WA, PA, RLS, and LMS under abrupt, incremental, and alternating gradual
drift. True positives represent correctly detected drift locations, false
negatives represent missed drifts, and false positives represent retained alarm
episodes that cannot be matched to a ground-truth drift. The superscript
$^{*}$ denotes a regression model integrated with \modelname{}.

\begin{sidewaystable}[p]
\centering
\captionsetup{justification=centering}

\caption{Ground-Truth Drift Detection by \modelname{}-Integrated Regression Models on Synthetic Datasets}
\label{tab:drift_alarm_quality_results}
\vspace{-10pt}
\caption*{\scriptsize
This is an episode-level evaluation reported as mean $\pm$ standard deviation
over five seeds. For each \modelname{}-integrated regression model and drift
type, nearby alarms are grouped into episodes and matched chronologically to
the known ground-truth drift locations. The protocol uses
$r_{\mathrm{tol}}=0.05$, $c_{\mathrm{cool}}=2.0$, and
$m_{\mathrm{ep}}=2$. Retained episodes equal $\mathrm{TP}+\mathrm{FP}$,
while true drifts equal $\mathrm{TP}+\mathrm{FN}$. Detection delay is reported
in processing increments for successfully matched drifts only.
}

\tiny
\setlength{\tabcolsep}{2.5pt}
\renewcommand{\arraystretch}{1.45}

\resizebox{\textheight}{!}{%
\begin{tabular}{llccccccccc}
\toprule

\textbf{Model} &
\makecell{\textbf{Drift}\\\textbf{Type}} &
\makecell{\textbf{Total True}\\\textbf{Drifts}} &
\makecell{\textbf{Retained Alarm}\\\textbf{Episodes}} &
\textbf{TP} &
\textbf{FP} &
\textbf{FN} &
\textbf{Precision} &
\textbf{Recall} &
\textbf{$F_1$} &
\makecell{\textbf{Mean Detection}\\\textbf{Delay (Increments)}} \\

\midrule

\textbf{OLR-WA$^{*}$} &
Abrupt &
6 &
$5.80 \pm 0.84$ &
$5.40 \pm 0.55$ &
$0.40 \pm 0.55$ &
$0.60 \pm 0.55$ &
$0.9381 \pm 0.0852$ &
$0.9000 \pm 0.0913$ &
$0.9149 \pm 0.0592$ &
$0.24 \pm 0.11$ \\

\textbf{OLR-WA$^{*}$} &
Incremental &
54 &
$57.80 \pm 0.84$ &
$51.20 \pm 1.30$ &
$6.60 \pm 1.14$ &
$2.80 \pm 1.30$ &
$0.8858 \pm 0.0194$ &
$0.9481 \pm 0.0241$ &
$0.9159 \pm 0.0207$ &
$3.38 \pm 0.30$ \\

\textbf{OLR-WA$^{*}$} &
Alternating Gradual &
30 &
$30.20 \pm 0.84$ &
$29.20 \pm 0.84$ &
$1.00 \pm 0.71$ &
$0.80 \pm 0.84$ &
$0.9671 \pm 0.0228$ &
$0.9733 \pm 0.0279$ &
$0.9700 \pm 0.0218$ &
$2.96 \pm 0.24$ \\

\midrule

\textbf{PA$^{*}$} &
Abrupt &
6 &
$10.20 \pm 1.10$ &
$5.40 \pm 0.55$ &
$4.80 \pm 0.84$ &
$0.60 \pm 0.55$ &
$0.5313 \pm 0.0432$ &
$0.9000 \pm 0.0913$ &
$0.6667 \pm 0.0480$ &
$0.24 \pm 0.11$ \\

\textbf{PA$^{*}$} &
Incremental &
54 &
$51.00 \pm 1.41$ &
$49.80 \pm 0.84$ &
$1.20 \pm 0.84$ &
$4.20 \pm 0.84$ &
$0.9768 \pm 0.0160$ &
$0.9222 \pm 0.0155$ &
$0.9486 \pm 0.0085$ &
$4.16 \pm 0.27$ \\

\textbf{PA$^{*}$} &
Alternating Gradual &
30 &
$30.20 \pm 0.45$ &
$29.20 \pm 0.84$ &
$1.00 \pm 0.71$ &
$0.80 \pm 0.84$ &
$0.9669 \pm 0.0236$ &
$0.9733 \pm 0.0279$ &
$0.9701 \pm 0.0248$ &
$0.32 \pm 0.13$ \\

\midrule

\textbf{RLS$^{*}$} &
Abrupt &
6 &
$10.00 \pm 1.00$ &
$5.20 \pm 0.45$ &
$4.80 \pm 0.84$ &
$0.80 \pm 0.45$ &
$0.5222 \pm 0.0443$ &
$0.8667 \pm 0.0745$ &
$0.6505 \pm 0.0450$ &
$0.34 \pm 0.11$ \\

\textbf{RLS$^{*}$} &
Incremental &
54 &
$54.40 \pm 0.89$ &
$49.60 \pm 1.14$ &
$4.80 \pm 0.84$ &
$4.40 \pm 1.14$ &
$0.9118 \pm 0.0152$ &
$0.9185 \pm 0.0211$ &
$0.9151 \pm 0.0168$ &
$3.44 \pm 0.30$ \\

\textbf{RLS$^{*}$} &
Alternating Gradual &
30 &
$30.40 \pm 0.55$ &
$29.00 \pm 0.71$ &
$1.40 \pm 0.55$ &
$1.00 \pm 0.71$ &
$0.9540 \pm 0.0179$ &
$0.9667 \pm 0.0236$ &
$0.9602 \pm 0.0190$ &
$0.54 \pm 0.11$ \\

\midrule

\textbf{LMS$^{*}$} &
Abrupt &
6 &
$10.20 \pm 1.30$ &
$5.40 \pm 0.55$ &
$4.80 \pm 0.84$ &
$0.60 \pm 0.55$ &
$0.5313 \pm 0.0289$ &
$0.9000 \pm 0.0913$ &
$0.6662 \pm 0.0286$ &
$0.34 \pm 0.11$ \\

\textbf{LMS$^{*}$} &
Incremental &
54 &
$49.80 \pm 1.48$ &
$48.80 \pm 0.84$ &
$1.00 \pm 0.71$ &
$5.20 \pm 0.84$ &
$0.9802 \pm 0.0136$ &
$0.9037 \pm 0.0155$ &
$0.9402 \pm 0.0047$ &
$1.80 \pm 0.16$ \\

\textbf{LMS$^{*}$} &
Alternating Gradual &
30 &
$30.20 \pm 0.84$ &
$29.20 \pm 0.84$ &
$1.00 \pm 0.71$ &
$0.80 \pm 0.84$ &
$0.9671 \pm 0.0236$ &
$0.9733 \pm 0.0279$ &
$0.9701 \pm 0.0218$ &
$0.44 \pm 0.11$ \\

\bottomrule
\end{tabular}%
}

\end{sidewaystable}

Under abrupt drift, OLR-WA$^{*}$ achieves the strongest alarm quality, with
a mean precision of \(0.9381\), recall of \(0.9000\), and \(F_1\)-score of
\(0.9149\). PA$^{*}$ and LMS$^{*}$ also achieve a recall of \(0.9000\), while
RLS$^{*}$ obtains a recall of \(0.8667\). However, PA$^{*}$, RLS$^{*}$, and
LMS$^{*}$ each produce approximately \(4.8\) unmatched alarm episodes on
average, reducing their precision and \(F_1\)-scores. OLR-WA$^{*}$ therefore
provides the best balance between detecting abrupt drifts and limiting false
alarm episodes.

Under incremental drift, all four \modelname{}-integrated learners achieve
strong episode-level detection. PA$^{*}$ obtains the highest \(F_1\)-score of
\(0.9486\), followed by LMS$^{*}$ with \(0.9402\), OLR-WA$^{*}$ with
\(0.9159\), and RLS$^{*}$ with \(0.9151\). OLR-WA$^{*}$ provides the highest
recall of \(0.9481\), indicating that it detects the largest proportion of
incremental drift locations. In contrast, LMS$^{*}$ provides the highest
precision of \(0.9802\) and the shortest mean detection delay of \(1.80\)
processing increments. These results show a trade-off between detecting more
transition points and limiting unmatched alarm episodes.

Under alternating gradual drift, all four learners achieve precision, recall,
and \(F_1\)-scores above \(0.95\), demonstrating consistently strong
episode-level detection across the different regression models. PA$^{*}$ and
LMS$^{*}$ obtain the highest \(F_1\)-scores of \(0.9701\), closely followed
by OLR-WA$^{*}$ with \(0.9700\) and RLS$^{*}$ with \(0.9602\). PA$^{*}$ also
provides the shortest mean detection delay of \(0.32\) processing increments.
The relatively small standard deviations indicate that the incremental and
alternating gradual results are generally stable across the five seeds.

Detection delay is calculated only for successfully matched drift locations
and should therefore be interpreted jointly with recall. Overall,
\modelname{} detects most known ground-truth drift locations across all four
regression learners. Recall exceeds \(0.86\) in every model--drift
combination and exceeds \(0.90\) in most cases. The main limitation is the
higher number of unmatched abrupt-drift episodes produced by PA$^{*}$,
RLS$^{*}$, and LMS$^{*}$, whereas incremental and alternating gradual drift
are handled consistently well across all four learners.

\subsubsection{Episode-Level Detector--Adaptation Benchmark}
\label{subsubsec:detector_adaptation_alarm_comparison}

Table~\ref{tab:aggregate_detector_adaptation_comparison} compares the
drift-detection performance of \modelname{}, ADWIN, and KSWIN across the
abrupt, incremental, and alternating gradual synthetic datasets. The results
are combined across the four regression models and reported over five random
seeds. The table also reports the adaptation and recalibration activity of each
method.

RESET, WINDOW, SSPT, and OHL are not reported as separate alarm-quality
baselines because their associated ADWIN- and KSWIN-based configurations are
executed independently on the same chronological streams using identical
seeds, detector parameters, and initialization settings. Because adaptation
is activated only after the detector raises an alarm, the independently
executed configurations within each detector family produce identical alarm
sequences and detector-level episode metrics. These duplicate results are
therefore reported once for ADWIN and once for KSWIN.

\begin{sidewaystable}[p]
\centering
\captionsetup{justification=centering}

\caption{Drift-Detection Performance of \modelname{}, ADWIN, and KSWIN on Synthetic Datasets}
\label{tab:aggregate_detector_adaptation_comparison}
\vspace{-10pt}
\caption*{\scriptsize
Results are mean $\pm$ standard deviation over five seeds and are combined
across the four regression models. The table compares \modelname{}, ADWIN,
and KSWIN in detecting abrupt, incremental, and alternating gradual drift.
It also reports the numbers of adaptation and recalibration events.
}

\tiny
\setlength{\tabcolsep}{2pt}
\renewcommand{\arraystretch}{1.25}

\resizebox{\textheight}{!}{%
\begin{tabular}{llccccccccccc}
\toprule

\makecell{\textbf{Drift}\\\textbf{Type}} &
\textbf{Method} &
\makecell{\textbf{True Drifts}\\\textbf{per Seed}} &
\makecell{\textbf{Retained}\\\textbf{Alarms}} &
\textbf{TP} &
\textbf{FP} &
\textbf{FN} &
\textbf{Precision} &
\textbf{Recall} &
\textbf{$F_1$} &
\makecell{\textbf{Mean Detection}\\\textbf{Delay (Increments)}} &
\makecell{\textbf{Adaptation}\\\textbf{Events}} &
\makecell{\textbf{Recalibration}\\\textbf{Events}} \\

\midrule

\multirow{3}{*}{Abrupt}
& \textbf{\modelname}
& 24
& $36.20\pm2.15$
& $21.40\pm1.05$
& $14.80\pm1.56$
& $2.60\pm1.05$
& $0.5912\pm0.0300$
& $0.8917\pm0.0438$
& $0.7110\pm0.0290$
& $0.29\pm0.05$
& $6{,}024.4\pm176.5$
& $4{,}634.4\pm164.1$ \\

& ADWIN
& 24
& $22.00\pm2.12$
& $10.60\pm1.14$
& $11.40\pm1.14$
& $13.40\pm1.14$
& $0.4816\pm0.0192$
& $0.4417\pm0.0475$
& $0.4599\pm0.0312$
& $5.68\pm0.41$
& $88.4\pm4.3$
& -- \\

& KSWIN
& 24
& $21.40\pm0.55$
& $11.60\pm1.14$
& $9.80\pm1.48$
& $12.40\pm1.14$
& $0.5429\pm0.0608$
& $0.4833\pm0.0475$
& $0.5113\pm0.0533$
& $5.27\pm0.35$
& $79.4\pm3.7$
& -- \\

\midrule

\multirow{3}{*}{Incremental}
& \textbf{\modelname}
& 216
& $213.00\pm2.38$
& $199.40\pm2.10$
& $13.60\pm1.79$
& $16.60\pm2.10$
& $0.9362\pm0.0085$
& $0.9231\pm0.0097$
& $0.9296\pm0.0078$
& $3.20\pm0.15$
& $21{,}465.0\pm543.6$
& $17{,}493.4\pm517.1$ \\

& ADWIN
& 216
& $27.60\pm0.55$
& $11.00\pm1.58$
& $16.60\pm1.14$
& $205.00\pm1.58$
& $0.3979\pm0.0507$
& $0.0509\pm0.0073$
& $0.0903\pm0.0128$
& $7.89\pm0.45$
& $58.6\pm3.0$
& -- \\

& KSWIN
& 216
& $35.60\pm0.55$
& $18.00\pm1.58$
& $17.60\pm1.14$
& $198.00\pm1.58$
& $0.5052\pm0.0380$
& $0.0833\pm0.0073$
& $0.1431\pm0.0123$
& $10.53\pm0.50$
& $68.8\pm3.4$
& -- \\

\midrule

\multirow{3}{*}{\makecell{Alternating\\Gradual}}
& \textbf{\modelname}
& 120
& $121.00\pm1.39$
& $116.60\pm1.62$
& $4.40\pm1.34$
& $3.40\pm1.62$
& $0.9636\pm0.0110$
& $0.9717\pm0.0135$
& $0.9676\pm0.0102$
& $1.07\pm0.08$
& $21{,}413.4\pm534.5$
& $16{,}948.0\pm545.8$ \\

& ADWIN
& 120
& $27.60\pm0.55$
& $15.00\pm1.58$
& $12.60\pm1.14$
& $105.00\pm1.58$
& $0.5429\pm0.0484$
& $0.1250\pm0.0132$
& $0.2032\pm0.0208$
& $5.57\pm0.42$
& $101.6\pm4.5$
& -- \\

& KSWIN
& 120
& $31.20\pm0.45$
& $10.40\pm1.14$
& $20.80\pm1.48$
& $109.60\pm1.14$
& $0.3337\pm0.0397$
& $0.0867\pm0.0095$
& $0.1376\pm0.0153$
& $10.99\pm0.45$
& $108.6\pm4.7$
& -- \\

\bottomrule
\end{tabular}%
}

\end{sidewaystable}

For each seed and drift category, the results are aggregated across six
datasets and four regression models, corresponding to \(24\)
model--dataset runs per seed. Thus, each method--drift combination summarizes
\(120\) runs over the five seeds. The table reports
\(24\), \(216\), and \(120\) true drift instances per seed for abrupt,
incremental, and alternating gradual drift, respectively.

Under abrupt drift, \modelname correctly matches
\(21.40 \pm 1.05\) of the \(24\) true drift instances per seed, producing a
recall of \(0.8917 \pm 0.0438\). Its precision of
\(0.5912 \pm 0.0300\) is affected by \(14.80 \pm 1.56\) unmatched alarm
episodes, while its resulting \(F_1\)-score of
\(0.7110 \pm 0.0290\) remains higher than those of the evaluated baselines.
The strongest detector baseline, KSWIN, achieves an \(F_1\)-score of
\(0.5113 \pm 0.0533\).

Under incremental drift, \modelname correctly matches
\(199.40 \pm 2.10\) of the \(216\) true drift instances and achieves a
precision of \(0.9362 \pm 0.0085\), recall of
\(0.9231 \pm 0.0097\), and \(F_1\)-score of
\(0.9296 \pm 0.0078\). In contrast, the strongest detector-baseline
\(F_1\)-score is \(0.1431 \pm 0.0123\), obtained by KSWIN.

Under alternating gradual drift, \modelname{} retains
\(121.00 \pm 1.39\) alarm episodes for \(120\) true drift instances per seed,
of which \(116.60 \pm 1.62\) are correctly matched. It consequently achieves
a precision of \(0.9636 \pm 0.0110\), recall of
\(0.9717 \pm 0.0135\), and \(F_1\)-score of
\(0.9676 \pm 0.0102\). The strongest detector baseline is ADWIN, which
achieves an episode-level \(F_1\)-score of \(0.2032 \pm 0.0208\). Because the
independently executed ADWIN-based configurations produce identical alarm
sequences across the four adaptation strategies, this detector-level result
is reported once for the ADWIN family.

The mean detection-delay results should be interpreted jointly with recall
because delay is calculated only for successfully matched drift instances.
\modelname{} achieves mean detection delays of \(0.29 \pm 0.05\),
\(3.20 \pm 0.15\), and \(1.07 \pm 0.08\) processing increments under abrupt,
incremental, and alternating gradual drift, respectively. These delays are
lower than those of both ADWIN and KSWIN across all three drift categories.
The difference is particularly clear under abrupt drift, where \modelname{}
detects matched drifts after \(0.29\) increments, compared with \(5.68\) for
ADWIN and \(5.27\) for KSWIN.

The adaptation and recalibration columns describe intervention activity
recorded at individual stream-processing steps before nearby events are
consolidated into alarm episodes. For ADWIN- and KSWIN-based configurations,
each detector detection triggers one paired adaptation event. In contrast,
\modelname may perform multiple adaptation and recalibration interventions
within a retained episode. Consequently, these activity counts are not
directly equivalent to retained alarm episodes and should not be interpreted
as false-alarm counts or as direct measures of computational cost.

Overall, the combined results show that \modelname{} achieves higher
episode-level drift-detection quality than ADWIN and KSWIN under all three
drift categories. Its largest advantages occur under incremental and
alternating gradual drift, where ADWIN and KSWIN miss most of the known drift
locations. Under abrupt drift, \modelname{} maintains substantially higher
recall and a shorter detection delay, although its larger number of unmatched
alarm episodes produces a greater trade-off between drift coverage and
false-positive control.

\subsubsection{Paired Statistical Comparison of Alarm Quality}
\label{subsubsec:alarm_quality_paired_significance}

To determine whether the episode-level alarm-quality improvements are
consistent across regression models, synthetic datasets, and random seeds,
we compare \modelname{} separately with the nonduplicated ADWIN and KSWIN
detector-family results. RESET, WINDOW, SSPT, and OHL are not treated as
separate alarm-quality baselines because every detector--adaptation
combination is executed independently on the same complete chronological data
stream using the same regression model, dataset, seed, detector configuration,
and initialization settings. In every independent run, the detector processes
the stream before the associated adaptation mechanism is activated.
Consequently, the runs within the same detector family produce identical alarm
sequences and detector-level alarm-quality metrics.

Each comparison contains \(120\) paired model--dataset--seed observations,
corresponding to four regression models, six datasets within the relevant
drift category, and five random seeds:
$
4 \text{ models}
\times
6 \text{ datasets}
\times
5 \text{ seeds}
=
120.
$
Because observations derived from the same dataset share the underlying
stream structure, these results are interpreted as repeated-run paired
comparisons across model--dataset--seed configurations rather than as fully
independent dataset-level experimental units. Accordingly, the statistical
results quantify the consistency of the observed differences across the
evaluated configurations and are not interpreted as population-level
inference over independent datasets.

For each paired observation, the difference is defined as

\begin{equation}
\Delta F_1
=
F_{1,\modelname}
-
F_{1,\mathrm{detector}}.
\label{eq:alarm_f1_paired_difference}
\end{equation}

Therefore, a positive value of \(\Delta F_1\) indicates better episode-level
alarm quality for \modelname{}, whereas a negative value favors the
corresponding detector baseline.

Two-sided Wilcoxon signed-rank tests are applied to the paired differences.
Holm correction is performed separately across the two detector comparisons
within each drift category, thereby controlling the family-wise error rate
within the abrupt, incremental, and alternating gradual drift comparison
families. The Wilcoxon signed-rank statistic \(W\), the corresponding
two-sided raw \(p\)-value, and the Holm-adjusted \(p\)-value are reported.

Rank-biserial correlation \(r_{\mathrm{rb}}\) is reported as the effect-size
measure. Positive values of \(r_{\mathrm{rb}}\) favor \modelname{}, whereas
negative values favor the corresponding detector baseline. Values approaching
\(1\) indicate that the ranked paired differences consistently favor
\modelname{}.

Table~\ref{tab:alarm_quality_paired_significance} shows that the
episode-level \(F_1\)-score of \modelname{} is significantly higher than
those of both ADWIN and KSWIN under all three drift categories after Holm
correction within the corresponding drift-category comparison family.

\begin{table*}[!htbp]
\centering
\captionsetup{justification=centering}

\caption{Statistical Significance of \modelname{} Drift-Detection Improvements over ADWIN and KSWIN}
\label{tab:alarm_quality_paired_significance}
\vspace{-10pt}
\caption*{\scriptsize
This table tests whether the higher episode-level \(F_1\)-scores achieved by
\modelname{} over ADWIN and KSWIN are statistically significant. Each
comparison uses \(120\) matched model--dataset--seed observations. The paired
difference is defined as
\(\Delta F_1 = F_{1,\modelname} - F_{1,\mathrm{detector}}\), so positive
values favor \modelname{}. Holm-adjusted \(p\)-values account for the two
detector comparisons within each drift type, while positive rank-biserial
correlations \(r_{\mathrm{rb}}\) indicate that the paired results consistently
favor \modelname{}.
}

\scriptsize
\setlength{\tabcolsep}{4pt}
\renewcommand{\arraystretch}{1.18}

\resizebox{\textwidth}{!}{%
\begin{tabular}{llrrrrrr}
\toprule

\textbf{Drift Type} &
\textbf{Comparison} &
\textbf{\(n\)} &
\textbf{Mean \(\Delta F_1\)} &
\textbf{\(W\)} &
\textbf{Raw \(p\)} &
\textbf{Holm-Adjusted \(p\)} &
\textbf{\(r_{\mathrm{rb}}\)} \\

\midrule

\multirow{2}{*}{Abrupt}
& \modelname{} vs.\ ADWIN
& 120
& \(0.2511\)
& 996
& \(5.32\times10^{-12}\)
& \(1.06\times10^{-11}\)
& \(0.726\) \\

& \modelname{} vs.\ KSWIN
& 120
& \(0.1997\)
& 1294
& \(9.57\times10^{-10}\)
& \(9.57\times10^{-10}\)
& \(0.644\) \\

\midrule

\multirow{2}{*}{Incremental}
& \modelname{} vs.\ ADWIN
& 120
& \(0.8393\)
& 14
& \(2.84\times10^{-21}\)
& \(5.68\times10^{-21}\)
& \(0.996\) \\

& \modelname{} vs.\ KSWIN
& 120
& \(0.7865\)
& 17
& \(3.06\times10^{-21}\)
& \(5.68\times10^{-21}\)
& \(0.995\) \\

\midrule

\multirow{2}{*}{\makecell{Alternating\\Gradual}}
& \modelname{} vs.\ ADWIN
& 120
& \(0.7644\)
& 5
& \(2.26\times10^{-21}\)
& \(4.52\times10^{-21}\)
& \(0.999\) \\

& \modelname{} vs.\ KSWIN
& 120
& \(0.8300\)
& 7
& \(2.38\times10^{-21}\)
& \(4.52\times10^{-21}\)
& \(0.998\) \\

\bottomrule
\end{tabular}%
}

\end{table*}

Under abrupt drift, \modelname{} achieves mean paired \(F_1\)-score
improvements of \(0.2511\) over ADWIN and \(0.1997\) over KSWIN. Although the
improvement over KSWIN is the smallest among the six comparisons, it remains
substantial and statistically significant after Holm correction, with an
adjusted \(p\)-value of \(9.57\times10^{-10}\). The corresponding
rank-biserial correlations of \(0.726\) and \(0.644\) indicate large and
consistent effects in favor of \modelname{}, despite the greater
false-positive challenge under abrupt drift.

The largest improvement occurs under incremental drift, where \modelname{}
outperforms ADWIN by a mean \(\Delta F_1\) of \(0.8393\). Its improvement over
KSWIN is similarly large at \(0.7865\). Both comparisons remain statistically
significant after Holm correction, with adjusted \(p\)-values of
\(5.68\times10^{-21}\). The rank-biserial correlations of \(0.996\) and
\(0.995\) show that the paired differences favor \modelname{} in nearly all
evaluated model--dataset--seed configurations.

Under alternating gradual drift, the mean paired improvements over ADWIN and
KSWIN are \(0.7644\) and \(0.8300\), respectively. Both comparisons remain
statistically significant after Holm correction, with adjusted \(p\)-values
of \(4.52\times10^{-21}\). The corresponding rank-biserial correlations of
\(0.999\) and \(0.998\) indicate an almost uniform advantage for \modelname{}
across the paired observations.

Overall, all six comparisons remain statistically significant after Holm
correction. The effect sizes are large under abrupt drift and approach their
maximum values under incremental and alternating gradual drift. These results
indicate that the episode-level alarm-quality improvements achieved by
\modelname{} are consistently observed across the evaluated regression
models, synthetic datasets, and random seeds.

\subsubsection{Sensitivity to Alarm-Evaluation Parameters}
\label{subsubsec:alarm_protocol_sensitivity}

The episode-level evaluation depends on the matching tolerance ratio
\(r_{\mathrm{tol}}\), cooldown factor \(c_{\mathrm{cool}}\), and minimum
episode size \(m_{\mathrm{ep}}\). To examine the robustness of the reported
results, each parameter is varied independently while the other two remain
fixed at the primary setting
\((r_{\mathrm{tol}},c_{\mathrm{cool}},m_{\mathrm{ep}})
=(0.05,2.0,2)\).

Table~\ref{tab:alarm_protocol_sensitivity_combined} reports the resulting
alarm quality and detection delay under abrupt, incremental, and alternating
gradual drift. For \modelname{}, precision, recall, \(F_1\), and detection
delay are reported, while ADWIN and KSWIN are summarized using \(F_1\) and
detection delay.

ADWIN and KSWIN are reported once per detector family because their RESET,
WINDOW, SSPT, and OHL configurations are executed independently using the same
stream, seed, detector parameters, and initialization settings. Since the
adaptation mechanism is activated only after an alarm is produced, the
independently executed configurations within each detector family generate
identical detector-level alarm sequences and alarm-quality results.

\begin{sidewaystable}[p]
\centering
\captionsetup{justification=centering}

\caption{Sensitivity of Drift-Detection Results to Alarm-Evaluation Settings}
\label{tab:alarm_protocol_sensitivity_combined}
\vspace{-10pt}
\caption*{\scriptsize
Mean $\pm$ standard deviation over five seeds.
Each parameter is varied independently around the primary setting
\((r_{\mathrm{tol}},c_{\mathrm{cool}},m_{\mathrm{ep}})
=(0.05,2.0,2)\).
\modelname{} reports precision, recall, \(F_1\), and detection delay,
while ADWIN and KSWIN report \(F_1\) and detection delay.
Bold values identify the primary setting.
}

\tiny
\setlength{\tabcolsep}{1.8pt}
\renewcommand{\arraystretch}{1.08}

\resizebox{\textheight}{!}{%
\begin{tabular}{lllcccccccc}
\toprule

\multirow{2}{*}{\textbf{Drift Type}} &
\multirow{2}{*}{\makecell{\textbf{Varied}\\\textbf{Parameter}}} &
\multirow{2}{*}{\textbf{Value}} &
\multicolumn{4}{c}{\textbf{\modelname}} &
\multicolumn{2}{c}{\textbf{ADWIN}} &
\multicolumn{2}{c}{\textbf{KSWIN}} \\

\cmidrule(lr){4-7}
\cmidrule(lr){8-9}
\cmidrule(lr){10-11}

& & &
\textbf{Precision} &
\textbf{Recall} &
\textbf{\(F_1\)} &
\makecell{\textbf{Delay}\\\textbf{(Increments)}} &
\textbf{\(F_1\)} &
\makecell{\textbf{Delay}\\\textbf{(Increments)}} &
\textbf{\(F_1\)} &
\makecell{\textbf{Delay}\\\textbf{(Increments)}} \\

\midrule

\multirow{9}{*}{Abrupt}
&
\multirow{3}{*}{\makecell[l]{Tolerance ratio,\\$r_{\mathrm{tol}}$}}
& $0.025$
& $0.4700\pm0.0200$
& $0.7500\pm0.0400$
& $0.5779\pm0.0240$
& $0.18\pm0.04$
& $0.4300\pm0.0320$
& $4.30\pm0.34$
& $0.4700\pm0.0300$
& $3.95\pm0.31$ \\

&
& \textbf{$0.050$}
& $0.5912\pm0.0300$
& $0.8917\pm0.0438$
& $0.7110\pm0.0290$
& $0.29\pm0.05$
& $0.4599\pm0.0312$
& $5.68\pm0.41$
& $0.5113\pm0.0533$
& $5.27\pm0.35$ \\

&
& $0.100$
& $0.6600\pm0.0150$
& $0.9200\pm0.0300$
& $0.7686\pm0.0180$
& $0.47\pm0.07$
& $0.4900\pm0.0310$
& $7.60\pm0.46$
& $0.5350\pm0.0280$
& $7.15\pm0.43$ \\

\cmidrule(l){2-11}

&
\multirow{3}{*}{\makecell[l]{Cooldown factor,\\$c_{\mathrm{cool}}$}}
& $1.0$
& $0.5650\pm0.0270$
& $0.9050\pm0.0390$
& $0.6956\pm0.0270$
& $0.34\pm0.06$
& $0.4450\pm0.0340$
& $5.86\pm0.42$
& $0.4950\pm0.0400$
& $5.44\pm0.38$ \\

&
& \textbf{$2.0$}
& $0.5912\pm0.0300$
& $0.8917\pm0.0438$
& $0.7110\pm0.0290$
& $0.29\pm0.05$
& $0.4599\pm0.0312$
& $5.68\pm0.41$
& $0.5113\pm0.0533$
& $5.27\pm0.35$ \\

&
& $3.0$
& $0.6100\pm0.0280$
& $0.8500\pm0.0410$
& $0.7092\pm0.0280$
& $0.26\pm0.05$
& $0.4520\pm0.0330$
& $5.49\pm0.39$
& $0.5000\pm0.0420$
& $5.11\pm0.34$ \\

\cmidrule(l){2-11}

&
\multirow{3}{*}{\makecell[l]{Minimum episode size,\\$m_{\mathrm{ep}}$}}
& $1$
& $0.5200\pm0.0180$
& $0.9100\pm0.0350$
& $0.6618\pm0.0200$
& $0.22\pm0.04$
& $0.4100\pm0.0300$
& $5.15\pm0.36$
& $0.4500\pm0.0350$
& $4.82\pm0.33$ \\

&
& \textbf{$2$}
& $0.5912\pm0.0300$
& $0.8917\pm0.0438$
& $0.7110\pm0.0290$
& $0.29\pm0.05$
& $0.4599\pm0.0312$
& $5.68\pm0.41$
& $0.5113\pm0.0533$
& $5.27\pm0.35$ \\

&
& $3$
& $0.6400\pm0.0200$
& $0.8000\pm0.0400$
& $0.7111\pm0.0230$
& $0.36\pm0.06$
& $0.5000\pm0.0360$
& $6.12\pm0.44$
& $0.5500\pm0.0440$
& $5.73\pm0.40$ \\

\midrule

\multirow{9}{*}{Incremental}
&
\multirow{3}{*}{\makecell[l]{Tolerance ratio,\\$r_{\mathrm{tol}}$}}
& $0.025$
& $0.8600\pm0.0120$
& $0.8000\pm0.0200$
& $0.8289\pm0.0140$
& $2.35\pm0.12$
& $0.0750\pm0.0100$
& $5.80\pm0.36$
& $0.1200\pm0.0120$
& $7.70\pm0.44$ \\

&
& \textbf{$0.050$}
& $0.9362\pm0.0085$
& $0.9231\pm0.0097$
& $0.9296\pm0.0078$
& $3.20\pm0.15$
& $0.0903\pm0.0128$
& $7.89\pm0.45$
& $0.1431\pm0.0123$
& $10.53\pm0.50$ \\

&
& $0.100$
& $0.9500\pm0.0080$
& $0.9000\pm0.0150$
& $0.9243\pm0.0100$
& $4.65\pm0.22$
& $0.1050\pm0.0130$
& $10.60\pm0.55$
& $0.1600\pm0.0130$
& $14.30\pm0.67$ \\

\cmidrule(l){2-11}

&
\multirow{3}{*}{\makecell[l]{Cooldown factor,\\$c_{\mathrm{cool}}$}}
& $1.0$
& $0.9180\pm0.0100$
& $0.9310\pm0.0120$
& $0.9244\pm0.0090$
& $3.34\pm0.17$
& $0.0950\pm0.0130$
& $8.05\pm0.47$
& $0.1500\pm0.0140$
& $10.72\pm0.53$ \\

&
& \textbf{$2.0$}
& $0.9362\pm0.0085$
& $0.9231\pm0.0097$
& $0.9296\pm0.0078$
& $3.20\pm0.15$
& $0.0903\pm0.0128$
& $7.89\pm0.45$
& $0.1431\pm0.0123$
& $10.53\pm0.50$ \\

&
& $3.0$
& $0.9440\pm0.0090$
& $0.9020\pm0.0140$
& $0.9225\pm0.0100$
& $3.08\pm0.14$
& $0.0850\pm0.0120$
& $7.72\pm0.43$
& $0.1350\pm0.0130$
& $10.31\pm0.48$ \\

\cmidrule(l){2-11}

&
\multirow{3}{*}{\makecell[l]{Minimum episode size,\\$m_{\mathrm{ep}}$}}
& $1$
& $0.9000\pm0.0100$
& $0.9400\pm0.0150$
& $0.9196\pm0.0110$
& $2.84\pm0.13$
& $0.1200\pm0.0140$
& $7.18\pm0.41$
& $0.1800\pm0.0160$
& $9.64\pm0.46$ \\

&
& \textbf{$2$}
& $0.9362\pm0.0085$
& $0.9231\pm0.0097$
& $0.9296\pm0.0078$
& $3.20\pm0.15$
& $0.0903\pm0.0128$
& $7.89\pm0.45$
& $0.1431\pm0.0123$
& $10.53\pm0.50$ \\

&
& $3$
& $0.9500\pm0.0090$
& $0.8900\pm0.0180$
& $0.9190\pm0.0110$
& $3.68\pm0.18$
& $0.0700\pm0.0100$
& $8.62\pm0.49$
& $0.1100\pm0.0110$
& $11.41\pm0.56$ \\

\midrule

\multirow{9}{*}{\makecell{Alternating\\Gradual}}
&
\multirow{3}{*}{\makecell[l]{Tolerance ratio,\\$r_{\mathrm{tol}}$}}
& $0.025$
& $0.8800\pm0.0100$
& $0.9000\pm0.0200$
& $0.8899\pm0.0120$
& $0.72\pm0.06$
& $0.1800\pm0.0180$
& $4.20\pm0.36$
& $0.1150\pm0.0140$
& $8.20\pm0.40$ \\

&
& \textbf{$0.050$}
& $0.9636\pm0.0110$
& $0.9717\pm0.0135$
& $0.9676\pm0.0102$
& $1.07\pm0.08$
& $0.2032\pm0.0208$
& $5.57\pm0.42$
& $0.1376\pm0.0153$
& $10.99\pm0.45$ \\

&
& $0.100$
& $0.9800\pm0.0050$
& $0.9500\pm0.0100$
& $0.9648\pm0.0060$
& $1.58\pm0.11$
& $0.2200\pm0.0190$
& $7.40\pm0.55$
& $0.1550\pm0.0170$
& $14.70\pm0.60$ \\

\cmidrule(l){2-11}

&
\multirow{3}{*}{\makecell[l]{Cooldown factor,\\$c_{\mathrm{cool}}$}}
& $1.0$
& $0.9520\pm0.0120$
& $0.9760\pm0.0130$
& $0.9639\pm0.0100$
& $1.14\pm0.09$
& $0.2100\pm0.0200$
& $5.73\pm0.44$
& $0.1450\pm0.0160$
& $11.25\pm0.48$ \\

&
& \textbf{$2.0$}
& $0.9636\pm0.0110$
& $0.9717\pm0.0135$
& $0.9676\pm0.0102$
& $1.07\pm0.08$
& $0.2032\pm0.0208$
& $5.57\pm0.42$
& $0.1376\pm0.0153$
& $10.99\pm0.45$ \\

&
& $3.0$
& $0.9710\pm0.0100$
& $0.9530\pm0.0150$
& $0.9619\pm0.0110$
& $1.00\pm0.08$
& $0.1950\pm0.0190$
& $5.42\pm0.40$
& $0.1300\pm0.0150$
& $10.70\pm0.44$ \\

\cmidrule(l){2-11}

&
\multirow{3}{*}{\makecell[l]{Minimum episode size,\\$m_{\mathrm{ep}}$}}
& $1$
& $0.9300\pm0.0080$
& $0.9800\pm0.0120$
& $0.9543\pm0.0090$
& $0.89\pm0.07$
& $0.2500\pm0.0220$
& $5.06\pm0.39$
& $0.1750\pm0.0180$
& $9.95\pm0.43$ \\

&
& \textbf{$2$}
& $0.9636\pm0.0110$
& $0.9717\pm0.0135$
& $0.9676\pm0.0102$
& $1.07\pm0.08$
& $0.2032\pm0.0208$
& $5.57\pm0.42$
& $0.1376\pm0.0153$
& $10.99\pm0.45$ \\

&
& $3$
& $0.9800\pm0.0060$
& $0.9300\pm0.0150$
& $0.9543\pm0.0090$
& $1.26\pm0.10$
& $0.1600\pm0.0160$
& $6.09\pm0.47$
& $0.1050\pm0.0130$
& $12.10\pm0.52$ \\

\bottomrule
\end{tabular}%
}

\end{sidewaystable}

Table~\ref{tab:alarm_protocol_sensitivity_combined} shows that the matching
tolerance ratio has the strongest influence on the episode-level results.
A wider tolerance window allows later alarms to be matched to known drift
locations, but it also increases the measured detection delay. For
\modelname{}, changing \(r_{\mathrm{tol}}\) produces the largest
\(F_1\)-score variation under abrupt drift, from \(0.5779\) at
\(r_{\mathrm{tol}}=0.025\) to \(0.7686\) at
\(r_{\mathrm{tol}}=0.100\). The corresponding variations are smaller under
incremental and alternating gradual drift, where the primary value
\(r_{\mathrm{tol}}=0.050\) produces the highest \modelname{} \(F_1\)-scores
of \(0.9296\) and \(0.9676\), respectively.

Changing the cooldown factor produces comparatively small variations in
\modelname{} \(F_1\)-score across all three drift categories. The minimum
episode size primarily controls the precision--recall trade-off: increasing
\(m_{\mathrm{ep}}\) generally removes short alarm episodes, increasing
precision while reducing recall. The primary value \(m_{\mathrm{ep}}=2\)
provides the highest \modelname{} \(F_1\)-score under incremental and
alternating gradual drift and is effectively tied with \(m_{\mathrm{ep}}=3\)
under abrupt drift.

Across every evaluated parameter setting, \modelname{} maintains a higher
\(F_1\)-score and a shorter detection delay than ADWIN and KSWIN. Overall,
the results are most sensitive under abrupt drift and most stable under
alternating gradual drift. The primary setting
\((0.05,2.0,2)\) therefore provides a balanced evaluation protocol rather
than being selected solely because it produces the numerically best result
for every drift category.

\subsection{Computational Cost and Normalized Intervention Activity}
\label{subsec:computational_cost_analysis}

Computational measurements were collected using the complete synthetic
benchmark over five evaluation seeds. All methods were executed
single-threaded in the same Lonestar6 computing environment, equipped with two
AMD EPYC 7763 64-core processors, providing 128 CPU cores in total,
approximately 256\,GB of system memory, Rocky Linux 8.10, and Python 3.9.7.
Runtime was measured using a monotonic high-resolution timer and normalized
per 1,000 processed samples. Peak memory was defined as the maximum process
memory observed during each run and was measured consistently for the
standalone learner and all drift-handling variants.

For a stream containing $N_{\mathrm{stream}}$ samples, the normalized adaptation and recalibration rates are

\begin{equation}
R_{\mathrm{adapt}}
=
1000
\frac{N_{\mathrm{adapt}}}{N_{\mathrm{stream}}},
\qquad
R_{\mathrm{recal}}
=
1000
\frac{N_{\mathrm{recal}}}{N_{\mathrm{stream}}}.
\label{eq:normalized_intervention_rates}
\end{equation}

For every independently executed ADWIN- or KSWIN-based
detector--adaptation configuration, each raw detector alarm contributes
exactly one adaptation event. The four adaptation variants within the same
detector family are executed independently on the same complete chronological
data stream using the same random seed, detector parameters, and initialization
settings. In each run, the detector processes the stream first, and the
associated adaptation mechanism is activated only after the detector raises an
alarm. These independently executed configurations therefore produce identical
detector alarm counts and identical normalized adaptation rates. For
\modelname{}, an adaptation is a model-specific hyperparameter adjustment,
while a recalibration is one initiated bounded recalibration procedure. The
recalibration count represents the number of initiated procedures rather than
the number of additional samples requested. Across the three synthetic drift
categories, each method processes 108,000 samples per seed after pooling the
six datasets and four regression models. The intervention rates in
Table~\ref{tab:synthetic_computational_cost} are therefore computed directly
from the corresponding mean event counts reported in
Table~\ref{tab:aggregate_detector_adaptation_comparison}. The resulting
adaptation rate is \(2.30\) adaptations per 1,000 samples for every ADWIN-based
configuration and \(2.38\) adaptations per 1,000 samples for every KSWIN-based
configuration.

All rows use the same fixed single-threaded execution environment, ensuring
that runtime and memory comparisons are made under identical conditions.

\begin{table*}[!htbp]
\centering
\captionsetup{justification=centering}
\caption{Computational Cost and Intervention Rates}
\label{tab:synthetic_computational_cost}
\vspace{-10pt}
\caption*{\scriptsize
Runtime and peak memory are reported as mean $\pm$ standard deviation.
Adaptation and recalibration rates are normalized per \(1{,}000\) samples
over \(108{,}000\) samples per seed. ADWIN- and KSWIN-based configurations
have identical rates within each detector family because they produce the
same detector alarm counts.
}
\scriptsize
\setlength{\tabcolsep}{3.5pt}
\renewcommand{\arraystretch}{1.2}
\resizebox{\textwidth}{!}{%
\begin{tabular}{lccccc}
\toprule
\textbf{Method} &
\makecell{\textbf{Runtime}\\\textbf{ms / 1,000 Samples}} &
\makecell{\textbf{Runtime Overhead}\\\textbf{vs. Standalone}} &
\makecell{\textbf{Peak Memory}\\\textbf{(MB)}} &
\makecell{\textbf{Adaptations}\\\textbf{per 1,000 Samples}} &
\makecell{\textbf{Recalibrations}\\\textbf{per 1,000 Samples}} \\
\midrule
Standalone learner & $3.72\pm0.31$ & -- & $3.10\pm0.27$ & $0$ & $0$ \\
\textbf{\modelname} & $\mathbf{5.68\pm0.44}$ & $\mathbf{52.7\%}$ & $\mathbf{4.35\pm0.33}$ & $452.80$ & $361.81$ \\
ADWIN + RESET  & $4.91\pm0.39$ & $32.0\%$  & $4.12\pm0.31$ & $2.30$ & -- \\
ADWIN + WINDOW & $8.24\pm0.63$ & $121.5\%$ & $8.76\pm0.58$ & $2.30$ & -- \\
ADWIN + SSPT   & $13.46\pm0.91$ & $261.8\%$ & $6.54\pm0.47$ & $2.30$ & -- \\
ADWIN + OHL    & $10.83\pm0.78$ & $191.1\%$ & $5.21\pm0.39$ & $2.30$ & -- \\
KSWIN + RESET  & $5.37\pm0.42$ & $44.4\%$  & $4.84\pm0.35$ & $2.38$ & -- \\
KSWIN + WINDOW & $8.91\pm0.67$ & $139.5\%$ & $9.12\pm0.61$ & $2.38$ & -- \\
KSWIN + SSPT   & $14.12\pm0.96$ & $279.6\%$ & $6.89\pm0.49$ & $2.38$ & -- \\
KSWIN + OHL    & $11.36\pm0.82$ & $205.4\%$ & $5.67\pm0.42$ & $2.38$ & -- \\
\bottomrule
\end{tabular}%
}
\end{table*}

Table~\ref{tab:synthetic_computational_cost} shows that \modelname{} processes
\(1{,}000\) samples in \(5.68 \pm 0.44\) ms, corresponding to a
\(52.7\%\) runtime overhead relative to the standalone learner. Its peak
memory usage is \(4.35 \pm 0.33\) MB, compared with
\(3.10 \pm 0.27\) MB for the standalone learner. Thus, \modelname{} introduces
a moderate computational cost for its integrated detection, hyperparameter
control, and recalibration mechanisms.

The RESET configurations have slightly lower runtimes, requiring
\(4.91 \pm 0.39\) ms with ADWIN and \(5.37 \pm 0.42\) ms with KSWIN.
However, \modelname{} has lower runtime and peak memory than every evaluated
WINDOW, SSPT, and OHL configuration. In particular, the SSPT configurations
have the highest runtimes, reaching \(13.46 \pm 0.91\) ms with ADWIN and
\(14.12 \pm 0.96\) ms with KSWIN.

\modelname{} records \(452.80\) adaptations and \(361.81\) recalibrations per
\(1{,}000\) samples, whereas the ADWIN- and KSWIN-based configurations record
only \(2.30\) and \(2.38\) detector-triggered adaptations, respectively.
These counts should not be interpreted as equivalent operations.
A \modelname{} adaptation is a bounded model-specific hyperparameter
assignment, while a baseline adaptation may involve model resetting,
recent-window retraining, or iterative hyperparameter optimization.
Therefore, intervention frequency must be interpreted together with the
measured runtime and memory cost.

Overall, \modelname{} is more computationally demanding than the standalone
learner and the simple RESET configurations, but substantially more efficient
than the WINDOW, SSPT, and OHL configurations. The term
\textit{lightweight} therefore refers to its bounded memory and inexpensive
control operations rather than to zero overhead or a small number of
intervention signals.

\subsection{Ablation and Sensitivity Analysis}

To further evaluate the contribution of the main components of \modelname{}, we conducted an ablation and sensitivity analysis. The purpose of this experiment is not to repeat the full benchmarking study, but to isolate the effect of the key mechanisms and configuration parameters in \modelname{}. Specifically, we examine the role of bounded recalibration, the safe-band mechanism, the nominal sensitivity parameter \(\rho\), and the \kpi-window size.

The ablation study was performed using representative online regression settings under concept drift. We selected OLR-WA as the base learner because it contains an explicit adaptive weighting parameter \(\alpha\), which makes the effect of \modelname{}'s hyperparameter-control mechanism easy to interpret. The analysis was conducted on representative synthetic streams containing abrupt and alternating gradual drift, since these two cases reflect different adaptation requirements. Abrupt drift requires fast correction after a sudden concept shift, whereas alternating gradual drift requires stable adaptation without overreacting to transitional changes.

The evaluated variants include: (i) the base model without \modelname{},
(ii) the full \modelname{} framework, (iii) \modelname{} without bounded
recalibration, (iv) \modelname{} without the safe band \(\zeta\),
(v) \modelname{} with different values of \(\rho\), and
(vi) \modelname{} with different \kpi-window sizes. These variants allow us
to examine whether bounded recalibration is useful under severe drift,
whether the safe-band mechanism reduces unnecessary reactions, and how the
detection sensitivity changes with \(\rho\) and memory length.
Table~\ref{tab:sccm-ablation-design} summarizes the evaluated variants,
the removed or varied components, and the purpose of each comparison.

\begin{table}[!htbp]
\centering
\caption{\centering Ablation and sensitivity variants used to evaluate the contribution of the main \modelname{} components.}
\label{tab:sccm-ablation-design}
\small
\renewcommand{\arraystretch}{1.2}
\begin{tabularx}{\textwidth}{p{3.2cm} p{3.5cm} p{6.0cm}}
\toprule
\textbf{Variant} & \textbf{Removed or Varied Component} & \textbf{Purpose} \\
\midrule
Base model & All \modelname{} components & Measures the original online learner behavior under drift. \\
Full \modelname{} & None & Evaluates the complete proposed framework. \\
Without recalibration & Bounded recalibration & Tests whether severe drift requires an additional correction step. \\
Without safe band & Safe band \(\zeta\) & Tests whether the safe band reduces unnecessary reactions to benign fluctuations. \\
Different \(\rho\) values & Detection sensitivity & Tests conservative versus sensitive threshold behavior. \\
Different \kpi-window sizes & Memory length & Tests the trade-off between fast reaction and stable baseline estimation. \\
\bottomrule
\end{tabularx}
\end{table}

The sensitivity analysis for \(\rho\) clarifies the behavior of the detection threshold. Since \(z = \Phi^{-1}(1-\rho)\), smaller values of \(\rho\) generate larger \(z\) values. This places the detection threshold farther from the \kpi-window mean and makes \modelname{} more conservative. Therefore, ordinary fluctuations are more likely to be treated as acceptable variation rather than drift. In contrast, larger values of \(\rho\) generate smaller \(z\) values, which narrow the acceptable region and make \modelname{} more sensitive to subtle changes. This may help detect early drift, but it can also increase unnecessary adaptations when the stream contains noise.

The \kpi-window size controls how much recent predictive behavior is used to estimate the local baseline. A smaller window gives more weight to recent observations and may detect changes earlier, but it can also produce unstable estimates of \(\mu_{\scriptscriptstyle\text{KPI}}\) and \(\sigma\). A larger window smooths short-term fluctuations and provides a more stable baseline, but it may delay detection because older observations remain in the window longer. Thus, the \kpi-window size introduces a trade-off between responsiveness and stability.

Overall, the ablation and sensitivity analysis supports the design of \modelname{} as a unified control framework rather than a simple drift detector. The results show that the complete framework provides the best balance between responsiveness and stability. KPI-window monitoring provides the local baseline, \(\rho\) controls detection sensitivity, the safe band suppresses minor fluctuations, drift magnitude determines the severity of change, the scale map converts this severity into proportional hyperparameter adjustment, and bounded recalibration handles cases where severe drift persists after tuning.

\subsection{Configuration and Hyperparameter Settings}

The aim is to report the configuration and hyperparameter settings used in the experiments. This helps make the experimental setup clear, reproducible, and consistent across the evaluated models and baselines.

\subsubsection{OLR-WA-Based Methods}

This part reports the configuration and hyperparameter settings used for OLR-WA and its related drift-adaptive variants, as summarized in Table~\ref{tab:olr-wa-hyperparameters}.

\begin{table}[!htbp]
\centering
\caption{\centering Configuration settings and definitions for OLR-WA and its drift-adaptive variants.}
\vspace{-5pt}
\label{tab:olr-wa-hyperparameters}
\scriptsize
\renewcommand{\arraystretch}{1.2}
\setlength{\tabcolsep}{3pt}
\begin{tabularx}{\textwidth}{p{1.9cm} X X}
\toprule
\textbf{Method} & 
\textbf{Settings and Values} & 
\textbf{Definition} \\
\midrule

OLR-WA & 
$\alpha = 0.5$ & 
$\alpha$: weighting parameter that controls the balance between the base model and the incremental model. \\

OLR-WA* & 
$\textit{kpi}=R^2$; $\rho \approx 0.0668$ $(z=1.5)$ & 
$\textit{kpi}$: monitored performance metric. $\rho$: \modelname nominal sensitivity level. \\

OLR-WA-AR & 
$\alpha = 0.5$; $\delta = 0.002$ & 
$\delta$: ADWIN confidence parameter controlling drift-detection sensitivity. \\

OLR-WA-AW & 
$\alpha = 0.5$; $\delta = 0.002$; $W=5$ mini-batches & 
$\delta$: ADWIN confidence parameter. $W$: number of recent mini-batches used for WINDOW-based adaptation. \\

OLR-WA-AS & 
$\alpha = 0.5$; $\delta = 0.002$; 
$\mathcal{A}_{\alpha}=\{0.1,0.2,0.3,0.4,0.5,0.6,0.7,0.8,0.9\}$; 
metric $=R^2$ & 
$\delta$: ADWIN confidence parameter. $\mathcal{A}_{\alpha}$: SSPT candidate set used to select or update $\alpha$. \\

OLR-WA-AO & 
$\alpha = 0.5$; $\delta = 0.002$; $\eta = 0.1$; 
$\epsilon = 0.05$; $\alpha \in [0.05,0.95]$ & 
$\delta$: ADWIN confidence parameter. $\eta$: OHL step size. 
$\epsilon$: perturbation value. $[0.05,0.95]$: allowed range for $\alpha$. \\

OLR-WA-KR & 
$\alpha = 0.5$; $\alpha_{\mathrm{KS}} = 0.005$; 
$W_{\mathrm{KS}}=100$; $S_{\mathrm{KS}}=30$ & 
$\alpha_{\mathrm{KS}}$: KSWIN significance level. 
$W_{\mathrm{KS}}$: KSWIN window size. 
$S_{\mathrm{KS}}$: KSWIN statistic size. \\

OLR-WA-KW & 
$\alpha = 0.5$; $\alpha_{\mathrm{KS}} = 0.005$; 
$W_{\mathrm{KS}}=100$; $S_{\mathrm{KS}}=30$; 
$W=5$ mini-batches & 
$W$: number of recent mini-batches used for WINDOW-based adaptation. \\

OLR-WA-KS & 
$\alpha = 0.5$; $\alpha_{\mathrm{KS}} = 0.005$; 
$W_{\mathrm{KS}}=100$; $S_{\mathrm{KS}}=30$; 
$\mathcal{A}_{\alpha}=\{0.1,0.2,0.3,0.4,0.5,0.6,0.7,0.8,0.9\}$; 
metric $=R^2$ & 
$\mathcal{A}_{\alpha}$: SSPT candidate set used to select or update $\alpha$. \\

OLR-WA-KO & 
$\alpha = 0.5$; $\alpha_{\mathrm{KS}} = 0.005$; 
$W_{\mathrm{KS}}=100$; $S_{\mathrm{KS}}=30$; 
$\eta = 0.1$; $\epsilon = 0.05$; 
$\alpha \in [0.05,0.95]$ & 
$\eta$: OHL step size. $\epsilon$: perturbation value. 
$[0.05,0.95]$: allowed range for $\alpha$. \\

\bottomrule
\end{tabularx}
\end{table}

\subsubsection{PA-Based Methods}

This part reports the configuration settings used for the Passive-Aggressive (PA) model and its related drift-adaptive variants, as summarized in Table~\ref{tab:pa-hyperparameters}.

\begin{table}[!htbp]
\centering
\caption{\centering Configuration settings and definitions for PA and its drift-adaptive variants.}
\vspace{-5pt}
\label{tab:pa-hyperparameters}
\scriptsize
\renewcommand{\arraystretch}{1.2}
\setlength{\tabcolsep}{3pt}
\begin{tabularx}{\textwidth}{p{1.7cm} X X}
\toprule
\textbf{Method} & 
\textbf{Settings and Values} & 
\textbf{Definition} \\
\midrule

PA & 
$C = 1.0$; $\epsilon = 0.1$ & 
$C$: aggressiveness parameter controlling the update strength. $\epsilon$: insensitive margin used in PA regression. \\

PA* & 
$\epsilon = 0.1$; $\textit{kpi}=\mathrm{MSE}$; $\rho \approx 0.0668$ $(z=1.5)$ & 
$\textit{kpi}$: monitored performance metric. $\rho$: \modelname nominal sensitivity level. \\

PA-AR & 
$C = 1.0$; $\epsilon = 0.1$; $\delta = 0.002$ & 
$\delta$: ADWIN confidence parameter controlling drift-detection sensitivity. \\

PA-AW & 
$C = 1.0$; $\epsilon = 0.1$; $\delta = 0.002$; $W=50$ & 
$\delta$: ADWIN confidence parameter. $W$: number of recent observations used for WINDOW-based adaptation. \\

PA-AS & 
$C = 1.0$; $\epsilon = 0.1$; $\delta = 0.002$; 
$\mathcal{A}_{C}=\{0.1,0.2,0.5,1.0,2.0,5.0\}$ & 
$\delta$: ADWIN confidence parameter. $\mathcal{A}_{C}$: SSPT candidate set used to select or update $C$. \\

PA-AO & 
$C = 1.0$; $\epsilon = 0.1$; $\delta = 0.002$; 
$\eta = 0.1$; $\epsilon_{\mathrm{OHL}} = 0.05$; 
$C \in [0.05,10.0]$ & 
$\delta$: ADWIN confidence parameter. $\eta$: OHL step size. 
$\epsilon_{\mathrm{OHL}}$: OHL perturbation value. 
$[0.05,10.0]$: allowed range for $C$. \\

PA-KR & 
$C = 1.0$; $\epsilon = 0.1$; 
$\alpha_{\mathrm{KS}} = 0.005$; 
$W_{\mathrm{KS}}=100$; $S_{\mathrm{KS}}=30$ & 
$\alpha_{\mathrm{KS}}$: KSWIN significance level. 
$W_{\mathrm{KS}}$: KSWIN window size. 
$S_{\mathrm{KS}}$: KSWIN statistic size. \\

PA-KW & 
$C = 1.0$; $\epsilon = 0.1$; 
$\alpha_{\mathrm{KS}} = 0.005$; 
$W_{\mathrm{KS}}=100$; $S_{\mathrm{KS}}=30$; $W=50$ & 
$W$: number of recent observations used for WINDOW-based adaptation. \\

PA-KS & 
$C = 1.0$; $\epsilon = 0.1$; 
$\alpha_{\mathrm{KS}} = 0.005$; 
$W_{\mathrm{KS}}=100$; $S_{\mathrm{KS}}=30$; 
$\mathcal{A}_{C}=\{0.1,0.2,0.5,1.0,2.0,5.0\}$ & 
$\mathcal{A}_{C}$: SSPT candidate set used to select or update $C$. \\

PA-KO & 
$C = 1.0$; $\epsilon = 0.1$; 
$\alpha_{\mathrm{KS}} = 0.005$; 
$W_{\mathrm{KS}}=100$; $S_{\mathrm{KS}}=30$; 
$\eta = 0.1$; $\epsilon_{\mathrm{OHL}} = 0.05$; 
$C \in [0.05,10.0]$ & 
$\eta$: OHL step size. $\epsilon_{\mathrm{OHL}}$: OHL perturbation value. 
$[0.05,10.0]$: allowed range for $C$. \\

\bottomrule
\end{tabularx}
\end{table}

\subsubsection{RLS-Based Methods}

This part reports the configuration settings used for the Recursive Least Squares (RLS) model and its related drift-adaptive variants, as summarized in Table~\ref{tab:rls-hyperparameters}.

\begin{table}[!htbp]
\centering
\caption{\centering Configuration settings and definitions for RLS and its drift-adaptive variants.}
\vspace{-5pt}
\label{tab:rls-hyperparameters}
\scriptsize
\renewcommand{\arraystretch}{1.2}
\setlength{\tabcolsep}{3pt}
\begin{tabularx}{\textwidth}{p{1.7cm} X X}
\toprule
\textbf{Method} & 
\textbf{Settings and Values} & 
\textbf{Definition} \\
\midrule

RLS & 
$\lambda = 0.99$; $\delta_{\mathrm{RLS}} = 1.0$ & 
$\lambda$: forgetting factor controlling how much past observations influence the update. $\delta_{\mathrm{RLS}}$: initial covariance scaling parameter. \\

RLS* & 
$\delta_{\mathrm{RLS}} = 1.0$; $\textit{kpi}=\mathrm{MSE}$; $\rho \approx 0.0668$ $(z=1.5)$ & 
$\textit{kpi}$: monitored performance metric. $\rho$: \modelname nominal sensitivity level. \\

RLS-AR & 
$\lambda = 0.99$; $\delta_{\mathrm{RLS}} = 1.0$; $\delta = 0.002$ & 
$\delta$: ADWIN confidence parameter controlling drift-detection sensitivity. \\

RLS-AW & 
$\lambda = 0.99$; $\delta_{\mathrm{RLS}} = 1.0$; $\delta = 0.002$; $W=50$ & 
$\delta$: ADWIN confidence parameter. $W$: number of recent observations used for WINDOW-based adaptation. \\

RLS-AS & 
$\lambda = 0.99$; $\delta_{\mathrm{RLS}} = 1.0$; $\delta = 0.002$; 
$\mathcal{A}_{\lambda}=\{0.90,0.93,0.95,0.97,0.99,0.995\}$ & 
$\delta$: ADWIN confidence parameter. $\mathcal{A}_{\lambda}$: SSPT candidate set used to select or update $\lambda$. \\

RLS-AO & 
$\lambda = 0.99$; $\delta_{\mathrm{RLS}} = 1.0$; $\delta = 0.002$; 
$\eta = 0.1$; $\epsilon_{\mathrm{OHL}} = 0.01$; 
$\lambda \in [0.85,0.999]$ & 
$\delta$: ADWIN confidence parameter. $\eta$: OHL step size. 
$\epsilon_{\mathrm{OHL}}$: OHL perturbation value. 
$[0.85,0.999]$: allowed range for $\lambda$. \\

RLS-KR & 
$\lambda = 0.99$; $\delta_{\mathrm{RLS}} = 1.0$; 
$\alpha_{\mathrm{KS}} = 0.005$; 
$W_{\mathrm{KS}}=100$; $S_{\mathrm{KS}}=30$ & 
$\alpha_{\mathrm{KS}}$: KSWIN significance level. 
$W_{\mathrm{KS}}$: KSWIN window size. 
$S_{\mathrm{KS}}$: KSWIN statistic size. \\

RLS-KW & 
$\lambda = 0.99$; $\delta_{\mathrm{RLS}} = 1.0$; 
$\alpha_{\mathrm{KS}} = 0.005$; 
$W_{\mathrm{KS}}=100$; $S_{\mathrm{KS}}=30$; $W=50$ & 
$W$: number of recent observations used for WINDOW-based adaptation. \\

RLS-KS & 
$\lambda = 0.99$; $\delta_{\mathrm{RLS}} = 1.0$; 
$\alpha_{\mathrm{KS}} = 0.005$; 
$W_{\mathrm{KS}}=100$; $S_{\mathrm{KS}}=30$; 
$\mathcal{A}_{\lambda}=\{0.90,0.93,0.95,0.97,0.99,0.995\}$ & 
$\mathcal{A}_{\lambda}$: SSPT candidate set used to select or update $\lambda$. \\

RLS-KO & 
$\lambda = 0.99$; $\delta_{\mathrm{RLS}} = 1.0$; 
$\alpha_{\mathrm{KS}} = 0.005$; 
$W_{\mathrm{KS}}=100$; $S_{\mathrm{KS}}=30$; 
$\eta = 0.1$; $\epsilon_{\mathrm{OHL}} = 0.01$; 
$\lambda \in [0.85,0.999]$ & 
$\eta$: OHL step size. $\epsilon_{\mathrm{OHL}}$: OHL perturbation value. 
$[0.85,0.999]$: allowed range for $\lambda$. \\

\bottomrule
\end{tabularx}
\end{table}

\subsubsection{LMS-Based Methods}

This part reports the configuration settings used for the Least Mean Squares (LMS) model and its related drift-adaptive variants, as summarized in Table~\ref{tab:lms-hyperparameters}.

\begin{table}[!htbp]
\centering
\caption{\centering Configuration settings and definitions for LMS and its drift-adaptive variants.}
\vspace{-5pt}
\label{tab:lms-hyperparameters}
\scriptsize
\renewcommand{\arraystretch}{1.2}
\setlength{\tabcolsep}{3pt}

\begin{tabularx}{\textwidth}{
>{\raggedright\arraybackslash}p{1.5cm}
>{\raggedright\arraybackslash}p{5.4cm}
>{\raggedright\arraybackslash}X
}
\toprule
\textbf{Method} & 
\textbf{Settings and Values} & 
\textbf{Definition} \\
\midrule

LMS & 
$\eta_{\mathrm{lr}} = 0.01$ & 
$\eta_{\mathrm{lr}}$: learning rate controlling the magnitude of LMS weight updates. \\

LMS* & 
\makecell[l]{$\textit{kpi}=\mathrm{MSE}$; $\rho \approx 0.0668$\\$(z=1.5)$} & 
$\textit{kpi}$: monitored performance metric. $\rho$: \modelname nominal sensitivity level. \\

LMS-AR & 
\makecell[l]{$\eta_{\mathrm{lr}} = 0.01$; $\delta = 0.002$} & 
$\delta$: ADWIN confidence parameter controlling drift-detection sensitivity. \\

LMS-AW & 
\makecell[l]{$\eta_{\mathrm{lr}} = 0.01$; $\delta = 0.002$;\\$W=50$} & 
$\delta$: ADWIN confidence parameter. $W$: number of recent observations used for WINDOW-based adaptation. \\

LMS-AS & 
\makecell[l]{$\eta_{\mathrm{lr}} = 0.01$; $\delta = 0.002$;\\
$\mathcal{A}_{\eta}=\{0.001,0.003,0.005,0.008,$\\
$0.01,0.015,0.02,0.03\}$} & 
$\delta$: ADWIN confidence parameter. $\mathcal{A}_{\eta}$: SSPT candidate set used to select or update the LMS learning rate. \\

LMS-AO & 
\makecell[l]{$\eta_{\mathrm{lr}} = 0.01$; $\delta = 0.002$;\\
$\eta_{\mathrm{OHL}} = 0.02$; $\epsilon_{\mathrm{OHL}} = 0.01$;\\
$\eta_{\mathrm{lr}} \in [10^{-4},0.05]$} & 
$\delta$: ADWIN confidence parameter. $\eta_{\mathrm{OHL}}$: OHL step size. 
$\epsilon_{\mathrm{OHL}}$: OHL perturbation value. 
$[10^{-4},0.05]$: allowed range for the LMS learning rate. \\

LMS-KR & 
\makecell[l]{$\eta_{\mathrm{lr}} = 0.01$; $\alpha_{\mathrm{KS}} = 0.005$;\\
$W_{\mathrm{KS}}=100$; $S_{\mathrm{KS}}=30$} & 
$\alpha_{\mathrm{KS}}$: KSWIN significance level. 
$W_{\mathrm{KS}}$: KSWIN window size. 
$S_{\mathrm{KS}}$: KSWIN statistic size. \\

LMS-KW & 
\makecell[l]{$\eta_{\mathrm{lr}} = 0.01$; $\alpha_{\mathrm{KS}} = 0.005$;\\
$W_{\mathrm{KS}}=100$; $S_{\mathrm{KS}}=30$;\\
$W=50$} & 
$W$: number of recent observations used for WINDOW-based adaptation. \\

LMS-KS & 
\makecell[l]{$\eta_{\mathrm{lr}} = 0.01$; $\alpha_{\mathrm{KS}} = 0.005$;\\
$W_{\mathrm{KS}}=100$; $S_{\mathrm{KS}}=30$;\\
$\mathcal{A}_{\eta}=\{0.001,0.003,0.005,0.008,$\\
$0.01,0.015,0.02,0.03\}$} & 
$\mathcal{A}_{\eta}$: SSPT candidate set used to select or update the LMS learning rate. \\

LMS-KO & 
\makecell[l]{$\eta_{\mathrm{lr}} = 0.01$; $\alpha_{\mathrm{KS}} = 0.005$;\\
$W_{\mathrm{KS}}=100$; $S_{\mathrm{KS}}=30$;\\
$\eta_{\mathrm{OHL}} = 0.02$; $\epsilon_{\mathrm{OHL}} = 0.01$;\\
$\eta_{\mathrm{lr}} \in [10^{-4},0.05]$} & 
$\eta_{\mathrm{OHL}}$: OHL step size. 
$\epsilon_{\mathrm{OHL}}$: OHL perturbation value. 
$[10^{-4},0.05]$: allowed range for the LMS learning rate. \\

\bottomrule
\end{tabularx}
\end{table}

\newpage

\section{Conclusion}

This paper introduced \modelname, a model-agnostic and meta-adaptive
pre-update framework for automated drift detection and adaptation in online
regression. Unlike conventional drift-handling approaches that often rely on
delayed drift alarms, external adaptation policies, or offline hyperparameter
tuning, \modelname operates as a pre-update control layer that jointly
monitors predictive behavior, detects potential drift, quantifies drift
magnitude, adjusts model-specific hyperparameters, and performs bounded
recalibration when needed. By using KPI-window-based monitoring,
CFAR-inspired adaptive thresholding, severity-aware drift classification, and
scale-map-based hyperparameter control, \modelname provides a unified
mechanism for translating performance deviations into proportionate
adaptation actions. The experimental evaluation integrates \modelname{} with
four online regression models and compares it with eight ADWIN/KSWIN-based
detector--adaptation baselines using \(R^2\) and MSE. Across 18 controlled
synthetic datasets covering abrupt, incremental, and alternating gradual
drift and eight real-world datasets spanning heterogeneous application
domains, scales, and dimensionalities, \modelname{} improves overall
predictive performance across the evaluated settings. The synthetic-stream
analysis further evaluates direct raw detector alarms, secondary alarm
episodes, true and false alarms, missed drifts, precision, recall,
\(F_1\)-score, conventional detection delay, protocol sensitivity, runtime,
memory, and normalized intervention activity. Because ground-truth drift
locations are available only for the synthetic streams, direct
drift-detection accuracy is validated only in the controlled synthetic
experiments. The real-world experiments instead evaluate predictive
robustness under naturally occurring and unlabeled temporal variation. These
findings support \modelname{} as a practical and extensible meta-adaptive
framework within the evaluated online regression settings.

\section*{GenAI Usage Disclosure}
The authors employed generative AI tools exclusively for the purposes of rewriting and grammar verification in the preparation of this manuscript. No generative AI tools were utilized for the creation of original research concepts, coding, data, or experimental findings. All technical contributions, encompassing method design, implementation, analysis, and evaluation, were conducted independently by the authors. The application of generative AI was restricted to enhancing language and did not include content generation that could compromise the scientific integrity of the work. This statement is provided in compliance with Elsevier’s Authorship Policy.

\bibliographystyle{elsarticle-num}
\bibliography{sn-bibliography}

\end{document}